\documentclass{article}

\PassOptionsToPackage{numbers}{natbib}

\usepackage[final]{neurips_2024}

\usepackage{graphicx}
\usepackage{amsmath}
\usepackage[utf8]{inputenc} 
\usepackage[T1]{fontenc}    
\usepackage{hyperref}       
\usepackage{url}            
\usepackage{booktabs}       
\usepackage{amsfonts}       
\usepackage{nicefrac}       
\usepackage{microtype}      
\usepackage{xcolor}
\usepackage{algorithm}
\usepackage{microtype}
\usepackage{algcompatible,amsmath}
\usepackage{mdframed} 
\usepackage{caption}
\usepackage{framed}
\usepackage{multirow}
\usepackage{comment}
\usepackage{amssymb}
\usepackage{pifont}
\usepackage{balance}
\usepackage{mathtools}
\usepackage{colortbl}
\usepackage{bbm}
\usepackage{enumitem}
\usepackage{marvosym}
\usepackage{textcomp}
\usepackage{array}
\usepackage{booktabs}
\usepackage{tabularx}
\usepackage{CJKutf8}
\usepackage{xspace}
\usepackage{pdfpages} 
\usepackage{soul}

\usepackage{placeins}
\usepackage{authblk}

\newcommand{\data}{MR-Ben\xspace}

\title{\data: A Meta-Reasoning Benchmark for Evaluating System-2 Thinking in LLMs}

\author{ \textbf{Zhongshen Zeng\textsuperscript{1}   \quad Yinhong Liu\textsuperscript{2} \quad Yingjia Wan\textsuperscript{2} }\\
\textbf{Jingyao Li\textsuperscript{1} \quad Pengguang Chen\textsuperscript{1} \quad Jianbo Dai\textsuperscript{3} \quad Yuxuan Yao\textsuperscript{4}} \\
\textbf{Rongwu Xu\textsuperscript{5} \quad Zehan Qi\textsuperscript{5} \quad Wanru Zhao\textsuperscript{2} \quad Linling Shen\textsuperscript{6}} \\
\textbf{Jianqiao Lu\textsuperscript{7} \quad Haochen Tan\textsuperscript{4} \quad Yukang Chen\textsuperscript{1} \quad Hao Zhang\textsuperscript{8}} \\
\textbf{Zhan Shi\textsuperscript{6} \quad Bailin Wang\textsuperscript{9} \quad Zhijiang Guo\textsuperscript{2\textsuperscript{$\dagger$}} \quad Jiaya Jia\textsuperscript{1\textsuperscript{$\dagger$}}
}}

\affil{\textsuperscript{1}Chinese University of Hong Kong \quad \textsuperscript{2}University of Cambridge \quad \textsuperscript{3}University of Edinburgh \\
\textsuperscript{4}City University of Hong Kong \quad \textsuperscript{5}Tsinghua University \quad \textsuperscript{6}University of Texas at Austin \\
\textsuperscript{7}University of Hong Kong \quad \textsuperscript{8}Nanyang Technological University \\ \textsuperscript{9}Massachusetts Institute of Technology}

\begin{document}

\maketitle

\def\thefootnote{$\dagger$} \footnotetext{Correspondence to: Zhijiang Guo (zg283@cam.ac.uk) and Jiaya Jia (leojia@cse.cuhk.edu.hk).}\def\thefootnote{\arabic{footnote}}

\begin{abstract}


Large language models (LLMs) have shown increasing capability in problem-solving and decision-making, largely based on the step-by-step chain-of-thought reasoning processes. However, evaluating these reasoning abilities has become increasingly challenging. Existing \textit{outcome-based} benchmarks are beginning to saturate, becoming less effective in tracking meaningful progress. To address this, we present a \textit{process-based} benchmark \data that demands a meta-reasoning skill, where LMs are asked to locate and analyse potential errors in automatically generated reasoning steps. Our meta-reasoning paradigm is especially suited for system-2 slow thinking, mirroring the human cognitive process of carefully examining assumptions, conditions, calculations, and logic to identify mistakes.
\data comprises 5,975 questions curated by human experts across a wide range of subjects, including physics, chemistry, logic, coding, and more.
Through our designed metrics for assessing meta-reasoning on this benchmark, we identify interesting limitations and weaknesses of current LLMs (open-source and closed-source models). For example, with models like the o1 series from OpenAI demonstrating strong performance by effectively scrutinizing the solution space, many other state-of-the-art models fall significantly behind on \data, exposing potential shortcomings in their training strategies and inference methodologies\footnote{Our dataset and codes are available on \url{https://randolph-zeng.github.io/Mr-Ben.github.io}.}.

\end{abstract}
\section{Introduction}
\label{sec:intro}

Reasoning, the cognitive process of using evidence, arguments, and logic to reach conclusions, is crucial for problem-solving, decision-making, and critical thinking~\citep{wason1972psychology, FaginH94}. With the rapid advancement of Large Language Models (LLMs), there is an increasing interest in exploring their reasoning capabilities~\citep{HuangSurvey2023,QiaoO0CYDTHC23}. Consequently, evaluating reasoning in LLMs reliably becomes paramount. Current evaluation methodologies primarily focus on the final result~\citep{CobbeGSM8K2021,HendrycksBKABTS21,GevaKSKRB21,SuzgunSSGTCCLCZ23}, disregarding the intricacies of the reasoning process. While effective to some extent, such evaluation practices may conceal underlying issues like logical errors or unnecessary steps that compromise the accuracy and efficiency of reasoning~\citep{DBLP:journals/corr/abs-2404-05692,liu2024measuring}. 

Therefore, it is important to complement outcome-based evaluation with an intrinsic evaluation of the quality of the reasoning process.  However, current benchmarks for evaluating LLMs' reasoning capabilities have certain limitations in terms of their scope and size. For instance, PRM800K~\citep{LightmanVerify2023} categorizes each reasoning step as positive, negative, or neutral. Similarly, BIG-Bench Mistake~\citep{Tyen2023} focuses on identifying errors in step-level answers. We follow the same meta-reasoning paradigm as MR-GSM8K~\citep{DBLP:journals/corr/abs-2312-17080} and MR-Math~\citep{DBLP:journals/corr/abs-2404-05692}, which go a step further by providing the error reason for the first negative step in the reasoning chain. However, these benchmarks are limited to a narrower task scope—MR-GSM8K and MR-Math focus solely on mathematical reasoning, while BIG-Bench Mistake mainly assesses logical reasoning. To ensure a comprehensive evaluation of reasoning abilities, it is crucial to identify reasoning errors and assess the LLMs' capacity to elucidate them across wider domains.

\begin{figure}[t]
    \centering
    \includegraphics[width=\textwidth]{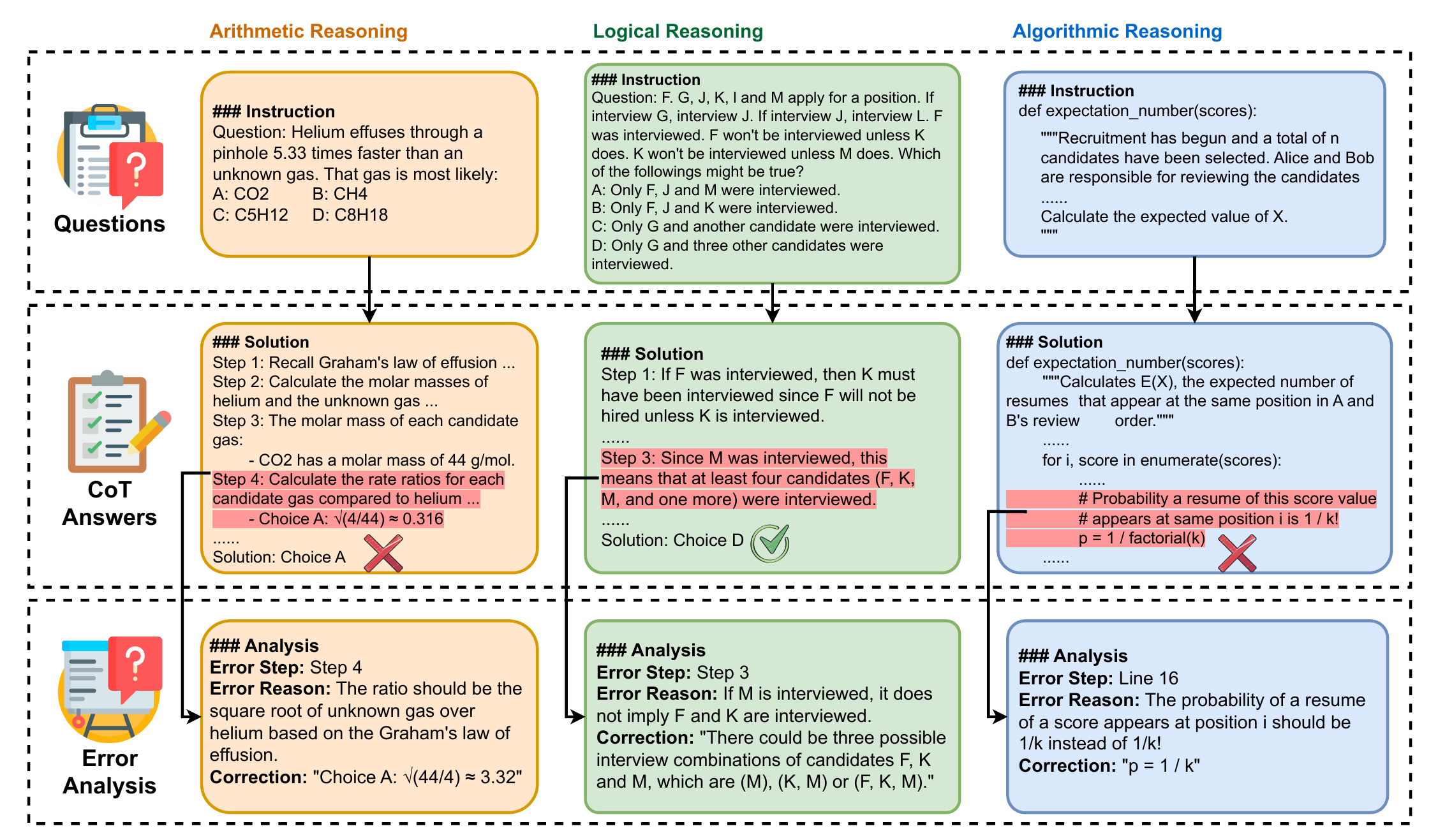}
    \vspace{-0.5cm}
    \caption{
    Overview of the evaluation paradigm and representative examples in \data. Each data point encompasses three key elements: a question, a Chain-of-Thought (CoT) answer, and an error analysis. The CoT answer is generated by various LLMs. Human experts annotate the error analyses, which include error steps, reasons behind the error, and subsequent corrections. The three examples shown are selected to represent arithmetic, logical, and algorithmic reasoning types.}
    \label{fig:main}
\end{figure}

To bridge this gap, we construct a comprehensive benchmark \data comprising 6k questions covering a wide range of subjects, including natural sciences like math, biology, and physics, as well as coding and logic. One unique aspect of \data is its meta-reasoning paradigm, which involves challenging LLMs to reason about different forms of reasoning. In this paradigm, LLMs take on the role of a teacher, evaluating the reasoning process by assessing correctness, analyzing potential errors, and providing corrections, as depicted in Figure~\ref{fig:main}. 

Our analysis of various LLMs~\citep{GPT35turbo, GPT4, Claude2, Mixtral24, Llama3} uncovers distinct limitations and previously unidentified weaknesses in their reasoning abilities. While many LLMs are capable of generating correct answers, they often struggle to identify errors within their reasoning processes and explain the underlying rationale. To excel under our meta-reasoning paradigm, models must meticulously scrutinize assumptions, conditions, calculations, and logical steps, even inferring step outcomes counterfactually. These requirements align with the characteristics of ``System-2'' slow thinking~\citep{kahneman2011thinking, bengio2020system2}, which we believe remains underdeveloped in most of the state-of-the-art models we evaluated.

We suspect that a key reason for this gap lies in current fine-tuning paradigms, which prioritize correct solutions and limit effective exploration of the broader solution space. Echoing this hypothesis, we observed that models like o1-preview~\citep{o1openai}, which reportedly incorporate effective search and disambiguation techniques across trajectories in the solution space, outperform other models by a large margin. Moreover, we found that leveraging high-quality and diverse synthetic data~\citep{abdin2024phi} significantly mitigates this issue, offering a promising path to enhance performance regardless of model size. Additionally, our results indicate that different LLMs excel in distinct reasoning paradigms, challenging the notion that domain-specific enhancements necessarily yield broad cognitive improvements. We hope that \data will guide researchers in comprehensively evaluating their models’ capabilities and foster the development of more robust AI reasoning frameworks.

Our key contributions are summarized as follows: 
\begin{itemize}
    \item We introduced \data, which includes around 6k questions across a wide range of subjects, from natural sciences to coding and logic, and employs a unique meta-reasoning paradigm.
    \item We conduct an extensive analysis of various LLMs on \data, revealing various limitations and previously unidentified weaknesses in their reasoning abilities.
    \item We offer potential pathways for enhancing the reasoning abilities of LLMs and challenge the assumption that domain-specific enhancements necessarily lead to broad improvements.
\end{itemize}

\section{Related Works}
\label{sec:related_works}

\paragraph{Reasoning Benchmarks}
Evaluating the reasoning capabilities of LLMs is crucial for understanding their potential and limitations. While existing benchmarks often assess reasoning by measuring performance on tasks that require reasoning, such as accuracy, they often focus on specific reasoning types like arithmetic, knowledge, logic, or algorithmic reasoning. Arithmetic reasoning, involving mathematical concepts and operations, has been explored in benchmarks ranging from elementary word problems~\citep{Koncel-Kedziorski16,AminiGLKCH19,PatelBG21,CobbeGSM8K2021} to more complex and large-scale tasks~\citep{HendrycksBKABTS21,MishraFLTWBRTSC22}. Knowledge reasoning, on the other hand, requires either internal (commonsense) or external knowledge, or a combination of both~\citep{ClarkARC2018, TalmorHLB19, GevaKSKRB21}. Logical reasoning benchmarks, encompassing deductive and inductive reasoning, use synthetic rule bases for the former~\citep{ClarkTR20, TafjordDC21, DalviJTXSPC21} and specific observations for the latter to formulate general principles~\citep{ZhangACRE2021, YangDDCCLGW24}. 
Algorithmic reasoning often involves understanding the coding problem description and performing multi-step reasoning to solve it~\citep{Dai2024mhpp, GuRLSS024}. Benchmarks like BBH~\citep{SrivastavaBBH2022} and MMLU~\citep{hendrycks2020measuring} indirectly assess reasoning by evaluating performance on tasks that require it. 
However, these benchmarks primarily focus on final results, neglecting the analysis of potential errors in the reasoning process. Unlike prior efforts, \data goes beyond accuracy by assessing the ability to locate potential errors in the reasoning process and provide explanations and corrections. Moreover, \data covers different types of reasoning, offering a more comprehensive assessment.

\paragraph{Evaluation Beyond Accuracy}
Many recent studies have shifted their focus from using only the final result to evaluating the reasoning quality beyond accuracy. This shift has led to the development of two approaches: reference-free and reference-based evaluation. Reference-free methods aim to assess reasoning quality without relying on human-provided solutions. For example, ROSCOE~\citep{GolovnevaCPCZFC23} evaluates reasoning chains by quantifying reasoning errors such as redundancy and hallucination. Other approaches convert reasoning steps into structured forms, like subject-verb-object frames~\citep{PrasadSZB23} or symbolic proofs~\citep{Saparov023}, allowing for automated analysis. Reference-based methods depend on human-generated step-by-step solutions. For instance, PRM800K~\citep{LightmanVerify2023} offers solutions to MATH problems~\citep{HendrycksBKABTS21}, categorizing each reasoning step as positive, negative, or neutral. Building on this, MR-GSM8K~\citep{DBLP:journals/corr/abs-2312-17080} and MR-Math~\citep{DBLP:journals/corr/abs-2404-05692} further provide the error reason behind the first negative step. MR-GSM8K focuses on elementary math problems, sampling questions from GSM8K~\citep{CobbeGSM8K2021}. MR-Math samples a smaller set of 459 questions from MATH~\citep{HendrycksBKABTS21}. Using the same annotation scheme, BIG-Bench Mistake~\citep{Tyen2023} focuses on symbolic reasoning. It encompasses 2,186 instances from 5 tasks in BBH~\citep{SrivastavaBBH2022}. Despite the progress made by these datasets, limitations in scope and size remain. To address this, we introduce \data, a benchmark consisting of 5,975 manually annotated instances covering a wide range of subjects, including natural sciences, coding, and logic. \data also features more challenging questions, spanning high school, graduate, and professional levels.

\section{\data: Dataset Construction}
\label{sec:dataset}

\subsection{Dataset Structure}
\label{ssec:structure}

To comprehensively evaluate the reasoning capabilities of LLMs, \data employs a meta-reasoning paradigm. This paradigm casts LLMs in the role of a teacher, where they assess the reasoning process by evaluating its correctness, analyzing errors, and providing corrections. As shown in Figure~\ref{fig:main}, each data point within \data consists of three key elements: a question, a CoT answer, and an error analysis. The construction pipeline is shown in Figure~\ref{fig:dataset_pipeline} in Appendix-\ref{app:Eval_Prompt}. 

\paragraph{Question} The questions in \data are designed to cover a diverse range of reasoning types and difficulty levels, spanning from high school to professional levels. To ensure this breadth, we curated questions from various subjects, including natural sciences (mathematics, biology, physics), coding, and logic. Specifically, we sampled questions from mathematics, physics, biology, chemistry, and medicine from MMLU~\citep{hendrycks2020measuring}, which comprehensively assesses LLMs across academic and professional domains. For logic questions, we draw from LogiQA~\citep{liu2020logiqa}, which encompasses a broad spectrum of logical reasoning types, including categorical, conditional, disjunctive, and conjunctive reasoning.  Finally, we select coding problems from MHPP~\citep{Dai2024mhpp}, which focuses on function-level code generation requiring advanced algorithmic reasoning. Questions in MMLU and LogiQA require a single-choice answer, while MHPP requires a snippet of code as the answer.



\paragraph{CoT Answer } We queried GPT-3.5-Turbo-0125~\citep{GPT35turbo}, Claude2~\citep{Claude2}, and Mistral-Medium~\citep{Mistral23} (as of February 2024) using a prompt template (provided in Figure-\ref{fig:sol_gen_prompt} in Appendix-\ref{app:Eval_Prompt}) designed to elicit step-by-step solutions~\citep{Wei0SBIXCLZ22}. For clarity, all LLMs were instructed to format their solutions with numbered steps, except for coding problems. To encourage diverse solutions, we set the temperature parameter to 1 during sampling. This empirical setting yielded satisfactory instruction following and desirable fine-grained reasoning errors, which annotators and evaluated models are expected to identify.


\subsection{Annotation Process}
\label{ssec:process}

After acquiring the questions and their corresponding Chain-of-Thought (CoT) answers, we engage annotators to provide error analyses. The annotation process is divided into three stages.

\paragraph{Answer Correctness} 
CoT answers that result in a final answer different from the ground truth are automatically flagged as incorrect. However, for cases where the final answer matches the ground truth, manual annotation is required. This is because there are instances where the reasoning process leading to the correct answer is flawed, as illustrated in the middle example of Figure~\ref{fig:main}. Therefore, annotators are tasked with meticulously examining the entire reasoning path to determine if the correct final answer is a direct result of the reasoning process.

\paragraph{Error Step} This stage is applicable for solutions with either an unmatched final output or a matched final output underpinned by flawed reasoning. Following the prior effort~\citep{LightmanVerify2023}, each step in the reasoning process is categorized as positive, neutral, or negative. Positive and neutral steps represent stages where the correct final output remains attainable. Conversely, negative steps indicate a divergence from the path leading to the correct solution. Annotators are required to identify the first step in the reasoning process where the conditions, assumptions, or calculations are incorrect, making the correct final result unreachable for the subsequent reasoning steps.

\paragraph{Error Reason and Correction} Annotators are tasked with conducting an in-depth analysis of the reasoning that led to the identified error. As shown in Figure~\ref{fig:main}, annotators are required to provide the error reason and the corresponding correction to this reasoning step. This comprehensive approach ensures a thorough understanding and rectification of errors in the reasoning process.






\subsection{Data Statistics}
\label{ssec:stats}


\begin{table}
\caption{Statistics of \data. The length of questions and solutions are measured inthe  number of words. Notice that the steps for coding denote the number of lines of code. They are not directly comparable with other subjects. }
\centering
\resizebox{\textwidth}{!}{
\begin{tabular}{lcccccccc}
\toprule
 & \textbf{Math} & \textbf{Medicine} & \textbf{Biology} & \textbf{Physics} & \textbf{Chemistry} & \textbf{Logic} & \textbf{Coding} & \textbf{Total} \\
\midrule
Question-Solution Pairs & 918 & 828 & 1035 & 667 & 848 & 1441 & 238 & 5975 \\
Correct Solution Ratio & 16.2\% & 31.0\% & 59.6\% & 47.8\% & 45.0\% & 51.1\% & 31.1\% & 40.3\% \\
Avg Solution Steps & 6.8 & 5.3 & 5.1 & 5.7 & 5.6 & 5.3 & 32.5* & 9.5 \\
Avg First Error Step  & 3.1 & 3.0 & 2.7 & 3.1 & 3.0 & 2.8 & 14.0* & 4.5 \\
Avg Length of Questions & 44.3 & 88.7 & 56.3 & 66.6 & 48.1 & 154.8 & 140.1 & 85.6 \\
Avg Length of Solutions & 205.9 & 206.1 & 187.6 & 199.4 & 194.5 & 217.7 & 950.3 & 308.8 \\
\bottomrule
\end{tabular}
}
\label{tab:stats}
\end{table}

Table~\ref{tab:stats} presents the statistics of \data. 
The benchmark exhibits a balanced distribution of correct and incorrect solutions, with an overall correct solution rate of 40.3\%. Solutions, on average, involve 9.5 steps, and errors typically manifest around the fourth step (4.5). The questions and solutions are substantial, with average lengths of 85.6 and 308.8 words, respectively. The subject-wise analysis reveals that Math is the most challenging, with a correct solution rate of a mere 16.2\%. This could be attributable to the intricacy of the arithmetic operations involved. Conversely, Biology emerges as the least daunting, with a high correct solution rate of 59.6\%. 
Coding problems have the longest solutions, averaging 950.3 number of words. This underscores the complexity and the detailed procedural reasoning inherent in coding tasks. Similarly, Logic problems have the longest questions, averaging 154.8 words. This is in line with the need for elaborate descriptions in logical reasoning. The typical step at which the first error occurs is fairly consistent across most subjects, usually around the 3rd step out of a total of 5. However, Coding deviates from this trend. The first error tends to appear earlier, specifically around the 14th line out of a total of 32.5 lines. This suggests that the problem-solving process in Coding may have distinct dynamics compared to other subjects.

\subsection{Quality Control}
\label{ssec:quality}

\paragraph{Annotators} 
Given the complexity of the questions, which span a range of subjects from high school to professional levels, we enlisted the services of an annotation company. This company meticulously recruited annotators, each holding a minimum of a bachelor's degree. Before their trial labeling, annotators are thoroughly trained and are required to review the annotation guidelines. We've included the guidelines for all subjects in Appendix~\ref{app:guideline} for reference. The selection of annotators is based on their performance on a balanced, small hold-out set of problems for each subject. In addition to the annotators, a team of 14 quality controllers diligently monitors the quality of the annotation weekly. As a final layer of assurance, we have 4 meta controllers who scrutinize the quality of the work.


\paragraph{Quality Assurance}  Every problem in \data undergoes a rigorous three-round quality assurance process to ensure its accuracy and clarity. Initially, each question is labeled by two different annotators. Any inconsistencies in the solution correctness or the first error step are identified and reviewed by a quality controller for arbitration. Following this, every annotated problem is subjected to a secondary review by annotators who were not involved in the initial labeling. This is to ensure that the annotations for different solutions to the same problem are consistent and coherent. In the final phase of the review, 10\% of the problems are randomly sampled and reviewed by the meta controllers. Throughout the entire evaluation process, all annotated fields are meticulously examined in multiple rounds for their accuracy and clarity. Any incorrect annotations or those with disagreements are progressively filtered out and rectified, ensuring a high-quality dataset. This rigorous process allows us to maintain a high level of annotation quality.


\paragraph{Dataset Artifacts \& Biases}
Table~\ref{tab:stats} reveals a relatively balanced distribution of correct and incorrect solutions. However, an exception was observed in mathematical subjects, where the distribution tends to skew towards incorrect solutions. This skew could suggest an inherent complexity or ambiguity in mathematical problem statements. Our analysis of the first error step across all subjects indicated that errors predominantly occur in the initial stages ($n \leq 7$) of problem-solving and are distributed relatively uniformly. This pattern was consistent across most subjects, with no significant skew towards later steps. More detailed discussions of biases are provided in the Appendix~\ref{app:bias_correlation}.


\section{Evaluation}
\label{sec:evaluation}

For each question-solution pair annotated, the evaluated model are supposed to decide the correctness of the solution and report the first-error-step and error-reason if any. 
The solution-correctness and first-error-step is scored automatically based on the manual annotation result. 
Only when the evaluated model correctly identified the incorrect solution and first-error-step will its error-reason be further examined manually or automatically by models. Therefore in order to provide a unified and normalized score to reflect the overall competence of the evaluated model, we follow the work of \cite{DBLP:journals/corr/abs-2312-17080} and apply a metric named \textbf{MR-Score}, which consist of three sub-metrics. 

The first one is the Matthews Correlation Coefficient (a.k.a MCC, \citealp{Matthews1975ComparisonOT}) for the binary classification of solution-correctness.
\begin{equation} \label{eq1}
\begin{aligned}
MCC = \frac{TP \times TN - FP \times FN}{\sqrt{(TP + FP) \times (TP + FN) \times (TN + FP) \times (TN + FN) }} 
\end{aligned}
\end{equation}
where TP, TN, FP, FN stand for true positive, true negative, false positive and false negative. The MCC score ranges from -1 to +1 with -1 means total disagreement between prediction and observation, 0 indicates near random performance and +1 represents perfect prediction. In the context of this paper, we interpret negative values as no better than random guess and set 0 as cut-off threshold for normalization purpose.

The second metric is the ratio between numbers of solutions with correct first-error-step predicted and the total number of incorrect solutions. 
  
\begin{equation} \label{eq2}
\begin{split}
ACC_{\text{step}} = \frac{N_{\text{correct\_first\_error\_step}}}{N_{\text{incorrect\_sols}}}
\end{split}
\end{equation}

The third metrics is likewise the ratio between number of solutions with correct first-error-step plus correct error-reason predicted and the total number of incorrect solutions.
\begin{equation} \label{eq3}
\begin{split}
ACC_{\text{reason}} = \frac{N_{\text{correct\_error\_reason}}}{N_{\text{incorrect\_sols}}}
\end{split}
\end{equation}

\textbf{MR-Score} is then a weighted combination of three metrics, given by  
\begin{equation} \label{eq4}
\begin{aligned}
MR\mbox{-}Score &= w_1 * \max(0,MCC) + w_2 * ACC_{\text{step}} 
 + w_3 * ACC_{\text{reason}} 
\end{aligned}
\end{equation}

For the weights $w_1, w_2$ and $w_3$, they are chosen based on our evaluation results to maximize the differentiation between different models. It is important to note that the Matthews Correlation Coefficient (MCC) and the accuracy of locating the first error step can be directly calculated by comparing the responses of the evaluated model with the ground truth annotations. However, assessing the accuracy of the error reason explained by the evaluated model presents more complexity. While consulting domain experts for annotations is a feasible approach, we instead utilized GPT-4-Turbo as a proxy to examine the error reasons, as detailed in Figure-\ref{fig:error_reason_prompt} in Appendix-\ref{app:Eval_Prompt}.

We operate under the assumption that while our benchmark presents a significant challenge for GPT-4 in evaluating complete solution correctness—identifying the first error step and explaining the error reason—it is comparatively easier for GPT-4 to assess whether the provided error reasons align with the ground truth. Specifically, in a hold-out set of sampled error reasons, there was a 92\% agreement rate between the manual annotations by the authors and those generated by GPT-4. For more detailed evaluations on the robustness of MR-Score and its design thinking, please refer to our discussion in Appendix-\ref{app:mr_score_robustness}.

\section{Experiments}
\label{sec:experiments}

\subsection{Experiment Setup}
\label{sec:experiment setup}

\begin{table*}[t]
\caption{Evaluation results on \data: This table presents a detailed breakdown of each model’s performance evaluated under metric MR-Score across different subjects, where K stands for the number of demo examples here.}
\setlength\extrarowheight{3.5pt}
\renewcommand{\arraystretch}{0.85}
\setlength{\tabcolsep}{2pt}

\resizebox{\textwidth}{!}{
\begin{tabular}{lrrp{1.5mm}rrp{1.5mm}rrp{1.5mm}rrp{1.5mm}rrp{1.5mm}rrp{1.5mm}rrp{1.5mm}rr}
\toprule
\multirow{2}{*}{\textbf{Model}} & \multicolumn{2}{c}{\textbf{Bio.}} & & \multicolumn{2}{c}{\textbf{Phy.}} && \multicolumn{2}{c}{\textbf{Math}} && \multicolumn{2}{c}{\textbf{Chem.}} && \multicolumn{2}{c}{\textbf{Med.}} && \multicolumn{2}{c}{\textbf{Logic}} && \multicolumn{2}{c}{\textbf{Coding}} && \multicolumn{2}{c}{\textbf{Avg.}} \\[1pt] 
\cline{2-3} \cline{5-6} \cline{8-9} \cline{11-12} \cline{14-15} \cline{17-18} \cline{20-21} \cline{23-24} 
& $k$=0 & $k$=1 && $k$=0  & $k$=1  && $k$=0 & $k$=1  && $k$=0 & $k$=1 && $k$=0  & $k$=1  && $k$=0 & $k$=1  && $k$=0 & $k$=1 && $k$=0  & $k$=1   \\ 
 
\midrule
\multicolumn{24}{c}{\textbf{Closed-Source LLMs}}\\
\midrule

Claude3-Haiku & 5.7 & 5.8 && 3.3 & 3.5 && 3.1 & 3.1 && 6.5 & 6.4 && 2.0 & 2.0 && 1.2 & 1.2 && 9.0 & 0.0 && 4.4 & 3.1  \\
GPT-3.5-Turbo & 3.6 & 6.6 && 5.7 & 6.7 && 5.7 & 5.4 && 4.9 & 6.7 && 3.6 & 4.4 && 1.7 & 4.5 && 3.0 & 4.1 && 4.0 & 5.5  \\

Doubao-pro-4k & 8.4 & 13.5 && 10.0 & 11.7 && 12.3 & 15.5 && 10.6 & 17.5 && 5.9 & 10.0 && 4.5 & 5.5 && 9.8 & 7.4 && 8.8 & 11.6  \\

Mistral-Large & 22.2 & 28.0 && 26.7 & 25.4 && 24.3 & 28.2 && 24.0 & 27.0 && 15.9 & 19.3 && 14.7 & 17.1 && 21.1 & 21.4 && 21.3 & 23.8  \\

Yi-Large & 35.3.& 40.7 && 37.2 & 36.8 && 36.5 & 20.6 && 40.0 & 39.1 && 29.3 & 32.1 && 25.1 & 31.3 && 21.9 & 25.7 && 32.2 & 32.3  \\

Moonshot-v1-8k & 35.0 & 36.8 && 33.8 & 33.8 && 34.9 & 33.0 && 36.7 & 35.0 && 29.4 & 32.3 && 25.0 & 29.2 && 32.7 & 31.2 && 32.5 & 33.0  \\

GPT-4o-mini & 37.7 & 38.9 && 38.5 & 37.4 && 44.4 & 40.4 && 39.2 & 37.0 && 33.9 & 25.1 && 23.6 & 17.7 && 41.6 & 34.9 && 37.0 & 33.1  \\

Zhipu-GLM-4 & 40.7 & 46.2 && 37.7 & 42.5 && 38.4 & 36.6 && 43.1 & 44.0 && 34.5 & 41.0 && 37.5 & 32.5 && 38.8 & 32.8 && 38.7 & 39.4  \\

GPT-4-Turbo & 44.7 & 47.3 && 42.8 & 45.2 && 44.3 & 45.4 && 44.0 & 46.0 && 38.8 & 38.4 && 34.1 & 33.6 && 53.6 & 57.3 && 43.2 & 44.7  \\

GPT-4o & 48.3 & 49.1 && 45.5 & 48.2 && 42.6 & 41.3 && 48.2 & 49.1 && 47.9 & 47.7 && 31.9 & 28.4 && 56.5 & 54.6 && 45.8 & 45.5  \\

o1-mini & 45.8 & 46.9 && 56.0 & 53.8 && 68.5 & 67.0 && 55.2 & 56.1 && 45.9 & 47.2 && 30.7 & 28.7 && 55.1 & 55.6 && 51.0 & 50.8  \\

o1-preview & 54.1 & 56.0 && 62.2 & 61.7 && 69.8 & 70.3 && 60.6 & 60.3 && 54.3 & 55.1 && 46.1 & 45.3 && 65.1 & 70.0 && \textbf{58.9} & \textbf{59.8}  \\

\midrule
\multicolumn{24}{c}{\textbf{Open-Source Small}}\\
\midrule
Qwen1.5-1.8B & 0.0 & 0.0 && 0.0 & 0.0 && 0.0 & 0.1 && 0.0 & 0.1 && 0.0 & 0.0 && 0.0 & 0.1 && 0.0 & 0.0 && 0.0 & 0.0  \\
Gemma-2B & 0.1 & 0.0 && 0.0 & 0.0 && 0.0 & 1.0 && 0.1 & 0.0 && 0.0 & 0.4 && 0.0 & 0.2 && 0.7 & 0.0 && 0.1 & 0.2 \\
Qwen2-1.5B & 2.2 & 2.8 && 2.2 & 1.3 && 3.3 & 6.3 && 2.5 & 3.3 && 2.9 & 11.2 && 1.5 & 9.4 && 0.0 & 3.6 && 2.1 & 5.4  \\
Phi3-3.8B & 13.4 & 12.5 && 12.7 & 10.8 && 13.3 & 13.1 && 16.4 & 17.1 && 10.2 & 8.1 && 8.4 & 5.3 && 9.1 & 10.2 && \textbf{11.9} & \textbf{11.0}   \\

\midrule
\multicolumn{24}{c}{\textbf{Open-Source LLMs Medium}}\\
\midrule

GLM-4-9B  & 4.4 & 2.4 && 9.6 & 1.2 && 8.1 & 4.7 && 8.7 & 2.9 && 2.3 & 1.9 && 2.5 & 1.6 && 11.4 & 0.0 && 6.7 & 2.1  \\

DeepSeek-7B & 5.7 & 6.2 && 4.7 & 2.6 && 4.9 & 5.2 && 4.2 & 4.9 && 3.1 & 1.6 && 3.0 & 3.8 && 0.0 & 1.2 && 3.7 & 3.6  \\

Deepseek-Coder-33B & 7.4 & 5.5 && 7.8 & 5.6 && 7.2 & 8.6 && 7.8 & 7.4 && 6.0 & 5.5 && 4.6 & 6.7 && 8.4 & 4.9 && 7.0 & 6.3  \\

DeepSeek-Coder-7B & 10.5 & 9.9 && 11.8 & 9.6 && 11.8 & 12.1 && 12.3 & 11.9 && 10.4 & 11.0 && 9.8 & 10.7 && 5.0 & 5.8 && 10.2 & 10.2  \\

LLaMA3-8B & 12.0 & 11.9 && 10.9 & 7.5 && 15.0 & 9.0 && 12.6 & 12.7 && 9.3 & 8.0 && 9.4 & 9.6 && 15.8 & 10.0 && \textbf{12.2} & 9.8  \\

Yi-1.5-9B & 10.4 & 14.8 && 11.9 & 12.9 && 12.5 & 15.6 && 13.1 & 14.4 && 9.5 & 14.8 && 9.1 & 9.5 && 4.8 & 6.3 && 10.2 & \textbf{12.6}  \\

\midrule
\multicolumn{24}{c}{\textbf{Open-Source LLMs Large}}\\
\midrule

Qwen1.5-72B & 15.3 & 19.2 && 12.9 & 13.6 && 12.0 & 10.0 && 13.9 & 16.3 && 11.7 & 14.7 && 10.4 & 12.9 && 3.9 & 5.9 && 11.5 & 13.3  \\

DeepSeek-67B & 17.1 & 19.7 && 14.9 & 17.3 && 15.4 & 16.2 && 16.3 & 20.6 && 14.7 & 12.2 && 13.6 & 14.3 && 14.5 & 15.2 && 15.2 & 16.5  \\

LLaMA3-70B & 20.4 & 27.1 && 17.4 & 20.5 && 14.9 & 15.8 && 19.5 & 25.1 && 16.3 & 19.3 && 16.3 & 16.8 && 29.8 & 16.7 && 19.2 & 20.2  \\

DeepSeek-V2-236B & 30.0 & 37.1 && 32.2 & 36.5 && 32.2 & 30.0 && 32.5 & 35.4 && 26.5 & 32.4 && 23.6 & 27.4 && 34.2 & 27.1 && 30.2 & 32.3  \\

Qwen2-72B & 36.0 & 40.8 && 36.7 & 40.9 && 38.0 & 38.7 && 37.2 & 38.8 && 28.3 & 29.3 && 25.6 & 20.5 && 31.3 & 30.4 && \textbf{33.3} & \textbf{34.2}  \\

\bottomrule
\end{tabular}
}
\label{tab:main_table}
\end{table*}

To evaluate the performance of different models on our new benchmark, we selected a diverse array of models based on size and source accessibility \footnote{ Note: All models used in our experiments are instruction-finetuned versions, although this is not indicated in their abbreviated names}. This included smaller models like Gemma-2B\cite{team2024gemma}, Phi-3\cite{abdin2024phi}, Qwen1.5-1.8B \cite{bai2023qwen}, as well as larger counterparts such as Llama3-70B \cite{Llama3}, Deepseek-67B\cite{DBLP:journals/corr/abs-2401-02954}, and Qwen1.5-72B\cite{bai2023qwen}. We also compared open-source models (e.g. models from the Llama3 and Qwen1.5/Qwen2 series) against closed-source models from the GPT \cite{GPT4}, Claude \cite{Claude3}, Mistral \cite{Mistral23}, GLM \cite{ChatGLM}, Yi \cite{LingYiWanWu}, Moonshot \cite{MoonshotKimi}, Doubao \cite{Doubao} families. Additionally, models from the Deepseek-Coder \cite{DBLP:journals/corr/abs-2401-02954} series were included to assess the impact of coding-focused pretraining on reasoning performance.

Given the complexity of our benchmark, even larger open-source models like Llama3-70B-Instruct struggle to produce accurate evaluation results without the use of prompting methods, often achieving MR-Scores near zero. Consequently, we employed a step-wise chain-of-thought prompting technique similar to those described in \citep{DBLP:journals/corr/abs-2312-17080, Tyen2023}. This approach guides models in systematically reasoning through solution traces before making final decisions, as detailed in Appendix-\ref{app:Eval_Prompt}.

Considering the complexity of the task, which includes question comprehension, reasoning through the provided solutions, and adhering to format constraints, few-shot demonstration setups are also explored to investigate if models can benefit from In-Context Learning (ICL) examples. Due to the context token limits, we report zero and one-shot results in the main result table (Table~\ref{tab:main_table})\footnote{For the breakdown performances of models in the sub-tasks, please refer to Table~\ref{tab:main_table_sub_metrics}}. The performance of additional few-shot configurations on a selection of models with various capabilities is further discussed in Section \ref{sec:few_shot}.       

\begin{figure}
\centering
  \vspace{-1mm}
  \begin{minipage}{0.51\textwidth}
    \centering
    \includegraphics[width=\textwidth]{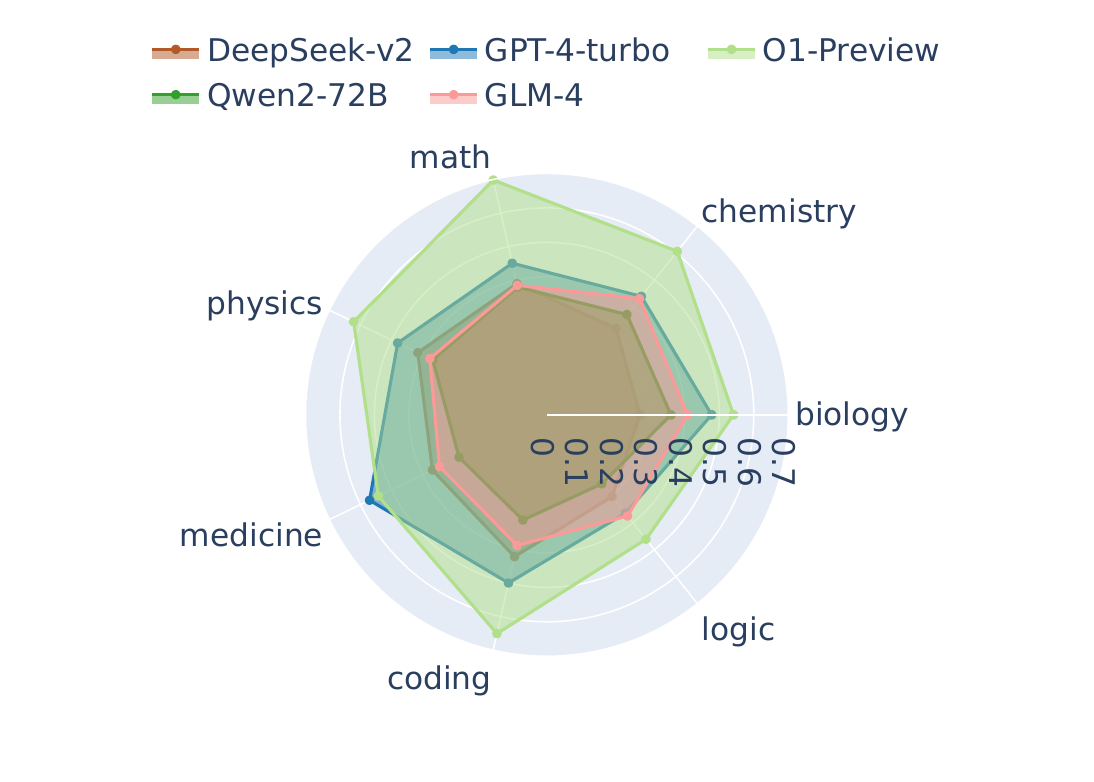}
    \caption{Model performance across subjects}
    \label{fig:radar_plot}
  \end{minipage}%
  \hfill
  \begin{minipage}{0.49\textwidth}
    \centering
    \includegraphics[width=\textwidth]{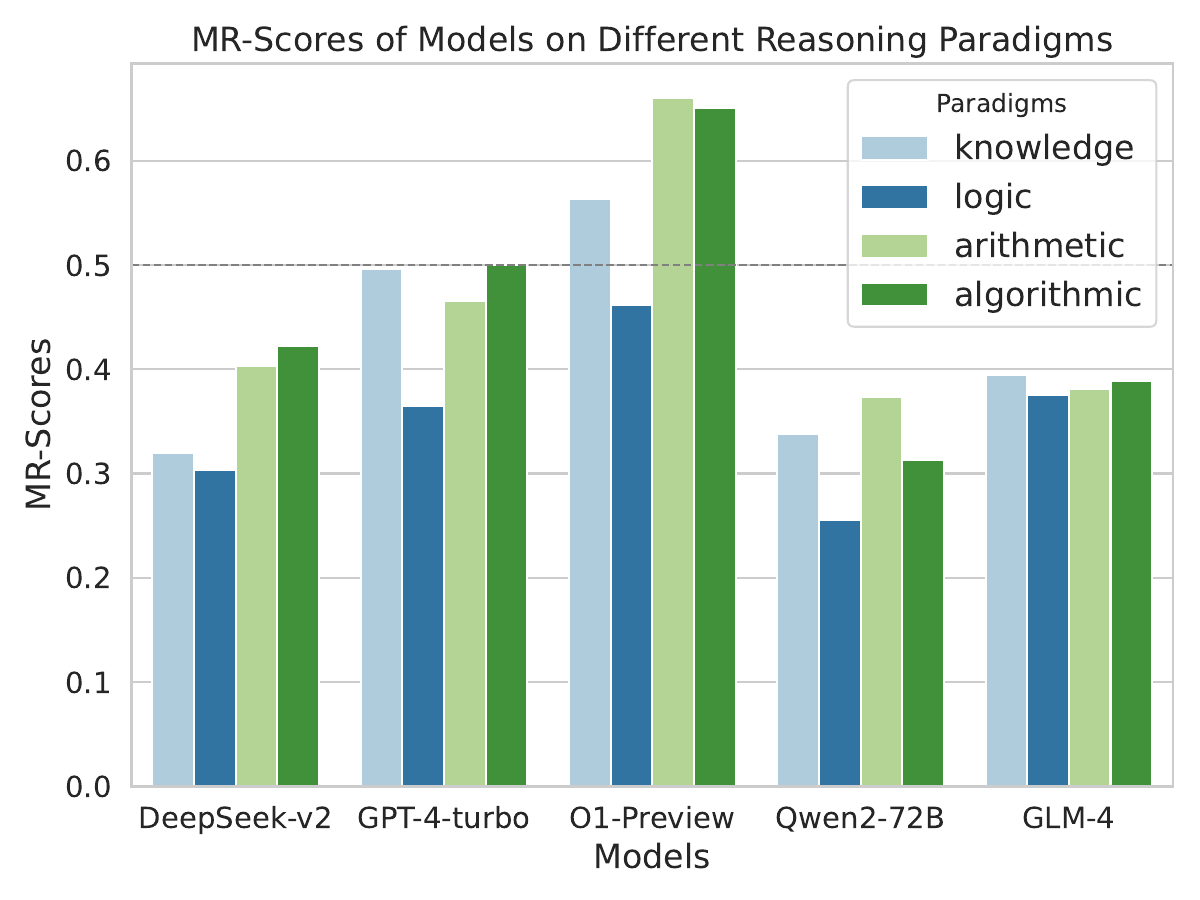}
    \caption{Model performance on different reasoning paradigms}
    \label{fig:reasoning_bars}
  \end{minipage}
  \label{fig:radar_bar_plot} 
\end{figure}

\subsection{Experiment Results}
The \data benchmark presents a significant shift in the challenge for state-of-the-art large language models, transitioning from question-answering to the nuanced role of question-solution scoring. This section details our findings, emphasizing variations in model performances and their implications.

\paragraph{Overall Performance} Among the evaluated models, o1-preview consistently achieves the highest MR-Scores across all subjects, significantly outperforming most competitors from both open and closed-source communities. Notably, the open-sourced Qwen2-72B and Deepseek-V2-236B models are performing exceptionally well, surpassing every other open-sourced model including Llama3 by a large margin. Their scores are even comparable to or greater than some of the most capable models from commercial companies, such as Mistral, Yi, and Moonshot AI. In the small language model category, the performance of Phi3-3.8B exceeds many of the mid-size models, including Deepseek-Coder-33B, whose size is around tenfold larger.     

\paragraph{Performance across Model Size and Reasoning Paradigm} Table~\ref{tab:main_table} reveals a general trend where larger models tend to perform better, highlighting the correlation between model size and the efficacy in complex reasoning tasks. However, this relationship is not strictly linear, as demonstrated by models like Phi3-3.8B, which excel despite their smaller size. Since \data challenges the language models to reason about the reasoning in the solution space among a diverse range of domains, models like Phi-3 that are trained with effective data synthesis techniques and broader coverage of the solution space, intuitively achieve higher MR-Score. This suggests that while larger model sizes generally yield superior performances, techniques like knowledge distillation can also significantly boost reasoning performance. Similarly, although the size of the o1 model series remains undisclosed, these models reportedly employ mechanisms that scale computation efficiently through effective exploration, frequent retrospection, and meticulous reflection within the solution space. These characteristics align closely with the principles of ``system-2'' thinking, which emphasizes deliberate, reflective problem-solving. As a result, the o1 models demonstrate a more effective reasoning process, achieving significantly higher MR-Scores than other models by a large margin.

\paragraph{Performance across Reasoning Types} Our categorization into four reasoning types—knowledge, arithmetic, algorithmic, and logic—illustrates the unique challenges each model faces within these paradigms (Figure~\ref{fig:reasoning_bars}). Logic reasoning emerges as the most formidable due to the intricate logical operations required by questions from the LogiQA dataset. In stark contrast, o1-Preview and GPT-4-turbo demonstrate exceptional prowess in algorithmic reasoning, where their capabilities markedly surpass other models. Notably, models excel in different reasoning paradigms, reflecting their varied strengths and training backgrounds. For instance, despite Deepseek-Coder's specialized pre-training focused on coding tasks, it does not necessarily confer superior abilities in algorithmic reasoning, underscoring that targeted pretraining does not guarantee enhanced performance across all reasoning types. Comparing the performance of the Deepseek-Coder with that of the Phi-3 model, which excels despite its much smaller size, highlights the potential significance of high-quality synthetic data in achieving broad-based reasoning capabilities.

\begin{figure}
\centering
  \vspace{-3mm}
  \begin{minipage}{0.48\textwidth}
        \centering
        \includegraphics[width=\textwidth]{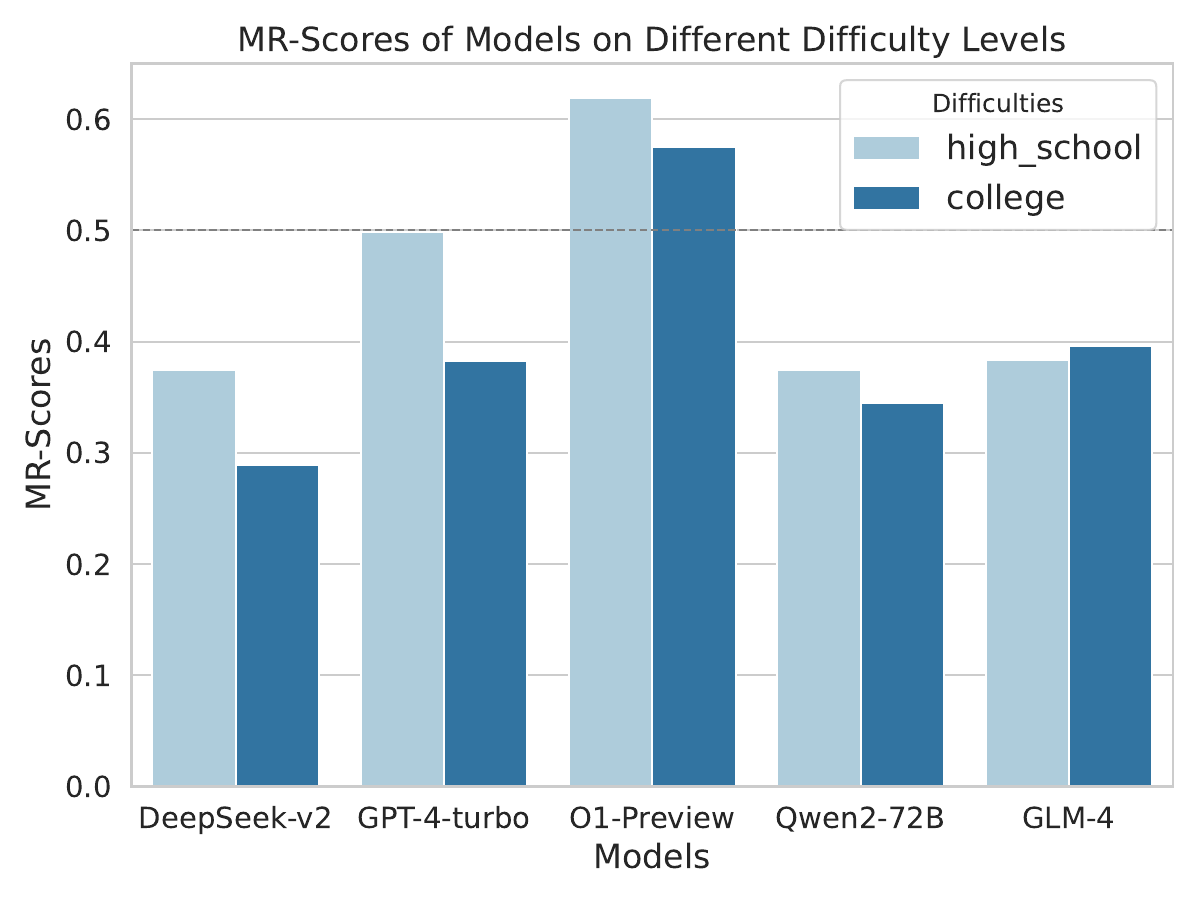}
        \vspace{-0.7cm}
        \caption{MR-Scores of different models on different levels of difficulty } 
        \label{fig:bar_chart} 
  \end{minipage}%
  \hfill
  \begin{minipage}{0.48\textwidth}
        \centering
        \includegraphics[width=\textwidth]{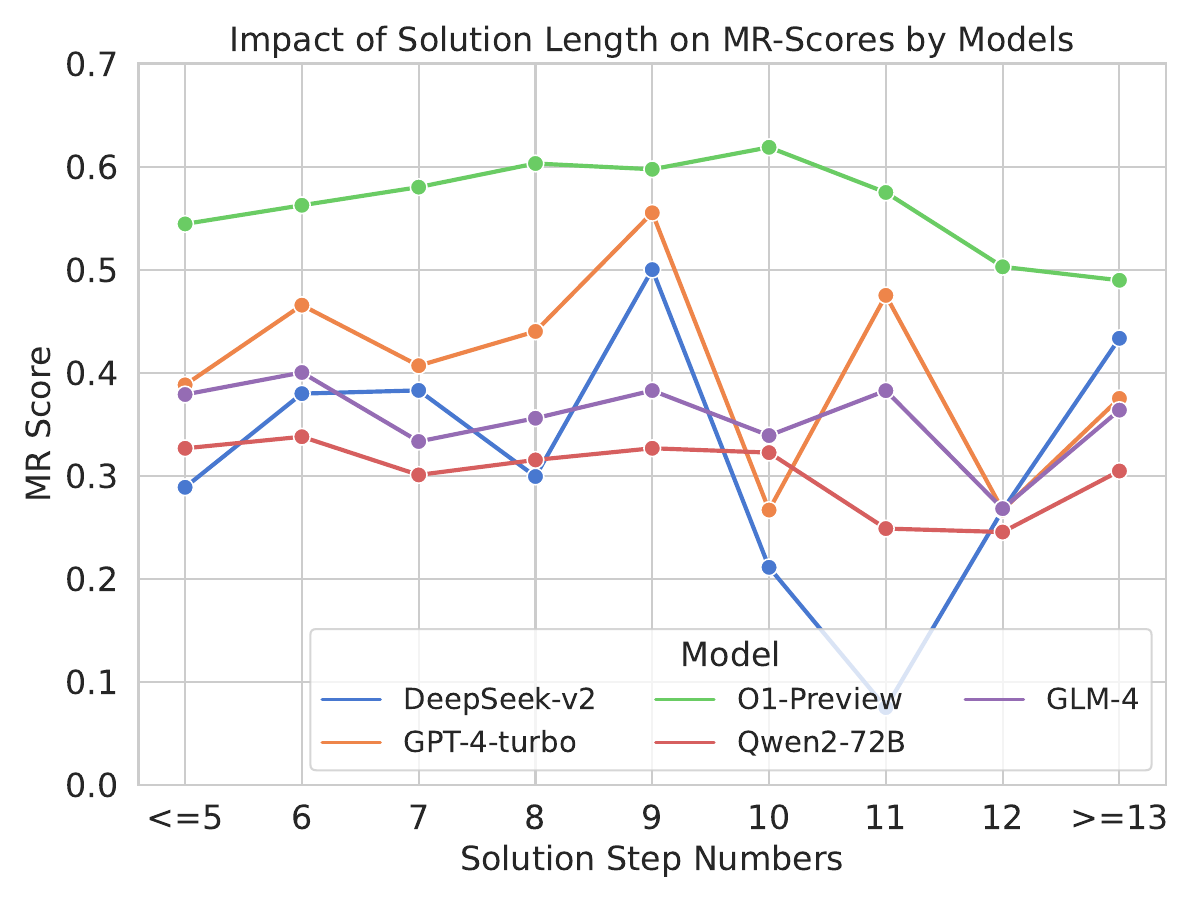}
        \vspace{-0.7cm}
        \caption{The MR-Scores of models on solutions with different step numbers.}
        \label{fig:line_chart} 
  \end{minipage}
  \label{fig:line_bar_plot} 
\end{figure}

\paragraph{Sensitivity to Task Difficulty and Solution Length} An examination across educational levels shows most models perform better at high school-level questions than college-level ones, indicating an intuitive level of sensitivity to the difficulty levels of the questions. Additionally, our analysis finds a minor negative correlation between the length of solution steps and MR-Scores, as detailed in Figure~\ref{fig:bar_chart} and Figure~\ref{fig:line_chart}.

\paragraph{Summary:} \data effectively differentiates model capabilities, often obscured in simpler settings. It not only identifies top performers but also underscores the influence of model size on outcomes, while demonstrating that techniques like knowledge distillation and test-time compute scaling, as seen with the Phi-3 and o1 models, can notably enhance smaller models' performance, challenging the dominance of larger models. The analysis further reveals that specialized training, such as in coding, does not guarantee superior algorithmic reasoning. This suggests the potential need for more balanced data approaches or improved data synthesis methods.

\section{Further Analysis \& Discussion}

\subsection{Few Shot Prompting}
\label{sec:few_shot}
As previously discussed and exemplified by our prompt template (Figure~\ref{fig:eval_prompt} in Appendix-\ref{app:Eval_Prompt}), our evaluation method is characterized by its high level of difficulty and complexity. In this experiment, we aimed to determine whether providing a few step-wise chain-of-thought (CoT) examples could improve model performance in terms of format adherence and reasoning quality. The results, as presented in Table~\ref{tab:few_shot_results} in Appendix, do not show a consistent pattern as the number of shots increases. While smaller language models like Gemma-2B exhibit performance improvements with additional shots, the performance of larger language models tends to fluctuate with an increasing number of shots. We hypothesize that for our complex tasks, the lengthy few-shot demonstrations may act more as a hindrance, providing distracting information rather than aiding in format adherence and reasoning. Our empirical findings suggest that a one-shot demonstration strikes the optimal balance between providing guidance and minimizing distraction. This supports our decision to focus on zero-shot versus one-shot comparisons in our primary experiments, as detailed in Table~\ref{tab:main_table}.

\subsection{Self Refine Prompting}
As suggested by \cite{huang2023large}, large language models typically cannot perform self-correction without external ground truth feedback. To explore whether this phenomenon occurs in our benchmark, we adopted a similar setting by prompting the language model to verify its own answer across a three-round interaction sequence: query, examine, and refine. Our prompting template, detailed in Figure~\ref{fig:self_refine_prompt} in Appendix~\ref{app:Eval_Prompt}, is minimalistic and designed solely to encourage the model to self-examine.

The results of this self-refinement process are recorded in Table~\ref{tab:cot_vs_self_refine}. Notably, models smaller than Llama3-70B exhibit performance degradation with self-refinement, while larger models, such as GPT-4, show marginal benefits from the process. Conversely, from Llama3-8B to Llama3-70B, despite a significant portion of correct predictions shifting to incorrect ones, as previously reported by \cite{huang2023large}, our benchmark shows an increasing trend of incorrect predictions shifting to correct ones as model size increases. This shift results in the significant performance improvements observed in models like Llama3-70B.

To understand the disproportionate improvement observed in the 70B model, we analyzed performance breakdown at the task level. These results are visualized and discussed in Figure~\ref{fig:self_refine_breakdown} of Appendix~\ref{sec:self_refine}. In short, we believe the lack of consistency does not necessarily indicate a more robust or advanced reasoning ability, despite the increase of the evaluation results.

\subsection{Solution Correctness Prior}
\begin{table}
\begin{minipage}{0.45\textwidth}
\setlength\extrarowheight{1pt}
\renewcommand{\arraystretch}{0.75}
\setlength{\tabcolsep}{2pt}
\caption{Comparison of average accuracy in identifying the first error step and the corresponding error reason, with and without prior knowledge of the solutions' correctness.}
\centering
\resizebox{\linewidth}{!}{%
    \begin{tabular}{lcccc}
    \toprule
    \multirow{2}{*}{\textbf{Model}} & \multicolumn{2}{c}{\textbf{Detection Acc.}} & \multicolumn{2}{c}{\textbf{Reason Acc.}} \\[-1pt]
    \cmidrule(lr){2-3} \cmidrule(lr){4-5}
    & w/o & with & w/o & with \\[-1pt]
    \midrule
    Gemma-2B & 0.3 & 0.1 & 0.1 & 0.0 \\
    Llama3-8B & 15.5 & 26.4 & 6.6 & 11.9 \\
    Llama3-70B & 14.5 & 34.6 & 9.1 & 25.7 \\
    GPT-4-Turbo & 40.9 & 41.6 & 37.9 & 38.0 \\
    \bottomrule
    \end{tabular}
}
\label{tab:PriorAccuracy}
\end{minipage}
\hfill
\begin{minipage}{0.48\textwidth}
\caption{Comparison of prompting methods: MR-Scores achieved by zero-shot step-wise CoT and Self-Refine technique.}
\centering
\setlength\extrarowheight{1pt}
\renewcommand{\arraystretch}{0.85}
\setlength{\tabcolsep}{6pt}
    \resizebox{\linewidth}{!}{%
    \begin{tabular}{lcc}
        \toprule
        \textbf{Model}     & \textbf{0-shot CoT}   & \textbf{Self-Refine} \\
        \midrule
        Gemma-2B & 0.1 & 0.2 \\
        Llama3-8B & 11.7  & 11.3 \\
        Llama3-70B & 17.7 & 27.5 \\
        GPT-4-Turbo  & 43.2 & 45.5 \\
        \bottomrule
    \end{tabular}\
    }
\label{tab:cot_vs_self_refine}
\end{minipage}
\end{table}

To verify the influence of external ground truth signals, we sampled 100 incorrect solutions from each subject respectively as our test set. By observing the same set of language models under a zero-shot CoT setting, we aim to determine whether the knowledge of the solution's incorrectness enhances their ability to identify the first error step and the reason for the error.

The results in Table~\ref{tab:PriorAccuracy} illustrate that the benefits of knowing the solution correctness prior generally increase with the model's competence but begin to plateau at the level of sophisticated models like GPT-4. Specifically, the Gemma-2b model struggles significantly in our benchmark, showing nearly zero performance due to its limited ability to follow formats and comprehend complex tasks. Consequently, having the solution correctness prior does not improve its performance metrics. In contrast, models with moderate capabilities benefit substantially from this prior knowledge, which aids in accurately locating the first error step and elucidating the error reason. However, as model capabilities improve, the incremental benefits of this prior knowledge quickly diminish. For instance, GPT-4 shows only a marginal improvement in identifying the first error step and an almost negligible impact on error reason analysis when provided with the prior.

\section{Conclusion}
\label{sec:conclusion}

This paper highlights the importance of evaluating the reasoning capabilities of LLMs with process-oriented design and presents a comprehensive benchmark called \data that addresses the limitations of existing evaluation methodologies. \data consists of questions from a diverse range of subjects and incorporates a meta-reasoning paradigm, where LLMs act as teachers to evaluate the reasoning process. Our evaluation of a diverse suite of LLMs on \data reveals several key limitations and weaknesses. Many models struggle with identifying and correcting errors within reasoning chains, demonstrating difficulty in performing system-2 style thinking—such as scrutinizing assumptions, calculations, and intermediate steps. Furthermore, even state-of-the-art models often fail to maintain consistency across reasoning paradigms, exposing gaps in their generalization abilities. Additionally, our findings emphasize the importance of searching and reflecting on the solution space during inference. Models like the o1 series showcase the potential of scaling test-time computation, where frequent retrospection and iterative search through multiple solution paths significantly enhance reasoning performance. Nevertheless, improving LLMs’ reasoning abilities on complex and nuanced tasks remains an open research question, and we encourage future work to develop upon \data. 



\section{Acknowledgement}
This work was supported in part by the Research Grants Council under the Areas of Excellence scheme grant AoE/E-601/22-R.

\clearpage
\bibliographystyle{plainnat}
\bibliography{custom}

\begin{thebibliography}{80}
\providecommand{\natexlab}[1]{#1}
\providecommand{\url}[1]{\texttt{#1}}
\expandafter\ifx\csname urlstyle\endcsname\relax
  \providecommand{\doi}[1]{doi: #1}\else
  \providecommand{\doi}{doi: \begingroup \urlstyle{rm}\Url}\fi

\bibitem[Abdin et~al.(2024)Abdin, Jacobs, Awan, Aneja, Awadallah, Awadalla,
  Bach, Bahree, Bakhtiari, Behl, et~al.]{abdin2024phi}
Marah Abdin, Sam~Ade Jacobs, Ammar~Ahmad Awan, Jyoti Aneja, Ahmed Awadallah,
  Hany Awadalla, Nguyen Bach, Amit Bahree, Arash Bakhtiari, Harkirat Behl,
  et~al.
\newblock Phi-3 technical report: A highly capable language model locally on
  your phone.
\newblock \emph{arXiv preprint arXiv:2404.14219}, 2024.

\bibitem[AI(2024{\natexlab{a}})]{MoonshotKimi}
Moonshot AI.
\newblock Moonshot ai, 2024{\natexlab{a}}.
\newblock URL \url{https://www.moonshot.cn/}.

\bibitem[AI(2024{\natexlab{b}})]{ChatGLM}
Zhipu AI.
\newblock Welcome to glm-4, 2024{\natexlab{b}}.
\newblock URL \url{https://en.chatglm.cn/}.

\bibitem[Amini et~al.(2019)Amini, Gabriel, Lin, Koncel{-}Kedziorski, Choi, and
  Hajishirzi]{AminiGLKCH19}
Aida Amini, Saadia Gabriel, Shanchuan Lin, Rik Koncel{-}Kedziorski, Yejin Choi,
  and Hannaneh Hajishirzi.
\newblock Mathqa: Towards interpretable math word problem solving with
  operation-based formalisms.
\newblock In Jill Burstein, Christy Doran, and Thamar Solorio, editors,
  \emph{Proceedings of the 2019 Conference of the North American Chapter of the
  Association for Computational Linguistics: Human Language Technologies,
  {NAACL-HLT} 2019, Minneapolis, MN, USA, June 2-7, 2019, Volume 1 (Long and
  Short Papers)}, pages 2357--2367. Association for Computational Linguistics,
  2019.
\newblock \doi{10.18653/V1/N19-1245}.
\newblock URL \url{https://doi.org/10.18653/v1/n19-1245}.

\bibitem[Anthropic(2024{\natexlab{a}})]{Claude2}
Anthropic.
\newblock Claude 2, 2024{\natexlab{a}}.
\newblock URL \url{https://www.anthropic.com/news/claude-2}.

\bibitem[Anthropic(2024{\natexlab{b}})]{Claude3}
Anthropic.
\newblock Introducing the next generation of claude, 2024{\natexlab{b}}.
\newblock URL \url{https://www.anthropic.com/news/claude-3-family}.

\bibitem[Bai et~al.(2023)Bai, Bai, Chu, Cui, Dang, Deng, Fan, Ge, Han, Huang,
  et~al.]{bai2023qwen}
Jinze Bai, Shuai Bai, Yunfei Chu, Zeyu Cui, Kai Dang, Xiaodong Deng, Yang Fan,
  Wenbin Ge, Yu~Han, Fei Huang, et~al.
\newblock Qwen technical report.
\newblock \emph{arXiv preprint arXiv:2309.16609}, 2023.

\bibitem[Bai et~al.(2022)Bai, Kadavath, Kundu, Askell, Kernion, Jones, Chen,
  Goldie, Mirhoseini, McKinnon, Chen, Olsson, Olah, Hernandez, Drain, Ganguli,
  Li, Tran{-}Johnson, Perez, Kerr, Mueller, Ladish, Landau, Ndousse, Lukosiute,
  Lovitt, Sellitto, Elhage, Schiefer, Mercado, DasSarma, Lasenby, Larson,
  Ringer, Johnston, Kravec, Showk, Fort, Lanham, Telleen{-}Lawton, Conerly,
  Henighan, Hume, Bowman, Hatfield{-}Dodds, Mann, Amodei, Joseph, McCandlish,
  Brown, and Kaplan]{ConstitutionalAI}
Yuntao Bai, Saurav Kadavath, Sandipan Kundu, Amanda Askell, Jackson Kernion,
  Andy Jones, Anna Chen, Anna Goldie, Azalia Mirhoseini, Cameron McKinnon,
  Carol Chen, Catherine Olsson, Christopher Olah, Danny Hernandez, Dawn Drain,
  Deep Ganguli, Dustin Li, Eli Tran{-}Johnson, Ethan Perez, Jamie Kerr, Jared
  Mueller, Jeffrey Ladish, Joshua Landau, Kamal Ndousse, Kamile Lukosiute,
  Liane Lovitt, Michael Sellitto, Nelson Elhage, Nicholas Schiefer,
  Noem{\'{\i}} Mercado, Nova DasSarma, Robert Lasenby, Robin Larson, Sam
  Ringer, Scott Johnston, Shauna Kravec, Sheer~El Showk, Stanislav Fort, Tamera
  Lanham, Timothy Telleen{-}Lawton, Tom Conerly, Tom Henighan, Tristan Hume,
  Samuel~R. Bowman, Zac Hatfield{-}Dodds, Ben Mann, Dario Amodei, Nicholas
  Joseph, Sam McCandlish, Tom Brown, and Jared Kaplan.
\newblock Constitutional {AI:} harmlessness from {AI} feedback.
\newblock \emph{CoRR}, abs/2212.08073, 2022.
\newblock \doi{10.48550/ARXIV.2212.08073}.
\newblock URL \url{https://doi.org/10.48550/arXiv.2212.08073}.

\bibitem[Bengio(2020)]{bengio2020system2}
Yoshua Bengio.
\newblock Deep learning for system 2 processing.
\newblock Presentation at the AAAI-20 Turing Award Winners 2018 Special Event,
  February 9 2020.

\bibitem[Bi et~al.(2024)Bi, Chen, Chen, Chen, Dai, Deng, Ding, Dong, Du, Fu,
  Gao, Gao, Gao, Ge, Guan, Guo, Guo, Hao, Hao, He, Hu, Huang, Li, Li, Li, Li,
  Li, Liang, Lin, Liu, Liu, Liu, Liu, Liu, Liu, Lu, Lu, Luo, Ma, Nie, Pei,
  Piao, Qiu, Qu, Ren, Ren, Ruan, Sha, Shao, Song, Su, Sun, Sun, Tang, Wang,
  Wang, Wang, Wang, Wang, Wu, Wu, Xie, Xie, Xie, Xiong, Xu, Xu, Xu, Yang, You,
  Yu, Yu, Zhang, Zhang, Zhang, Zhang, Zhang, Zhang, Zhang, Zhang, Zhao, Zhao,
  Zhou, Zhou, Zhu, and Zou]{DBLP:journals/corr/abs-2401-02954}
Xiao Bi, Deli Chen, Guanting Chen, Shanhuang Chen, Damai Dai, Chengqi Deng,
  Honghui Ding, Kai Dong, Qiushi Du, Zhe Fu, Huazuo Gao, Kaige Gao, Wenjun Gao,
  Ruiqi Ge, Kang Guan, Daya Guo, Jianzhong Guo, Guangbo Hao, Zhewen Hao, Ying
  He, Wenjie Hu, Panpan Huang, Erhang Li, Guowei Li, Jiashi Li, Yao Li, Y.~K.
  Li, Wenfeng Liang, Fangyun Lin, Alex~X. Liu, Bo~Liu, Wen Liu, Xiaodong Liu,
  Xin Liu, Yiyuan Liu, Haoyu Lu, Shanghao Lu, Fuli Luo, Shirong Ma, Xiaotao
  Nie, Tian Pei, Yishi Piao, Junjie Qiu, Hui Qu, Tongzheng Ren, Zehui Ren,
  Chong Ruan, Zhangli Sha, Zhihong Shao, Junxiao Song, Xuecheng Su, Jingxiang
  Sun, Yaofeng Sun, Minghui Tang, Bingxuan Wang, Peiyi Wang, Shiyu Wang, Yaohui
  Wang, Yongji Wang, Tong Wu, Y.~Wu, Xin Xie, Zhenda Xie, Ziwei Xie, Yiliang
  Xiong, Hanwei Xu, R.~X. Xu, Yanhong Xu, Dejian Yang, Yuxiang You, Shuiping
  Yu, Xingkai Yu, B.~Zhang, Haowei Zhang, Lecong Zhang, Liyue Zhang, Mingchuan
  Zhang, Minghua Zhang, Wentao Zhang, Yichao Zhang, Chenggang Zhao, Yao Zhao,
  Shangyan Zhou, Shunfeng Zhou, Qihao Zhu, and Yuheng Zou.
\newblock Deepseek {LLM:} scaling open-source language models with longtermism.
\newblock \emph{CoRR}, abs/2401.02954, 2024.
\newblock \doi{10.48550/ARXIV.2401.02954}.
\newblock URL \url{https://doi.org/10.48550/arXiv.2401.02954}.

\bibitem[Brown et~al.(2020)Brown, Mann, Ryder, Subbiah, Kaplan, Dhariwal,
  Neelakantan, Shyam, Sastry, Askell, Agarwal, Herbert{-}Voss, Krueger,
  Henighan, Child, Ramesh, Ziegler, Wu, Winter, Hesse, Chen, Sigler, Litwin,
  Gray, Chess, Clark, Berner, McCandlish, Radford, Sutskever, and Amodei]{gpt3}
Tom~B. Brown, Benjamin Mann, Nick Ryder, Melanie Subbiah, Jared Kaplan,
  Prafulla Dhariwal, Arvind Neelakantan, Pranav Shyam, Girish Sastry, Amanda
  Askell, Sandhini Agarwal, Ariel Herbert{-}Voss, Gretchen Krueger, Tom
  Henighan, Rewon Child, Aditya Ramesh, Daniel~M. Ziegler, Jeffrey Wu, Clemens
  Winter, Christopher Hesse, Mark Chen, Eric Sigler, Mateusz Litwin, Scott
  Gray, Benjamin Chess, Jack Clark, Christopher Berner, Sam McCandlish, Alec
  Radford, Ilya Sutskever, and Dario Amodei.
\newblock Language models are few-shot learners.
\newblock \emph{CoRR}, abs/2005.14165, 2020.
\newblock URL \url{https://arxiv.org/abs/2005.14165}.

\bibitem[Bytedance(2024)]{Doubao}
Bytedance.
\newblock Doubao team - crafting the industry's most advanced llms., 2024.
\newblock URL \url{https://www.doubao.com/chat/}.

\bibitem[Chung et~al.(2022)Chung, Hou, Longpre, Zoph, Tay, Fedus, Li, Wang,
  Dehghani, Brahma, Webson, Gu, Dai, Suzgun, Chen, Chowdhery, Narang, Mishra,
  Yu, Zhao, Huang, Dai, Yu, Petrov, Chi, Dean, Devlin, Roberts, Zhou, Le, and
  Wei]{Chung2022}
Hyung~Won Chung, Le~Hou, Shayne Longpre, Barret Zoph, Yi~Tay, William Fedus,
  Eric Li, Xuezhi Wang, Mostafa Dehghani, Siddhartha Brahma, Albert Webson,
  Shixiang~Shane Gu, Zhuyun Dai, Mirac Suzgun, Xinyun Chen, Aakanksha
  Chowdhery, Sharan Narang, Gaurav Mishra, Adams Yu, Vincent~Y. Zhao, Yanping
  Huang, Andrew~M. Dai, Hongkun Yu, Slav Petrov, Ed~H. Chi, Jeff Dean, Jacob
  Devlin, Adam Roberts, Denny Zhou, Quoc~V. Le, and Jason Wei.
\newblock Scaling instruction-finetuned language models.
\newblock \emph{CoRR}, abs/2210.11416, 2022.
\newblock \doi{10.48550/ARXIV.2210.11416}.
\newblock URL \url{https://doi.org/10.48550/arXiv.2210.11416}.

\bibitem[Clark et~al.(2018)Clark, Cowhey, Etzioni, Khot, Sabharwal, Schoenick,
  and Tafjord]{ClarkARC2018}
Peter Clark, Isaac Cowhey, Oren Etzioni, Tushar Khot, Ashish Sabharwal, Carissa
  Schoenick, and Oyvind Tafjord.
\newblock Think you have solved question answering? try arc, the {AI2}
  reasoning challenge.
\newblock \emph{CoRR}, abs/1803.05457, 2018.
\newblock URL \url{http://arxiv.org/abs/1803.05457}.

\bibitem[Clark et~al.(2020)Clark, Tafjord, and Richardson]{ClarkTR20}
Peter Clark, Oyvind Tafjord, and Kyle Richardson.
\newblock Transformers as soft reasoners over language.
\newblock In Christian Bessiere, editor, \emph{Proceedings of the Twenty-Ninth
  International Joint Conference on Artificial Intelligence, {IJCAI} 2020},
  pages 3882--3890. ijcai.org, 2020.
\newblock \doi{10.24963/IJCAI.2020/537}.
\newblock URL \url{https://doi.org/10.24963/ijcai.2020/537}.

\bibitem[Cobbe et~al.(2021)Cobbe, Kosaraju, Bavarian, Chen, Jun, Kaiser,
  Plappert, Tworek, Hilton, Nakano, Hesse, and Schulman]{CobbeGSM8K2021}
Karl Cobbe, Vineet Kosaraju, Mohammad Bavarian, Mark Chen, Heewoo Jun, Lukasz
  Kaiser, Matthias Plappert, Jerry Tworek, Jacob Hilton, Reiichiro Nakano,
  Christopher Hesse, and John Schulman.
\newblock Training verifiers to solve math word problems.
\newblock \emph{CoRR}, abs/2110.14168, 2021.
\newblock URL \url{https://arxiv.org/abs/2110.14168}.

\bibitem[Dai et~al.(2024)Dai, Lu, Feng, Ruan, Cheng, Tan, and Guo]{Dai2024mhpp}
Jianbo Dai, Jianqiao Lu, Yunlong Feng, Rongju Ruan, Ming Cheng, Haochen Tan,
  and Zhijiang Guo.
\newblock Mhpp: Exploring the capabilities and limitations of language models
  beyond basic code generation.
\newblock \emph{arXiv preprint arXiv:2405.11430}, 2024.

\bibitem[Dalvi et~al.(2021)Dalvi, Jansen, Tafjord, Xie, Smith, Pipatanangkura,
  and Clark]{DalviJTXSPC21}
Bhavana Dalvi, Peter Jansen, Oyvind Tafjord, Zhengnan Xie, Hannah Smith,
  Leighanna Pipatanangkura, and Peter Clark.
\newblock Explaining answers with entailment trees.
\newblock In Marie{-}Francine Moens, Xuanjing Huang, Lucia Specia, and
  Scott~Wen{-}tau Yih, editors, \emph{Proceedings of the 2021 Conference on
  Empirical Methods in Natural Language Processing, {EMNLP} 2021, Virtual Event
  / Punta Cana, Dominican Republic, 7-11 November, 2021}, pages 7358--7370.
  Association for Computational Linguistics, 2021.
\newblock \doi{10.18653/V1/2021.EMNLP-MAIN.585}.
\newblock URL \url{https://doi.org/10.18653/v1/2021.emnlp-main.585}.

\bibitem[Fagin and Halpern(1994)]{FaginH94}
Ronald Fagin and Joseph~Y. Halpern.
\newblock Reasoning about knowledge and probability.
\newblock \emph{J. {ACM}}, 41\penalty0 (2):\penalty0 340--367, 1994.
\newblock \doi{10.1145/174652.174658}.
\newblock URL \url{https://doi.org/10.1145/174652.174658}.

\bibitem[Fernandes et~al.(2023)Fernandes, Madaan, Liu, Farinhas, Martins,
  Bertsch, de~Souza, Zhou, Wu, Neubig, and Martins]{fernandes2023bridging}
Patrick Fernandes, Aman Madaan, Emmy Liu, Ant{\'{o}}nio Farinhas,
  Pedro~Henrique Martins, Amanda Bertsch, Jos{\'{e}} G.~C. de~Souza, Shuyan
  Zhou, Tongshuang Wu, Graham Neubig, and Andr{\'{e}} F.~T. Martins.
\newblock Bridging the gap: {A} survey on integrating (human) feedback for
  natural language generation.
\newblock \emph{CoRR}, abs/2305.00955, 2023.
\newblock \doi{10.48550/ARXIV.2305.00955}.
\newblock URL \url{https://doi.org/10.48550/arXiv.2305.00955}.

\bibitem[Gao et~al.(2023)Gao, Dai, Pasupat, Chen, Chaganty, Fan, Zhao, Lao,
  Lee, Juan, and Guu]{RARR}
Luyu Gao, Zhuyun Dai, Panupong Pasupat, Anthony Chen, Arun~Tejasvi Chaganty,
  Yicheng Fan, Vincent~Y. Zhao, Ni~Lao, Hongrae Lee, Da{-}Cheng Juan, and
  Kelvin Guu.
\newblock {RARR:} researching and revising what language models say, using
  language models.
\newblock In Anna Rogers, Jordan~L. Boyd{-}Graber, and Naoaki Okazaki, editors,
  \emph{Proceedings of the 61st Annual Meeting of the Association for
  Computational Linguistics (Volume 1: Long Papers), {ACL} 2023, Toronto,
  Canada, July 9-14, 2023}, pages 16477--16508. Association for Computational
  Linguistics, 2023.
\newblock \doi{10.18653/V1/2023.ACL-LONG.910}.
\newblock URL \url{https://doi.org/10.18653/v1/2023.acl-long.910}.

\bibitem[Geva et~al.(2021)Geva, Khashabi, Segal, Khot, Roth, and
  Berant]{GevaKSKRB21}
Mor Geva, Daniel Khashabi, Elad Segal, Tushar Khot, Dan Roth, and Jonathan
  Berant.
\newblock Did aristotle use a laptop? {A} question answering benchmark with
  implicit reasoning strategies.
\newblock \emph{Trans. Assoc. Comput. Linguistics}, 9:\penalty0 346--361, 2021.
\newblock \doi{10.1162/TACL\_A\_00370}.
\newblock URL \url{https://doi.org/10.1162/tacl\_a\_00370}.

\bibitem[Golovneva et~al.(2023)Golovneva, Chen, Poff, Corredor, Zettlemoyer,
  Fazel{-}Zarandi, and Celikyilmaz]{GolovnevaCPCZFC23}
Olga Golovneva, Moya Chen, Spencer Poff, Martin Corredor, Luke Zettlemoyer,
  Maryam Fazel{-}Zarandi, and Asli Celikyilmaz.
\newblock {ROSCOE:} {A} suite of metrics for scoring step-by-step reasoning.
\newblock In \emph{The Eleventh International Conference on Learning
  Representations, {ICLR} 2023, Kigali, Rwanda, May 1-5, 2023}. OpenReview.net,
  2023.
\newblock URL \url{https://openreview.net/pdf?id=xYlJRpzZtsY}.

\bibitem[Gou et~al.(2023)Gou, Shao, Gong, Shen, Yang, Duan, and Chen]{Critic}
Zhibin Gou, Zhihong Shao, Yeyun Gong, Yelong Shen, Yujiu Yang, Nan Duan, and
  Weizhu Chen.
\newblock {CRITIC:} large language models can self-correct with
  tool-interactive critiquing.
\newblock \emph{CoRR}, abs/2305.11738, 2023.
\newblock \doi{10.48550/ARXIV.2305.11738}.
\newblock URL \url{https://doi.org/10.48550/arXiv.2305.11738}.

\bibitem[Gu et~al.(2024)Gu, Rozi{\`{e}}re, Leather, Solar{-}Lezama, Synnaeve,
  and Wang]{GuRLSS024}
Alex Gu, Baptiste Rozi{\`{e}}re, Hugh~James Leather, Armando Solar{-}Lezama,
  Gabriel Synnaeve, and Sida Wang.
\newblock Cruxeval: {A} benchmark for code reasoning, understanding and
  execution.
\newblock In \emph{Forty-first International Conference on Machine Learning,
  {ICML} 2024, Vienna, Austria, July 21-27, 2024}. OpenReview.net, 2024.
\newblock URL \url{https://openreview.net/forum?id=Ffpg52swvg}.

\bibitem[Gunasekar et~al.(2023)Gunasekar, Zhang, Aneja, Mendes, Giorno, Gopi,
  Javaheripi, Kauffmann, de~Rosa, Saarikivi, Salim, Shah, Behl, Wang, Bubeck,
  Eldan, Kalai, Lee, and Li]{Gunasekar2023}
Suriya Gunasekar, Yi~Zhang, Jyoti Aneja, Caio C{\'{e}}sar~Teodoro Mendes,
  Allie~Del Giorno, Sivakanth Gopi, Mojan Javaheripi, Piero Kauffmann, Gustavo
  de~Rosa, Olli Saarikivi, Adil Salim, Shital Shah, Harkirat~Singh Behl, Xin
  Wang, S{\'{e}}bastien Bubeck, Ronen Eldan, Adam~Tauman Kalai, Yin~Tat Lee,
  and Yuanzhi Li.
\newblock Textbooks are all you need.
\newblock \emph{CoRR}, abs/2306.11644, 2023.
\newblock \doi{10.48550/ARXIV.2306.11644}.
\newblock URL \url{https://doi.org/10.48550/arXiv.2306.11644}.

\bibitem[Hendrycks et~al.(2021{\natexlab{a}})Hendrycks, Burns, Basart, Zou,
  Mazeika, Song, and Steinhardt]{hendrycks2020measuring}
Dan Hendrycks, Collin Burns, Steven Basart, Andy Zou, Mantas Mazeika, Dawn
  Song, and Jacob Steinhardt.
\newblock Measuring massive multitask language understanding.
\newblock In \emph{9th International Conference on Learning Representations,
  {ICLR} 2021, Virtual Event, Austria, May 3-7, 2021}. OpenReview.net,
  2021{\natexlab{a}}.
\newblock URL \url{https://openreview.net/forum?id=d7KBjmI3GmQ}.

\bibitem[Hendrycks et~al.(2021{\natexlab{b}})Hendrycks, Burns, Kadavath, Arora,
  Basart, Tang, Song, and Steinhardt]{HendrycksBKABTS21}
Dan Hendrycks, Collin Burns, Saurav Kadavath, Akul Arora, Steven Basart, Eric
  Tang, Dawn Song, and Jacob Steinhardt.
\newblock Measuring mathematical problem solving with the {MATH} dataset.
\newblock In Joaquin Vanschoren and Sai{-}Kit Yeung, editors, \emph{Proceedings
  of the Neural Information Processing Systems Track on Datasets and Benchmarks
  1, NeurIPS Datasets and Benchmarks 2021, December 2021, virtual},
  2021{\natexlab{b}}.
\newblock URL
  \url{https://datasets-benchmarks-proceedings.neurips.cc/paper/2021/hash/be83ab3ecd0db773eb2dc1b0a17836a1-Abstract-round2.html}.

\bibitem[Huang et~al.(2024)Huang, Dai, Weng, Wu, Qing, Zhang, Cui, and
  Guo]{Huang2024Soap}
Dong Huang, Jianbo Dai, Han Weng, Puzhen Wu, Yuhao Qing, Jie~M. Zhang, Heming
  Cui, and Zhijiang Guo.
\newblock {SOAP:} enhancing efficiency of generated code via self-optimization.
\newblock \emph{CoRR}, abs/2405.15189, 2024.
\newblock \doi{10.48550/ARXIV.2405.15189}.
\newblock URL \url{https://doi.org/10.48550/arXiv.2405.15189}.

\bibitem[Huang and Chang(2023)]{HuangSurvey2023}
Jie Huang and Kevin~Chen{-}Chuan Chang.
\newblock Towards reasoning in large language models: {A} survey.
\newblock In Anna Rogers, Jordan~L. Boyd{-}Graber, and Naoaki Okazaki, editors,
  \emph{Findings of the Association for Computational Linguistics: {ACL} 2023,
  Toronto, Canada, July 9-14, 2023}, pages 1049--1065. Association for
  Computational Linguistics, 2023.
\newblock \doi{10.18653/V1/2023.FINDINGS-ACL.67}.
\newblock URL \url{https://doi.org/10.18653/v1/2023.findings-acl.67}.

\bibitem[Huang et~al.(2023)Huang, Chen, Mishra, Zheng, Yu, Song, and
  Zhou]{huang2023large}
Jie Huang, Xinyun Chen, Swaroop Mishra, Huaixiu~Steven Zheng, Adams~Wei Yu,
  Xinying Song, and Denny Zhou.
\newblock Large language models cannot self-correct reasoning yet.
\newblock \emph{arXiv preprint arXiv:2310.01798}, 2023.

\bibitem[Jiang et~al.(2023)Jiang, Sablayrolles, Mensch, Bamford, Chaplot,
  de~Las~Casas, Bressand, Lengyel, Lample, Saulnier, Lavaud, Lachaux, Stock,
  Scao, Lavril, Wang, Lacroix, and Sayed]{Mistral23}
Albert~Q. Jiang, Alexandre Sablayrolles, Arthur Mensch, Chris Bamford,
  Devendra~Singh Chaplot, Diego de~Las~Casas, Florian Bressand, Gianna Lengyel,
  Guillaume Lample, Lucile Saulnier, L{\'{e}}lio~Renard Lavaud, Marie{-}Anne
  Lachaux, Pierre Stock, Teven~Le Scao, Thibaut Lavril, Thomas Wang,
  Timoth{\'{e}}e Lacroix, and William~El Sayed.
\newblock Mistral 7b.
\newblock \emph{CoRR}, abs/2310.06825, 2023.
\newblock \doi{10.48550/ARXIV.2310.06825}.
\newblock URL \url{https://doi.org/10.48550/arXiv.2310.06825}.

\bibitem[Jiang et~al.(2024)Jiang, Sablayrolles, Roux, Mensch, Savary, Bamford,
  Chaplot, de~Las~Casas, Hanna, Bressand, Lengyel, Bour, Lample, Lavaud,
  Saulnier, Lachaux, Stock, Subramanian, Yang, Antoniak, Scao, Gervet, Lavril,
  Wang, Lacroix, and Sayed]{Mixtral24}
Albert~Q. Jiang, Alexandre Sablayrolles, Antoine Roux, Arthur Mensch, Blanche
  Savary, Chris Bamford, Devendra~Singh Chaplot, Diego de~Las~Casas, Emma~Bou
  Hanna, Florian Bressand, Gianna Lengyel, Guillaume Bour, Guillaume Lample,
  L{\'{e}}lio~Renard Lavaud, Lucile Saulnier, Marie{-}Anne Lachaux, Pierre
  Stock, Sandeep Subramanian, Sophia Yang, Szymon Antoniak, Teven~Le Scao,
  Th{\'{e}}ophile Gervet, Thibaut Lavril, Thomas Wang, Timoth{\'{e}}e Lacroix,
  and William~El Sayed.
\newblock Mixtral of experts.
\newblock \emph{CoRR}, abs/2401.04088, 2024.
\newblock \doi{10.48550/ARXIV.2401.04088}.
\newblock URL \url{https://doi.org/10.48550/arXiv.2401.04088}.

\bibitem[Jung et~al.(2022)Jung, Qin, Welleck, Brahman, Bhagavatula, Bras, and
  Choi]{MaieuticPrompt}
Jaehun Jung, Lianhui Qin, Sean Welleck, Faeze Brahman, Chandra Bhagavatula,
  Ronan~Le Bras, and Yejin Choi.
\newblock Maieutic prompting: Logically consistent reasoning with recursive
  explanations.
\newblock In Yoav Goldberg, Zornitsa Kozareva, and Yue Zhang, editors,
  \emph{Proceedings of the 2022 Conference on Empirical Methods in Natural
  Language Processing, {EMNLP} 2022, Abu Dhabi, United Arab Emirates, December
  7-11, 2022}, pages 1266--1279. Association for Computational Linguistics,
  2022.
\newblock \doi{10.18653/V1/2022.EMNLP-MAIN.82}.
\newblock URL \url{https://doi.org/10.18653/v1/2022.emnlp-main.82}.

\bibitem[Kahneman(2011)]{kahneman2011thinking}
Daniel Kahneman.
\newblock Thinking, fast and slow.
\newblock \emph{Farrar, Straus and Giroux}, 2011.

\bibitem[Kojima et~al.(2022)Kojima, Gu, Reid, Matsuo, and
  Iwasawa]{KojimaGRMI22}
Takeshi Kojima, Shixiang~Shane Gu, Machel Reid, Yutaka Matsuo, and Yusuke
  Iwasawa.
\newblock Large language models are zero-shot reasoners.
\newblock In Sanmi Koyejo, S.~Mohamed, A.~Agarwal, Danielle Belgrave, K.~Cho,
  and A.~Oh, editors, \emph{Advances in Neural Information Processing Systems
  35: Annual Conference on Neural Information Processing Systems 2022, NeurIPS
  2022, New Orleans, LA, USA, November 28 - December 9, 2022}, 2022.
\newblock URL
  \url{http://papers.nips.cc/paper\_files/paper/2022/hash/8bb0d291acd4acf06ef112099c16f326-Abstract-Conference.html}.

\bibitem[Koncel{-}Kedziorski et~al.(2016)Koncel{-}Kedziorski, Roy, Amini,
  Kushman, and Hajishirzi]{Koncel-Kedziorski16}
Rik Koncel{-}Kedziorski, Subhro Roy, Aida Amini, Nate Kushman, and Hannaneh
  Hajishirzi.
\newblock {MAWPS:} {A} math word problem repository.
\newblock In Kevin Knight, Ani Nenkova, and Owen Rambow, editors, \emph{{NAACL}
  {HLT} 2016, The 2016 Conference of the North American Chapter of the
  Association for Computational Linguistics: Human Language Technologies, San
  Diego California, USA, June 12-17, 2016}, pages 1152--1157. The Association
  for Computational Linguistics, 2016.
\newblock \doi{10.18653/V1/N16-1136}.
\newblock URL \url{https://doi.org/10.18653/v1/n16-1136}.

\bibitem[Lightman et~al.(2023)Lightman, Kosaraju, Burda, Edwards, Baker, Lee,
  Leike, Schulman, Sutskever, and Cobbe]{LightmanVerify2023}
Hunter Lightman, Vineet Kosaraju, Yura Burda, Harrison Edwards, Bowen Baker,
  Teddy Lee, Jan Leike, John Schulman, Ilya Sutskever, and Karl Cobbe.
\newblock Let's verify step by step.
\newblock \emph{CoRR}, abs/2305.20050, 2023.
\newblock \doi{10.48550/ARXIV.2305.20050}.
\newblock URL \url{https://doi.org/10.48550/arXiv.2305.20050}.

\bibitem[LingYiWanWu(2024)]{LingYiWanWu}
LingYiWanWu.
\newblock Yi ai, 2024.
\newblock URL \url{https://platform.lingyiwanwu.com/}.

\bibitem[Liu et~al.(2020)Liu, Cui, Liu, Huang, Wang, and Zhang]{liu2020logiqa}
Jian Liu, Leyang Cui, Hanmeng Liu, Dandan Huang, Yile Wang, and Yue Zhang.
\newblock Logiqa: {A} challenge dataset for machine reading comprehension with
  logical reasoning.
\newblock In Christian Bessiere, editor, \emph{Proceedings of the Twenty-Ninth
  International Joint Conference on Artificial Intelligence, {IJCAI} 2020},
  pages 3622--3628. ijcai.org, 2020.
\newblock \doi{10.24963/IJCAI.2020/501}.
\newblock URL \url{https://doi.org/10.24963/ijcai.2020/501}.

\bibitem[Liu et~al.(2024{\natexlab{a}})Liu, Guo, Liang, Shareghi, Vuli{\'c},
  and Collier]{liu2024measuring}
Yinhong Liu, Zhijiang Guo, Tianya Liang, Ehsan Shareghi, Ivan Vuli{\'c}, and
  Nigel Collier.
\newblock Measuring, evaluating and improving logical consistency in large
  language models.
\newblock \emph{arXiv preprint arXiv:2410.02205}, 2024{\natexlab{a}}.

\bibitem[Liu et~al.(2024{\natexlab{b}})Liu, Zhou, Guo, Shareghi, Vulić,
  Korhonen, and Collier]{liu2024aligning}
Yinhong Liu, Han Zhou, Zhijiang Guo, Ehsan Shareghi, Ivan Vulić, Anna
  Korhonen, and Nigel Collier.
\newblock Aligning with human judgement: The role of pairwise preference in
  large language model evaluators, 2024{\natexlab{b}}.

\bibitem[Liu et~al.(2024{\natexlab{c}})Liu, Moosavi, and Lin]{liu2024llms}
Yiqi Liu, Nafise~Sadat Moosavi, and Chenghua Lin.
\newblock Llms as narcissistic evaluators: When ego inflates evaluation scores,
  2024{\natexlab{c}}.

\bibitem[Lu et~al.(2024{\natexlab{a}})Lu, Dou, Wang, Cao, Dai, Wan, Huang, and
  Guo]{Lu2024Auto}
Jianqiao Lu, Zhiyang Dou, Hongru Wang, Zeyu Cao, Jianbo Dai, Yingjia Wan, Yinya
  Huang, and Zhijiang Guo.
\newblock Autocv: Empowering reasoning with automated process labeling via
  confidence variation.
\newblock \emph{CoRR}, abs/2405.16802, 2024{\natexlab{a}}.
\newblock \doi{10.48550/ARXIV.2405.16802}.
\newblock URL \url{https://doi.org/10.48550/arXiv.2405.16802}.

\bibitem[Lu et~al.(2024{\natexlab{b}})Lu, Liu, Wan, Huang, Wang, Yang, Tang,
  and Guo]{Lu2024PDA}
Jianqiao Lu, Zhengying Liu, Yingjia Wan, Yinya Huang, Haiming Wang, Zhicheng
  Yang, Jing Tang, and Zhijiang Guo.
\newblock Process-driven autoformalization in lean 4.
\newblock \emph{CoRR}, abs/2406.01940, 2024{\natexlab{b}}.
\newblock \doi{10.48550/ARXIV.2406.01940}.
\newblock URL \url{https://doi.org/10.48550/arXiv.2406.01940}.

\bibitem[Matthews(1975)]{Matthews1975ComparisonOT}
Brian~W. Matthews.
\newblock Comparison of the predicted and observed secondary structure of t4
  phage lysozyme.
\newblock \emph{Biochimica et biophysica acta}, 405 2:\penalty0 442--51, 1975.
\newblock URL \url{https://api.semanticscholar.org/CorpusID:44596673}.

\bibitem[Meta(2024)]{Llama3}
Meta.
\newblock Introducing meta llama 3: The most capable openly available llm to
  date, 2024.
\newblock URL \url{https://ai.meta.com/blog/meta-llama-3/}.

\bibitem[Mishra et~al.(2022)Mishra, Finlayson, Lu, Tang, Welleck, Baral,
  Rajpurohit, Tafjord, Sabharwal, Clark, and Kalyan]{MishraFLTWBRTSC22}
Swaroop Mishra, Matthew Finlayson, Pan Lu, Leonard Tang, Sean Welleck, Chitta
  Baral, Tanmay Rajpurohit, Oyvind Tafjord, Ashish Sabharwal, Peter Clark, and
  Ashwin Kalyan.
\newblock {LILA:} {A} unified benchmark for mathematical reasoning.
\newblock In Yoav Goldberg, Zornitsa Kozareva, and Yue Zhang, editors,
  \emph{Proceedings of the 2022 Conference on Empirical Methods in Natural
  Language Processing, {EMNLP} 2022, Abu Dhabi, United Arab Emirates, December
  7-11, 2022}, pages 5807--5832. Association for Computational Linguistics,
  2022.
\newblock \doi{10.18653/V1/2022.EMNLP-MAIN.392}.
\newblock URL \url{https://doi.org/10.18653/v1/2022.emnlp-main.392}.

\bibitem[Mukherjee et~al.(2023)Mukherjee, Mitra, Jawahar, Agarwal, Palangi, and
  Awadallah]{Mukherjee2023}
Subhabrata Mukherjee, Arindam Mitra, Ganesh Jawahar, Sahaj Agarwal, Hamid
  Palangi, and Ahmed Awadallah.
\newblock Orca: Progressive learning from complex explanation traces of
  {GPT-4}.
\newblock \emph{CoRR}, abs/2306.02707, 2023.
\newblock \doi{10.48550/ARXIV.2306.02707}.
\newblock URL \url{https://doi.org/10.48550/arXiv.2306.02707}.

\bibitem[OpenAI(2023{\natexlab{a}})]{GPT35turbo}
OpenAI.
\newblock {GPT-3.5} {T}urbo, 2023{\natexlab{a}}.
\newblock URL \url{https://platform.openai.com/docs/models/gpt-3-5}.

\bibitem[OpenAI(2023{\natexlab{b}})]{GPT4}
OpenAI.
\newblock {GPT-4} {T}echnical {R}eport.
\newblock \emph{CoRR}, abs/2303.08774, 2023{\natexlab{b}}.
\newblock \doi{10.48550/arXiv.2303.08774}.
\newblock URL \url{https://doi.org/10.48550/arXiv.2303.08774}.

\bibitem[OpenAI(2024)]{o1openai}
OpenAI.
\newblock Introducing openai o1-preview, 2024.
\newblock URL \url{https://openai.com/index/introducing-openai-o1-preview/}.

\bibitem[Ouyang et~al.(2022)Ouyang, Wu, Jiang, Almeida, Wainwright, Mishkin,
  Zhang, Agarwal, Slama, Ray, Schulman, Hilton, Kelton, Miller, Simens, Askell,
  Welinder, Christiano, Leike, and Lowe]{InstructGPT}
Long Ouyang, Jeffrey Wu, Xu~Jiang, Diogo Almeida, Carroll~L. Wainwright, Pamela
  Mishkin, Chong Zhang, Sandhini Agarwal, Katarina Slama, Alex Ray, John
  Schulman, Jacob Hilton, Fraser Kelton, Luke Miller, Maddie Simens, Amanda
  Askell, Peter Welinder, Paul~F. Christiano, Jan Leike, and Ryan Lowe.
\newblock Training language models to follow instructions with human feedback.
\newblock In Sanmi Koyejo, S.~Mohamed, A.~Agarwal, Danielle Belgrave, K.~Cho,
  and A.~Oh, editors, \emph{Advances in Neural Information Processing Systems
  35: Annual Conference on Neural Information Processing Systems 2022, NeurIPS
  2022, New Orleans, LA, USA, November 28 - December 9, 2022}, 2022.
\newblock URL
  \url{http://papers.nips.cc/paper\_files/paper/2022/hash/b1efde53be364a73914f58805a001731-Abstract-Conference.html}.

\bibitem[Panickssery et~al.(2024)Panickssery, Bowman, and
  Feng]{panickssery2024llm}
Arjun Panickssery, Samuel~R. Bowman, and Shi Feng.
\newblock Llm evaluators recognize and favor their own generations, 2024.

\bibitem[Patel et~al.(2021)Patel, Bhattamishra, and Goyal]{PatelBG21}
Arkil Patel, Satwik Bhattamishra, and Navin Goyal.
\newblock Are {NLP} models really able to solve simple math word problems?
\newblock In Kristina Toutanova, Anna Rumshisky, Luke Zettlemoyer, Dilek
  Hakkani{-}T{\"{u}}r, Iz~Beltagy, Steven Bethard, Ryan Cotterell, Tanmoy
  Chakraborty, and Yichao Zhou, editors, \emph{Proceedings of the 2021
  Conference of the North American Chapter of the Association for Computational
  Linguistics: Human Language Technologies, {NAACL-HLT} 2021, Online, June
  6-11, 2021}, pages 2080--2094. Association for Computational Linguistics,
  2021.
\newblock \doi{10.18653/V1/2021.NAACL-MAIN.168}.
\newblock URL \url{https://doi.org/10.18653/v1/2021.naacl-main.168}.

\bibitem[Prasad et~al.(2023)Prasad, Saha, Zhou, and Bansal]{PrasadSZB23}
Archiki Prasad, Swarnadeep Saha, Xiang Zhou, and Mohit Bansal.
\newblock Receval: Evaluating reasoning chains via correctness and
  informativeness.
\newblock In Houda Bouamor, Juan Pino, and Kalika Bali, editors,
  \emph{Proceedings of the 2023 Conference on Empirical Methods in Natural
  Language Processing, {EMNLP} 2023, Singapore, December 6-10, 2023}, pages
  10066--10086. Association for Computational Linguistics, 2023.
\newblock \doi{10.18653/V1/2023.EMNLP-MAIN.622}.
\newblock URL \url{https://doi.org/10.18653/v1/2023.emnlp-main.622}.

\bibitem[Qiao et~al.(2023)Qiao, Ou, Zhang, Chen, Yao, Deng, Tan, Huang, and
  Chen]{QiaoO0CYDTHC23}
Shuofei Qiao, Yixin Ou, Ningyu Zhang, Xiang Chen, Yunzhi Yao, Shumin Deng,
  Chuanqi Tan, Fei Huang, and Huajun Chen.
\newblock Reasoning with language model prompting: {A} survey.
\newblock In Anna Rogers, Jordan~L. Boyd{-}Graber, and Naoaki Okazaki, editors,
  \emph{Proceedings of the 61st Annual Meeting of the Association for
  Computational Linguistics (Volume 1: Long Papers), {ACL} 2023, Toronto,
  Canada, July 9-14, 2023}, pages 5368--5393. Association for Computational
  Linguistics, 2023.
\newblock \doi{10.18653/V1/2023.ACL-LONG.294}.
\newblock URL \url{https://doi.org/10.18653/v1/2023.acl-long.294}.

\bibitem[Saparov and He(2023)]{Saparov023}
Abulhair Saparov and He~He.
\newblock Language models are greedy reasoners: {A} systematic formal analysis
  of chain-of-thought.
\newblock In \emph{The Eleventh International Conference on Learning
  Representations, {ICLR} 2023, Kigali, Rwanda, May 1-5, 2023}. OpenReview.net,
  2023.
\newblock URL \url{https://openreview.net/pdf?id=qFVVBzXxR2V}.

\bibitem[Srivastava et~al.(2022)Srivastava, Rastogi, Rao, Shoeb, Abid, Fisch,
  Brown, Santoro, Gupta, Garriga{-}Alonso, Kluska, Lewkowycz, Agarwal, Power,
  Ray, Warstadt, Kocurek, Safaya, Tazarv, Xiang, Parrish, Nie, Hussain, Askell,
  Dsouza, Rahane, Iyer, Andreassen, Santilli, Stuhlm{\"{u}}ller, Dai, La,
  Lampinen, Zou, Jiang, Chen, Vuong, Gupta, Gottardi, Norelli, Venkatesh,
  Gholamidavoodi, Tabassum, Menezes, Kirubarajan, Mullokandov, Sabharwal,
  Herrick, Efrat, Erdem, Karakas, and et~al.]{SrivastavaBBH2022}
Aarohi Srivastava, Abhinav Rastogi, Abhishek Rao, Abu Awal~Md Shoeb, Abubakar
  Abid, Adam Fisch, Adam~R. Brown, Adam Santoro, Aditya Gupta, Adri{\`{a}}
  Garriga{-}Alonso, Agnieszka Kluska, Aitor Lewkowycz, Akshat Agarwal, Alethea
  Power, Alex Ray, Alex Warstadt, Alexander~W. Kocurek, Ali Safaya, Ali Tazarv,
  Alice Xiang, Alicia Parrish, Allen Nie, Aman Hussain, Amanda Askell, Amanda
  Dsouza, Ameet Rahane, Anantharaman~S. Iyer, Anders Andreassen, Andrea
  Santilli, Andreas Stuhlm{\"{u}}ller, Andrew~M. Dai, Andrew La, Andrew~K.
  Lampinen, Andy Zou, Angela Jiang, Angelica Chen, Anh Vuong, Animesh Gupta,
  Anna Gottardi, Antonio Norelli, Anu Venkatesh, Arash Gholamidavoodi, Arfa
  Tabassum, Arul Menezes, Arun Kirubarajan, Asher Mullokandov, Ashish
  Sabharwal, Austin Herrick, Avia Efrat, Aykut Erdem, Ayla Karakas, and et~al.
\newblock Beyond the imitation game: Quantifying and extrapolating the
  capabilities of language models.
\newblock \emph{CoRR}, abs/2206.04615, 2022.
\newblock \doi{10.48550/ARXIV.2206.04615}.
\newblock URL \url{https://doi.org/10.48550/arXiv.2206.04615}.

\bibitem[Suzgun et~al.(2023)Suzgun, Scales, Sch{\"{a}}rli, Gehrmann, Tay,
  Chung, Chowdhery, Le, Chi, Zhou, and Wei]{SuzgunSSGTCCLCZ23}
Mirac Suzgun, Nathan Scales, Nathanael Sch{\"{a}}rli, Sebastian Gehrmann,
  Yi~Tay, Hyung~Won Chung, Aakanksha Chowdhery, Quoc~V. Le, Ed~H. Chi, Denny
  Zhou, and Jason Wei.
\newblock Challenging big-bench tasks and whether chain-of-thought can solve
  them.
\newblock In Anna Rogers, Jordan~L. Boyd{-}Graber, and Naoaki Okazaki, editors,
  \emph{Findings of the Association for Computational Linguistics: {ACL} 2023,
  Toronto, Canada, July 9-14, 2023}, pages 13003--13051. Association for
  Computational Linguistics, 2023.
\newblock \doi{10.18653/V1/2023.FINDINGS-ACL.824}.
\newblock URL \url{https://doi.org/10.18653/v1/2023.findings-acl.824}.

\bibitem[Tafjord et~al.(2021)Tafjord, Dalvi, and Clark]{TafjordDC21}
Oyvind Tafjord, Bhavana Dalvi, and Peter Clark.
\newblock Proofwriter: Generating implications, proofs, and abductive
  statements over natural language.
\newblock In Chengqing Zong, Fei Xia, Wenjie Li, and Roberto Navigli, editors,
  \emph{Findings of the Association for Computational Linguistics: {ACL/IJCNLP}
  2021, Online Event, August 1-6, 2021}, volume {ACL/IJCNLP} 2021 of
  \emph{Findings of {ACL}}, pages 3621--3634. Association for Computational
  Linguistics, 2021.
\newblock \doi{10.18653/V1/2021.FINDINGS-ACL.317}.
\newblock URL \url{https://doi.org/10.18653/v1/2021.findings-acl.317}.

\bibitem[Talmor et~al.(2019)Talmor, Herzig, Lourie, and Berant]{TalmorHLB19}
Alon Talmor, Jonathan Herzig, Nicholas Lourie, and Jonathan Berant.
\newblock Commonsenseqa: {A} question answering challenge targeting commonsense
  knowledge.
\newblock In Jill Burstein, Christy Doran, and Thamar Solorio, editors,
  \emph{Proceedings of the 2019 Conference of the North American Chapter of the
  Association for Computational Linguistics: Human Language Technologies,
  {NAACL-HLT} 2019, Minneapolis, MN, USA, June 2-7, 2019, Volume 1 (Long and
  Short Papers)}, pages 4149--4158. Association for Computational Linguistics,
  2019.
\newblock \doi{10.18653/V1/N19-1421}.
\newblock URL \url{https://doi.org/10.18653/v1/n19-1421}.

\bibitem[Team et~al.(2024)Team, Mesnard, Hardin, Dadashi, Bhupatiraju, Pathak,
  Sifre, Rivi{\`e}re, Kale, Love, et~al.]{team2024gemma}
Gemma Team, Thomas Mesnard, Cassidy Hardin, Robert Dadashi, Surya Bhupatiraju,
  Shreya Pathak, Laurent Sifre, Morgane Rivi{\`e}re, Mihir~Sanjay Kale,
  Juliette Love, et~al.
\newblock Gemma: Open models based on gemini research and technology.
\newblock \emph{arXiv preprint arXiv:2403.08295}, 2024.

\bibitem[Tyen et~al.(2023)Tyen, Mansoor, Chen, Mak, and Carbune]{Tyen2023}
Gladys Tyen, Hassan Mansoor, Peter Chen, Tony Mak, and Victor Carbune.
\newblock Llms cannot find reasoning errors, but can correct them!
\newblock \emph{CoRR}, abs/2311.08516, 2023.
\newblock \doi{10.48550/ARXIV.2311.08516}.
\newblock URL \url{https://doi.org/10.48550/arXiv.2311.08516}.

\bibitem[Wason and Johnson-Laird(1972)]{wason1972psychology}
Peter~Cathcart Wason and Philip~Nicholas Johnson-Laird.
\newblock \emph{Psychology of reasoning: Structure and content}, volume~86.
\newblock Harvard University Press, 1972.

\bibitem[Wei et~al.(2022)Wei, Wang, Schuurmans, Bosma, Ichter, Xia, Chi, Le,
  and Zhou]{Wei0SBIXCLZ22}
Jason Wei, Xuezhi Wang, Dale Schuurmans, Maarten Bosma, Brian Ichter, Fei Xia,
  Ed~H. Chi, Quoc~V. Le, and Denny Zhou.
\newblock Chain-of-thought prompting elicits reasoning in large language
  models.
\newblock In Sanmi Koyejo, S.~Mohamed, A.~Agarwal, Danielle Belgrave, K.~Cho,
  and A.~Oh, editors, \emph{Advances in Neural Information Processing Systems
  35: Annual Conference on Neural Information Processing Systems 2022, NeurIPS
  2022, New Orleans, LA, USA, November 28 - December 9, 2022}, 2022.
\newblock URL
  \url{http://papers.nips.cc/paper\_files/paper/2022/hash/9d5609613524ecf4f15af0f7b31abca4-Abstract-Conference.html}.

\bibitem[Welleck et~al.(2023)Welleck, Lu, West, Brahman, Shen, Khashabi, and
  Choi]{SelfCorrect}
Sean Welleck, Ximing Lu, Peter West, Faeze Brahman, Tianxiao Shen, Daniel
  Khashabi, and Yejin Choi.
\newblock Generating sequences by learning to self-correct.
\newblock In \emph{The Eleventh International Conference on Learning
  Representations, {ICLR} 2023, Kigali, Rwanda, May 1-5, 2023}. OpenReview.net,
  2023.
\newblock URL \url{https://openreview.net/pdf?id=hH36JeQZDaO}.

\bibitem[Xia et~al.(2024)Xia, Li, Liu, Wu, and
  Liu]{DBLP:journals/corr/abs-2404-05692}
Shijie Xia, Xuefeng Li, Yixin Liu, Tongshuang Wu, and Pengfei Liu.
\newblock Evaluating mathematical reasoning beyond accuracy.
\newblock \emph{CoRR}, abs/2404.05692, 2024.
\newblock \doi{10.48550/ARXIV.2404.05692}.
\newblock URL \url{https://doi.org/10.48550/arXiv.2404.05692}.

\bibitem[Xiong et~al.(2024)Xiong, Li, Zheng, Guo, Yin, Xie, Yang, Cao, Wang,
  Han, Tang, Li, and Liang]{Xiong2024DQ}
Jing Xiong, Zixuan Li, Chuanyang Zheng, Zhijiang Guo, Yichun Yin, Enze Xie,
  Zhicheng Yang, Qingxing Cao, Haiming Wang, Xiongwei Han, Jing Tang, Chengming
  Li, and Xiaodan Liang.
\newblock Dq-lore: Dual queries with low rank approximation re-ranking for
  in-context learning.
\newblock In \emph{The Twelfth International Conference on Learning
  Representations, {ICLR} 2024, Vienna, Austria, May 7-11, 2024}.
  OpenReview.net, 2024.
\newblock URL \url{https://openreview.net/forum?id=qAoxvePSlq}.

\bibitem[Yang et~al.(2022)Yang, Tian, Peng, and Klein]{Re3}
Kevin Yang, Yuandong Tian, Nanyun Peng, and Dan Klein.
\newblock Re3: Generating longer stories with recursive reprompting and
  revision.
\newblock \emph{CoRR}, abs/2210.06774, 2022.
\newblock \doi{10.48550/ARXIV.2210.06774}.
\newblock URL \url{https://doi.org/10.48550/arXiv.2210.06774}.

\bibitem[Yang et~al.(2024)Yang, Dong, Du, Cheng, Cambria, Liu, Gao, and
  Wei]{YangDDCCLGW24}
Zonglin Yang, Li~Dong, Xinya Du, Hao Cheng, Erik Cambria, Xiaodong Liu,
  Jianfeng Gao, and Furu Wei.
\newblock Language models as inductive reasoners.
\newblock In Yvette Graham and Matthew Purver, editors, \emph{Proceedings of
  the 18th Conference of the European Chapter of the Association for
  Computational Linguistics, {EACL} 2024 - Volume 1: Long Papers, St. Julian's,
  Malta, March 17-22, 2024}, pages 209--225. Association for Computational
  Linguistics, 2024.
\newblock URL \url{https://aclanthology.org/2024.eacl-long.13}.

\bibitem[Yao et~al.(2023)Yao, Yu, Zhao, Shafran, Griffiths, Cao, and
  Narasimhan]{ToT}
Shunyu Yao, Dian Yu, Jeffrey Zhao, Izhak Shafran, Tom Griffiths, Yuan Cao, and
  Karthik Narasimhan.
\newblock Tree of thoughts: Deliberate problem solving with large language
  models.
\newblock In Alice Oh, Tristan Naumann, Amir Globerson, Kate Saenko, Moritz
  Hardt, and Sergey Levine, editors, \emph{Advances in Neural Information
  Processing Systems 36: Annual Conference on Neural Information Processing
  Systems 2023, NeurIPS 2023, New Orleans, LA, USA, December 10 - 16, 2023},
  2023.
\newblock URL
  \url{http://papers.nips.cc/paper\_files/paper/2023/hash/271db9922b8d1f4dd7aaef84ed5ac703-Abstract-Conference.html}.

\bibitem[Yao et~al.(2024)Yao, Wu, Guo, Zhou, Gao, Luo, Hou, Fu, and
  Song]{Yao2024LeCo}
Yuxuan Yao, Han Wu, Zhijiang Guo, Biyan Zhou, Jiahui Gao, Sichun Luo, Hanxu
  Hou, Xiaojin Fu, and Linqi Song.
\newblock Learning from correctness without prompting makes {LLM} efficient
  reasoner.
\newblock \emph{CoRR}, abs/2403.19094, 2024.
\newblock \doi{10.48550/ARXIV.2403.19094}.
\newblock URL \url{https://doi.org/10.48550/arXiv.2403.19094}.

\bibitem[Yasunaga et~al.(2023)Yasunaga, Chen, Li, Pasupat, Leskovec, Liang,
  Chi, and Zhou]{Analogical}
Michihiro Yasunaga, Xinyun Chen, Yujia Li, Panupong Pasupat, Jure Leskovec,
  Percy Liang, Ed~H. Chi, and Denny Zhou.
\newblock Large language models as analogical reasoners.
\newblock \emph{CoRR}, abs/2310.01714, 2023.
\newblock \doi{10.48550/ARXIV.2310.01714}.
\newblock URL \url{https://doi.org/10.48550/arXiv.2310.01714}.

\bibitem[Ye and Durrett(2022)]{YeD22}
Xi~Ye and Greg Durrett.
\newblock The unreliability of explanations in few-shot prompting for textual
  reasoning.
\newblock In Sanmi Koyejo, S.~Mohamed, A.~Agarwal, Danielle Belgrave, K.~Cho,
  and A.~Oh, editors, \emph{Advances in Neural Information Processing Systems
  35: Annual Conference on Neural Information Processing Systems 2022, NeurIPS
  2022, New Orleans, LA, USA, November 28 - December 9, 2022}, 2022.
\newblock URL
  \url{http://papers.nips.cc/paper\_files/paper/2022/hash/c402501846f9fe03e2cac015b3f0e6b1-Abstract-Conference.html}.

\bibitem[Yu et~al.(2023)Yu, Zhang, Liang, Jiang, and Sabharwal]{PlugandPlay}
Wenhao Yu, Zhihan Zhang, Zhenwen Liang, Meng Jiang, and Ashish Sabharwal.
\newblock Improving language models via plug-and-play retrieval feedback.
\newblock \emph{CoRR}, abs/2305.14002, 2023.
\newblock \doi{10.48550/ARXIV.2305.14002}.
\newblock URL \url{https://doi.org/10.48550/arXiv.2305.14002}.

\bibitem[Zeng et~al.(2023)Zeng, Chen, Liu, Jiang, and
  Jia]{DBLP:journals/corr/abs-2312-17080}
Zhongshen Zeng, Pengguang Chen, Shu Liu, Haiyun Jiang, and Jiaya Jia.
\newblock Mr-gsm8k: A meta-reasoning benchmark for large language model
  evaluation.
\newblock \emph{CoRR}, abs/2312.17080, 2023.
\newblock \doi{10.48550/ARXIV.2312.17080}.
\newblock URL \url{https://doi.org/10.48550/arXiv.2312.17080}.

\bibitem[Zhang et~al.(2021)Zhang, Jia, Edmonds, Zhu, and Zhu]{ZhangACRE2021}
Chi Zhang, Baoxiong Jia, Mark Edmonds, Song{-}Chun Zhu, and Yixin Zhu.
\newblock {ACRE:} abstract causal reasoning beyond covariation.
\newblock In \emph{{IEEE} Conference on Computer Vision and Pattern
  Recognition, {CVPR} 2021, virtual, June 19-25, 2021}, pages 10643--10653.
  Computer Vision Foundation / {IEEE}, 2021.
\newblock \doi{10.1109/CVPR46437.2021.01050}.
\newblock URL
  \url{https://openaccess.thecvf.com/content/CVPR2021/html/Zhang\_ACRE\_Abstract\_Causal\_REasoning\_Beyond\_Covariation\_CVPR\_2021\_paper.html}.

\bibitem[Zhou et~al.(2024)Zhou, Wan, Proleev, Mincu, Chen, Heller, and
  Roy]{zhou2024batch}
Han Zhou, Xingchen Wan, Lev Proleev, Diana Mincu, Jilin Chen, Katherine~A
  Heller, and Subhrajit Roy.
\newblock Batch calibration: Rethinking calibration for in-context learning and
  prompt engineering.
\newblock In \emph{The Twelfth International Conference on Learning
  Representations}, 2024.
\newblock URL \url{https://openreview.net/forum?id=L3FHMoKZcS}.

\bibitem[Zhou et~al.(2023)Zhou, Muresanu, Han, Paster, Pitis, Chan, and
  Ba]{ZhouMHPPCB23}
Yongchao Zhou, Andrei~Ioan Muresanu, Ziwen Han, Keiran Paster, Silviu Pitis,
  Harris Chan, and Jimmy Ba.
\newblock Large language models are human-level prompt engineers.
\newblock In \emph{The Eleventh International Conference on Learning
  Representations, {ICLR} 2023, Kigali, Rwanda, May 1-5, 2023}. OpenReview.net,
  2023.
\newblock URL \url{https://openreview.net/pdf?id=92gvk82DE-}.

\end{thebibliography}
\clearpage 
\appendix
\section{Appendix}

\subsection{Limitations}
\label{app:limitations}

The meta-reasoning evaluation framework in \data, while innovative, is not without its limitations. Firstly, its applicability may be restricted when it comes to subjects that are inherently holistic or creative in nature, such as humanities or sociology. These subjects often require a comprehensive understanding and modification (e.g. essay writing), which can be challenging to break down into specific, sequential reasoning steps and corrections. Secondly, \data is currently confined to questions in English. This could potentially limit the scope of reasoning challenges that can be explored, as different languages may present unique cognitive and linguistic hurdles. Lastly, the analysis and correction of errors in the reasoning steps are currently based on solutions generated by three LLMs, namely GPT-3.5, Mistral-Medium, and Claude 2. It's important to note that different LLMs and different individuals, may exhibit distinct reasoning and error patterns. Therefore, it would be beneficial to broaden the spectrum of solutions analyzed, incorporating a more diverse range of LLMs and even human responses. This would not only enhance the robustness of the evaluation framework but also provide a more nuanced understanding of the reasoning processes at play.

\subsection{Broader Impact}
\label{app:impacts}

\paragraph{Positive Societal Impacts}
The proposed dataset \data has the potential to bring about significant positive societal impacts. It can contribute to the development and enhancement of LLMs by providing a comprehensive benchmark suite, which researchers and developers can use to identify and address the limitations and weaknesses of their models. This can lead to more accurate, efficient, and reliable LLMs. The meta-reasoning paradigm might open new avenues in AI research, leading to a deeper understanding of reasoning capabilities and the development of innovative methodologies for their evaluation and improvement. Moreover, with a wide range of subjects, \data can be a valuable resource for educational AI tools, providing personalized learning experiences and helping students understand and improve their reasoning skills. AI systems with improved reasoning capabilities can also be instrumental in various sectors, including healthcare, finance, and environmental management, aiding in complex decision-making and problem-solving tasks.

\paragraph{Negative Societal Impacts}
\data may also present potential negative societal impacts. As with any technology, there is a risk of LLMs being misused or used maliciously. For instance, LLMs with advanced reasoning capabilities could be used to manipulate information or deceive people. The use of LLMs in decision-making and problem-solving tasks could lead to an over-reliance on these systems, potentially undermining human judgment and critical thinking skills. Advanced LLMs, especially those used in sensitive sectors like healthcare and finance, need to handle vast amounts of data, which can raise privacy and security concerns if not managed properly.

\subsection{Additional Related Work}
\label{app:add_related_work}

\paragraph{Improving Reasoning Abilities of LLMs}
To enhance the reasoning capabilities of LLMs, prior research primarily focuses on specific prompting techniques~\citep{gpt3}. Existing efforts include few-shot prompting with intermediate steps augmented demonstrations~\citep{Wei0SBIXCLZ22,ToT,Xiong2024DQ} or zero-shot prompting with specific instructions~\citep{KojimaGRMI22,Analogical}. Although these methods have shown promising results, their effectiveness is often constrained by their task-specific nature and the labour-intensive process of designing prompts, leading to inconsistent outcomes across different tasks~\citep{YeD22, ZhouMHPPCB23}. Another strategy to facilitate reasoning involves instruction tuning or knowledge distillation, which elicits reasoning paths from LLMs without explicit prompting~\citep{Chung2022,Mukherjee2023,Gunasekar2023,Lu2024Auto}.  These approaches typically involve resource-intensive fine-tuning over LLMs and require a large set of examples annotated with CoT.

\paragraph{Learning From Feedback}
Improving LLMs through learning from feedback has become a prevalent strategy, notably through reinforcement learning from human feedback, which seeks to align LLMs with human values by refining their outputs based on feedback~\citep{InstructGPT,ConstitutionalAI}. However, this method faces challenges such as high costs due to manual labor and a lack of real-time feedback capabilities~\citep{fernandes2023bridging}. An alternative strategy involves using self-correcting LLMs, which rely on automated feedback to iteratively adapt and understand the consequences of their actions without relying on humans. This feedback can be derived from outside sources such as other models~\citep{Re3,Lu2024PDA}, tools~\citep{Critic,Huang2024Soap}, knowledge bases~\citep{RARR,PlugandPlay}, evaluation metrics~\citep{MaieuticPrompt,SelfCorrect} or generation logits~\citep{Yao2024LeCo}.

\section{Robustness of MR-Score}
\label{app:mr_score_robustness}

\begin{table*}[t]
\centering
\begin{tabular}{|l|c|c|c|c|c|c|c|}
\hline
\textbf{Model} & \textbf{Coding} & \textbf{Phy.} & \textbf{Bio.} & \textbf{Math} & \textbf{Med.} & \textbf{Chem.} & \textbf{Logic} \\ 
\hline
gpt-4-turbo     & 83/55   & 137/15  & 164/11  & 305/46 & 194/25   & 166/27    & 192/16 \\ 
deepseek\_coder & 100/38  & 145/7   & 167/8   & 321/30 & 200/19   & 172/21    & 193/15 \\ 
Qwen2-72B       & 99/39   & 142/10  & 167/8   & 312/39 & 195/24   & 172/21    & 200/8  \\ 
\hline
\end{tabular}
\caption{Scoring of error reasons from different models across subjects.}
\label{tab:model_error_reason}
\end{table*}

\begin{table*}[t]
\centering
\begin{tabular}{|l|c|c|c|c|c|c|c|}
\hline
\textbf{} & \textbf{Coding} & \textbf{Phy.} & \textbf{Bio.} & \textbf{Med.} & \textbf{Chem.} & \textbf{Logic} & \textbf{Math} \\ 
\hline
Agreement Ratio & 7/8 & 12/13 & 21/21 & 12/12 & 15/17 & 15/16 & 10/13 \\ 
\hline
\end{tabular}
\caption{Agreement ratio between the author and the proxy scoring model across different subjects.}
\label{tab:human_model_agreement}
\end{table*}

\textbf{Question}: Does the \texttt{ACC\_reason} metric’s dependency on the judgments of different LLMs or human evaluators lead to variability in scoring ?

\textbf{Answer}: We would like to argue that due to the careful design of our evaluation mechanism, the automatic scoring of error reasons is both robust and economically feasible:

\begin{itemize}
    \item \textbf{Multiple annotators}: During the annotation stage, we collected multiple annotations for the first error reasons and potential error rectification from different annotators who agreed on the solution correctness and the first error step.

    \item \textbf{Proxy Model Evaluation}: Based on the ground truth annotations collected from various perspectives, the proxy language model (e.g., GPT-4-Turbo) then examines the \textbf{error reasons} provided by evaluating models. Given the question/solution pair and information regarding the first error step, error reasons, and rectification, the potential flaws of the error reasons provided by the evaluating models are easy to diagnose under contrast.

    \item \textbf{ACC\_reason robustness}: Table-\ref{tab:model_error_reason} shows the scores of error reasons sampled from our evaluation results. For the same set of error reasons collected in each subject, three different models made their predictions on correctness/incorrectness. We can clearly see the consistency of their predictions among the three models over questions in all subjects. Since the MR-Score is a weighted metric, the final score variability is less than 1 percent in total.
\end{itemize}

\noindent

\textbf{Human-Model Agreement Rate}: As mentioned in \ref{sec:dataset}, the agreement rate between manual annotations and the GPT-4 predictions over 100 samples randomly collected from all subjects is 92\%. Below is the exact detail of our setup:

We randomly collected 100 data instances where the evaluating model correctly identified the solution correctness and the first error step across all subjects. We then manually examined whether the proxy scoring model (e.g., GPT-4-Turbo-2024-04-09) correctly scored the error reasons of the evaluating models. Table-\ref{tab:human_model_agreement} is the detailed composition of the ratio in which the author agrees with the proxy scoring model. The annotation time varies significantly across subjects, as some problems—such as coding and chemistry—can take more than 10 minutes to evaluate, while subjects like biology are easier to assess. This high agreement rate further supports the reliability of our evaluation, thus avoiding the need for manual annotation of potentially 138,000 problems (6,000 benchmark size times 23 models evaluated).

\begin{table*}[t]
\caption{Evaluation results breakdown on MR-Ben: This table presents a detailed breakdown of each model’s performance evaluated under metric MCC/ACC-step/ACC-reason across different subjects. Here k stands for number of shot and every model we used in this experiment are instruction-tuned.}
\setlength\extrarowheight{3.5pt}
\renewcommand{\arraystretch}{0.85}
\setlength{\tabcolsep}{2pt}

\resizebox{\textwidth}{!}{
\begin{tabular}{lrrp{1.5mm}rrp{1.5mm}rrp{1.5mm}rrp{1.5mm}rrp{1.5mm}rrp{1.5mm}rrp{1.5mm}rr}
\toprule
\multirow{2}{*}{\textbf{Model}} & \multicolumn{2}{c}{\textbf{Bio.}} & & \multicolumn{2}{c}{\textbf{Phy.}} && \multicolumn{2}{c}{\textbf{Math}} && \multicolumn{2}{c}{\textbf{Chem.}} && \multicolumn{2}{c}{\textbf{Med.}} && \multicolumn{2}{c}{\textbf{Logic}} && \multicolumn{2}{c}{\textbf{Coding}} && \multicolumn{2}{c}{\textbf{Avg.}} \\[1pt] 
\cline{2-3} \cline{5-6} \cline{8-9} \cline{11-12} \cline{14-15} \cline{17-18} \cline{20-21} \cline{23-24} 
& $k$=0 & $k$=1 && $k$=0  & $k$=1  && $k$=0 & $k$=1  && $k$=0 & $k$=1 && $k$=0  & $k$=1  && $k$=0 & $k$=1  && $k$=0 & $k$=1 && $k$=0  & $k$=1   \\ 
 
\midrule
\multicolumn{24}{c}{\textbf{MR-Scores}}\\
\midrule

Claude3-Haiku & 5.7 & 5.8 && 3.3 & 3.5 && 3.1 & 3.1 && 6.5 & 6.4 && 2.0 & 2.0 && 1.2 & 1.2 && 9.0 & 0.0 && 4.4 & 3.1  \\

GPT-3.5-Turbo & 3.6 & 6.6 && 5.7 & 6.7 && 5.7 & 5.4 && 4.9 & 6.7 && 3.6 & 4.4 && 1.7 & 4.5 && 3.0 & 4.1 && 4.0 & 5.5  \\

Phi3-3.8B & 13.4 & 12.5 && 12.7 & 10.8 && 13.3 & 13.1 && 16.4 & 17.1 && 10.2 & 8.1 && 8.4 & 5.3 && 9.1 & 10.2 && \textbf{11.9} & \textbf{11.0}   \\

Deepseek-Coder-33B & 7.4 & 5.5 && 7.8 & 5.6 && 7.2 & 8.6 && 7.8 & 7.4 && 6.0 & 5.5 && 4.6 & 6.7 && 8.4 & 4.9 && 7.0 & 6.3  \\

DeepSeek-Coder-7B & 10.5 & 9.9 && 11.8 & 9.6 && 11.8 & 12.1 && 12.3 & 11.9 && 10.4 & 11.0 && 9.8 & 10.7 && 5.0 & 5.8 && 10.2 & 10.2  \\

LLaMA3-8B & 12.0 & 11.9 && 10.9 & 7.5 && 15.0 & 9.0 && 12.6 & 12.7 && 9.3 & 8.0 && 9.4 & 9.6 && 15.8 & 10.0 && \textbf{12.2} & 9.8  \\

Qwen1.5-72B & 15.3 & 19.2 && 12.9 & 13.6 && 12.0 & 10.0 && 13.9 & 16.3 && 11.7 & 14.7 && 10.4 & 12.9 && 3.9 & 5.9 && 11.5 & 13.3  \\

DeepSeek-67B & 17.1 & 19.7 && 14.9 & 17.3 && 15.4 & 16.2 && 16.3 & 20.6 && 14.7 & 12.2 && 13.6 & 14.3 && 14.5 & 15.2 && 15.2 & 16.5  \\

LLaMA3-70B & 20.4 & 27.1 && 17.4 & 20.5 && 14.9 & 15.8 && 19.5 & 25.1 && 16.3 & 19.3 && 16.3 & 16.8 && 29.8 & 16.7 && 19.2 & 20.2  \\

Mistral-Large & 22.2 & 28.0 && 26.7 & 25.4 && 24.3 & 28.2 && 24.0 & 27.0 && 15.9 & 19.3 && 14.7 & 17.1 && 21.1 & 21.4 && 21.3 & 23.8  \\

DeepSeek-V2-236B & 30.0 & 37.1 && 32.2 & 36.5 && 32.2 & 30.0 && 32.5 & 35.4 && 26.5 & 32.4 && 23.6 & 27.4 && 34.2 & 27.1 && 30.2 & 32.3  \\

GPT-4-Turbo & 44.7 & 47.3 && 42.8 & 45.2 && 44.3 & 45.4 && 44.0 & 46.0 && 38.8 & 38.4 && 34.1 & 33.6 && 53.6 & 57.3 && \textbf{43.2} & \textbf{44.7}  \\

\midrule
\multicolumn{24}{c}{\textbf{MCC-Matthews Correlation Coefficient}}\\
\midrule

Claude3-Haiku & 13.96 &  17.72  && 16.47 &  13.62  && 15.09 &  10.74  && 16.54 &  19.96  && 8.52 &  8.35  && 6.21 &  4.94  && 4.36 &  0  &&  11.59 & 10.76  \\ 

GPT-3.5-Turbo & 10.72 &  19.44  && 16.66 &  21.33  && 17.48 &  17.45  && 18.24 &  12.6  && 11.19 &  13.28  && 4.07 &  0  && 12.35 &  12.35  &&  12.96 & 13.78  \\ 

Deepseek-Coder-33B & 7.51 &  8.57  && 11.73 &  6.81  && 9.69 &  21.06  && 9.98 &  7.94  && 1.62 &  6.28  && 0 &  0  && 26.18 &  15.44  &&  9.53 & 9.44  \\ 

Deepseek-Coder-7B & 4.96 &  9.79  && 8.77 &  6.72  && 9.05 &  10.82  && 10.49 &  9.39  && 5.02 &  3.17  && 3.22 &  2.58  && 10.91 &  6.27  &&  7.49 & 6.96  \\ 

LlaMA3-8B & 19.37 &  21.15  && 16.24 &  18.64  && 26.55 &  21.87  && 25.99 &  28.6  && 14.92 &  18.95  && 11.8 &  16.24  && 14.54 &  15.72  &&  18.49 & 20.17  \\ 

Phi3-3.8B & 27.66 &  28.48  && 21.61 &  21.44  && 22.29 &  25.17  && 30.92 &  33.37  && 17.36 &  14.9  && 13.03 &  9.56  && 14.48 &  18.76  &&  21.05 & 21.67  \\ 

Qwen1.5-72B & 33.64 &  42.44  && 31.4 &  31.56  && 29.2 &  23.28  && 35.47 &  36.47  && 21.76 &  29.64  && 24.42 &  27.74  && 13.8 &  15.69  &&  27.1 & 29.55  \\ 

Deepseek-67B & 43.61 &  41.73  && 24.16 &  28.77  && 24.95 &  23.87  && 36.58 &  37.29  && 27.8 &  28.93  && 26.74 &  25.09  && 28.23 &  29.06  &&  30.3 & 30.68  \\ 

LlaMA3-70B & 45.67 &  56.14  && 40.34 &  41.3  && 32.76 &  30.94  && 41.72 &  52.12  && 33.18 &  37.75  && 32.0 &  33.87  && 47.86 &  29.67  &&  39.08 & 40.26  \\ 

Mistral-Large & 41.67 &  49.0  && 34.24 &  33.47  && 29.0 &  37.05  && 41.99 &  47.07  && 23.76 &  32.05  && 25.66 &  33.25  && 37.05 &  33.52  &&  33.34 & 37.92  \\ 

Deepseek-v2-236B & 52.96 &  53.38  && 41.81 &  46.48  && 43.75 &  40.53  && 54.32 &  50.15  && 37.61 &  44.53  && 36.36 &  35.41  && 45.89 &  35.7  &&  44.67 & 43.74  \\ 

GPT-4-Turbo & 63.33 &  62.59  && 52.9 &  52.7  && 50.67 &  52.84  && 53.05 &  54.59  && 56.79 &  54.66  && 40.95 &  42.94  && 52.5 &  57.53  &&  52.88 & 53.98  \\

\midrule
\multicolumn{24}{c}{\textbf{Accuracy of First Error Step}}\\
\midrule
Claude3-Haiku & 2.15 &  3.1  && 1.4 &  1.12  && 2.38 &  1.59  && 1.77 &  4.42  && 1.69 &  0.68  && 1.01 &  0.29  && 0.0 &  0.0  &&  1.49 & 1.6  \\ 

GPT-3.5-Turbo & 2.86 &  4.53  && 4.2 &  4.76  && 4.37 &  3.84  && 2.87 &  8.17  && 2.37 &  3.05  && 1.73 &  7.63  && 0.61 &  2.44  &&  2.72 & 4.92  \\ 

Deepseek-Coder-33B & 14.83 &  10.29  && 14.94 &  12.36  && 14.69 &  10.92  && 15.67 &  16.31  && 14.54 &  12.43  && 12.22 &  18.18  && 5.49 &  3.05  &&  13.2 & 11.93  \\ 

Deepseek-Coder-7B & 21.77 &  18.18  && 23.28 &  19.83  && 23.41 &  20.03  && 24.46 &  23.18  && 23.29 &  26.09  && 20.03 &  24.72  && 4.27 &  6.1  &&  20.07 & 19.73  \\ 

LlaMA3-8B & 14.35 &  14.35  && 17.53 &  8.62  && 20.29 &  7.8  && 14.16 &  11.59  && 13.13 &  8.41  && 13.64 &  11.36  && 17.68 &  9.76  &&  15.83 & 10.27  \\ 

Phi3-3.8B & 12.68 &  11.48  && 16.38 &  12.07  && 17.69 &  16.12  && 18.03 &  17.17  && 12.78 &  8.76  && 10.23 &  6.96  && 8.54 &  9.15  &&  13.76 & 11.67  \\ 

Qwen1.5-72B & 11.48 &  15.31  && 10.63 &  11.49  && 10.79 &  9.88  && 12.45 &  14.38  && 11.03 &  13.49  && 8.1 &  10.94  && 1.83 &  4.27  &&  9.47 & 11.39  \\ 

Deepseek-67B & 13.16 &  19.14  && 19.25 &  21.84  && 20.81 &  22.11  && 17.17 &  23.39  && 14.71 &  12.08  && 12.78 &  13.49  && 12.2 &  14.02  &&  15.72 & 18.01  \\ 

LlaMA3-70B & 15.79 &  22.25  && 14.66 &  18.39  && 13.65 &  15.08  && 17.81 &  22.32  && 14.36 &  16.64  && 14.91 &  14.91  && 26.83 &  13.41  &&  16.86 & 17.57  \\ 

Mistral-Large & 18.38 &  25.54  && 26.33 &  28.29  && 27.28 &  33.25  && 26.05 &  27.59  && 16.92 &  19.29  && 14.53 &  14.96  && 19.51 &  19.51  &&  21.29 & 24.06  \\ 

Deepseek-v2-236B & 27.51 &  35.41  && 37.64 &  40.23  && 36.28 &  34.33  && 33.91 &  37.55  && 27.5 &  32.57  && 22.87 &  27.56  && 33.54 &  26.83  &&  31.32 & 33.5  \\ 

GPT-4-Turbo & 41.77 &  46.06  && 42.58 &  46.5  && 46.49 &  47.81  && 42.6 &  46.8  && 37.06 &  36.72  && 29.93 &  33.24  && 50.61 &  59.15  &&  41.58 & 45.18  \\

\midrule
\multicolumn{24}{c}{\textbf{Accuracy of First Error Reason}}\\
\midrule
Claude3-Haiku & 1.67 &  2.63  && 0.56 &  0.84  && 1.19 &  0.93  && 0.22 &  2.21  && 1.18 &  0.34  && 0.72 &  0.29  && 0.0 &  0.0  &&  0.79 & 1.03  \\ 

GPT-3.5-Turbo & 1.19 &  2.63  && 2.24 &  1.96  && 1.85 &  1.46  && 0.88 &  3.53  && 1.35 &  1.69  && 0.72 &  4.17  && 0.61 &  1.83  &&  1.26 & 2.47  \\ 

Deepseek-Coder-33B & 2.87 &  1.44  && 2.01 &  1.15  && 1.69 &  2.21  && 2.15 &  1.93  && 2.63 &  1.05  && 1.85 &  2.56  && 3.05 &  1.83  &&  2.32 & 1.74  \\ 

Deepseek-Coder-7B & 5.98 &  5.02  && 6.03 &  4.6  && 5.98 &  7.93  && 5.79 &  6.22  && 4.9 &  5.08  && 6.25 &  5.54  && 3.05 &  5.49  &&  5.43 & 5.7  \\ 

LlaMA3-8B & 7.66 &  6.7  && 4.89 &  2.3  && 7.28 &  4.55  && 6.22 &  7.08  && 4.73 &  3.33  && 5.97 &  5.97  && 15.24 &  7.93  &&  7.43 & 5.41  \\ 

Phi3-3.8B & 8.13 &  6.7  && 6.9 &  5.75  && 7.02 &  6.37  && 9.66 &  10.52  && 5.78 &  5.08  && 5.4 &  2.7  && 7.32 &  7.32  &&  7.17 & 6.35  \\ 

Qwen1.5-72B & 10.29 &  12.2  && 6.9 &  7.76  && 5.85 &  4.81  && 6.22 &  9.44  && 8.06 &  9.46  && 6.11 &  8.24  && 1.22 &  3.05  &&  6.38 & 7.85  \\ 

Deepseek-67B & 8.85 &  11.24  && 8.62 &  10.06  && 8.32 &  9.49  && 7.73 &  12.23  && 9.46 &  5.6  && 8.81 &  10.37  && 10.37 &  10.37  &&  8.88 & 9.91  \\ 

LlaMA3-70B & 13.16 &  18.42  && 9.77 &  13.51  && 8.58 &  10.27  && 11.59 &  15.88  && 10.68 &  13.49  && 10.94 &  11.08  && 24.39 &  13.41  &&  12.73 & 13.72  \\ 

Mistral-Large & 15.27 &  21.0  && 19.05 &  20.45  && 15.1 &  21.72  && 16.56 &  18.54  && 13.03 &  14.21  && 11.22 &  11.94  && 17.07 &  17.68  &&  15.33 & 17.94  \\ 

Deepseek-v2-236B & 22.25 &  31.58  && 25.57 &  30.17  && 25.1 &  23.28  && 22.96 &  28.11  && 21.54 &  27.5  && 18.89 &  24.15  && 29.88 &  23.78  &&  23.74 & 26.94  \\ 

GPT-4-Turbo & 39.14 &  42.0  && 38.38 &  41.46  && 40.4 &  41.06  && 36.64 &  42.16  && 32.83 &  32.83  && 27.63 &  30.07  && 50.61 &  56.1  &&  37.95 & 40.81  \\

\bottomrule
\end{tabular}
}
\label{tab:main_table_sub_metrics}
\end{table*}

\textbf{Question}: Is the MR-Score sensitive to different weightings? Is MR-Score a robust unified metric?

Table-\ref{tab:main_table_sub_metrics} shows breakdown performance for models in all four metrics (MR-Score, MCC, ACC\_step, and ACC\_reason):

\begin{enumerate}
    \item \textbf{Metric Robustness}: Due to the progressive nature of the definitions of our subtasks (e.g., the success of subsequent tasks depends on the previous ones), we can see the diminishing trend in the scores of MCC, ACC\_step, and ACC\_reason. However, thanks to the design of our evaluation mechanism and metrics, the \textbf{score rankings of different models stay in relatively stable order across metrics}. In other words, we have not observed any model that excels in determining the solution correctness (thus high in MCC) but is unable to explain the rationale behind it (e.g., low in ACC\_reason).

    \item \textbf{Task Difficulties}: As shown in the breakdown table, the ACC\_reason metric is more discriminative than the MCC metric for competent models but vice versa for the less competent ones. This aligns with our intuition that generally more difficult questions are more discriminative for strong candidates, while weaker ones are simply incapable of solving them. \textbf{This phenomenon could in part explain why in general the MR-Score is not very sensitive to minor changes in the weightings assigned to the subtasks}, since the differentiability of the subtask metrics tends to reconcile with each other under different scenarios.

    \item \textbf{Differentiability and Interpretability}: The weights of the MR-Score are ultimately decided by considering both the discriminative ability and the interpretability. To best differentiate models with different evaluation results, we conducted a thorough grid search to investigate the impact of the weightings. Since the weightings calculated returned a few optimal instances, we deliberately selected the one that assigns higher scores to more difficult tasks. \textbf{We believe the current weighting ratio strikes a good balance between interpretability and differentiation}: For example, GPT-4-Turbo, Deepseek-v2-236B, and Mistral-Large achieve 86.4\%, 78.5\%, and 81.2\% respectively in MMLU but score 43.2\%, 29.4\%, and 21.3\% in our benchmark.
\end{enumerate}

\section{More Discussion on Biases}
\label{app:bias_correlation}

To quantitatively assess the relationship between the length of solutions and their correctness, Pearson-Correlation-Coefficients were calculated and reported in Table-\ref{tab:pearson_results} in the Appendix. The result suggests varying dynamics across disciplines regarding how solution length impacts the likelihood of correctness. For subjects such as coding, chemistry and math, longer solutions are less likely to be correct, which could suggest that complexity or elaboration in responses may lead to mistakes or incorrect reasoning. For medicine, despite being weak, there's a tendency for longer solutions to be slightly more correct, possibly due to more detailed or thorough explanations being favorable. For the other subjects, length of solution does not appear to significantly affect correctness, indicating that other factors likely play a more dominant role in determining solution quality. The overall Pearson Coefficients analysis reflects the distinct nature of problem-solving in each field of our benchmark.

\begin{table}[ht]
\centering
\caption{Pearson Correlation Between Solution Length and Correctness}
\label{tab:pearson_results}
\begin{tabular}{lcc}
\toprule
\textbf{Subject} & \textbf{Pearson Correlation} & \textbf{P-value} \\
\midrule
Medicine       & 0.094  & 0.0072 \\
Physics        & -0.061 & 0.111 \\
Biology        & 0.009  & 0.783 \\
Chemistry      & -0.127 & 0.00018 \\
Coding         & -0.199 & 0.0021 \\
Logic          & 0.0002 & 0.995 \\
Math           & -0.115 & 0.00049 \\
\bottomrule
\end{tabular}
\end{table}
\begin{table}
\caption{Evaluation Results of Models on MR-Bean in few-shot settings: This table presents a detailed breakdown of each model’s performance evaluated under metric MR-Score across different subjects.}
\centering

\renewcommand{\arraystretch}{0.75}
\setlength{\tabcolsep}{8pt}

\resizebox{\textwidth}{!}{ 
\begin{tabular}{lcrrrrrrrr}
\toprule
\textbf{Model} & \textbf{k-shot} & \textbf{Bio.} &  \textbf{Phy.} & \textbf{Math} & \textbf{Chem.} & \textbf{Med.} & \textbf{Logic} & \textbf{Coding} & \textbf{Avg.} \\ 

\midrule
\multirow{4}{*}{Gemma-2B} & 0 & 0.1 & 0.0 & 0.0 & 0.1 & 0.0 & 0.0 & 0.7 & 0.1 \\
 & 1 & 0.0 & 0.0 & 1.0 & 0.0 & 0.4 & 0.2 & 0.0 & 0.2 \\
 & 2 & 0.1 & 0.2 & 0.7 & 0.6 & 0.7 & 0.2 & 0.0 & \textbf{0.4} \\
 & 3 & 0.1 & 0.3 & 1.1 & 0.1 & 0.7 & 0.3 & 0.0 & 0.4 \\

\midrule
\multirow{4}{*}{Llama3-8B} & 0 & 11.1 & 14.9 & 14.8 & 12.8 & 9.4 & 9.6 & 9.1 & \textbf{11.7} \\
& 1 & 11.7 & 8.1 & 7.8 & 12.8 & 7.3 & 10.7 & 5.7 & 9.2 \\
& 2 & 9.7 & 7.8 & 11.1 & 8.8 & 6.4 & 6.2 & 2.4 & 7.5 \\
& 3 & 10.0 & 10.7 & 8.3 & 8.2 & 5.5 & 5.3 & 3.0 & 7.3 \\


\midrule
\multirow{4}{*}{Llama3-70B} & 0 & 19.9 & 15.4 & 15.0 & 17.6 & 14.6 & 13.5 & 28.2 & 17.7 \\
& 1 & 30.5 & 21.4 & 16.8 & 26.2 & 16.9 & 16.0 & 15.3 & \textbf{20.4} \\
& 2 & 27.2 & 19.9 & 16.8 & 22.0 & 15.9 & 17.5 & 19.5 & 19.8 \\
& 3 & 27.2 & 20.6 & 16.3 & 21.1 & 16.0 & 14.6 & 19.4 & 19.3 \\


\midrule
\multirow{4}{*}{GPT-4-Turbo} & 0 & 44.7 & 42.8 & 44.3 & 44.0 & 38.8 & 34.1 & 53.6 & 43.2 \\
& 1 & 47.3 & 45.2 & 45.4 & 46.0 & 38.4 & 33.6 & 57.3 & \textbf{44.7} \\
& 2 & 46.6 & 42.7 & 44.9 & 43.3 & 42.1 & 35.9 & 53.0 & 44.1 \\
& 3 & 44.0 & 44.8 & 46.5 & 44.4 & 41.2 & 33.7 & 56.6 & 44.5 \\
\bottomrule
\end{tabular}
}
\label{tab:few_shot_results}
\end{table}

\begin{figure}[t]
    \centering
    \includegraphics[width=\textwidth]{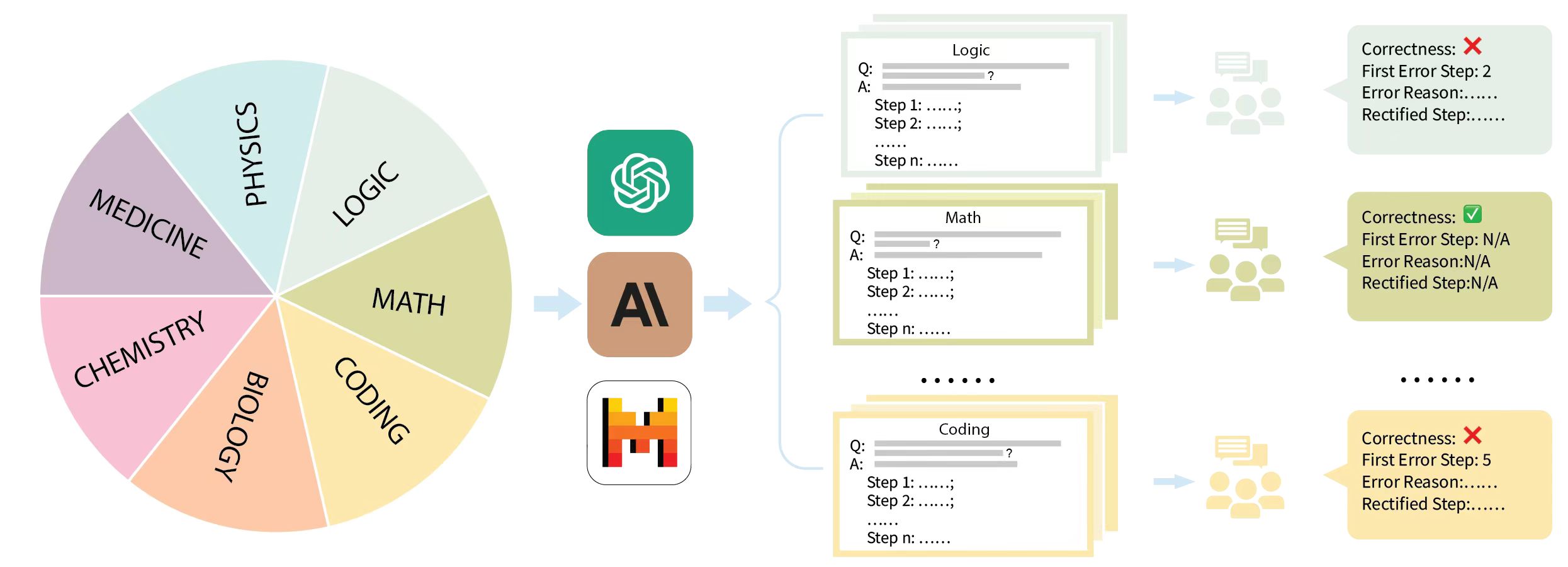}
    \vspace{-0.5cm}
    \caption{This is the illustration of the dataset creation pipeline of \data~. We first compile a set of questions from different subjects and then collect solutions from different LLMs. For each subject, a group of domain experts is recruited to annotate each question solution pair on its solution correctness, first error step, and error reasons.}
    \label{fig:dataset_pipeline}
\end{figure}

\begin{figure}[t]
    \centering
    \includegraphics[width=\textwidth]{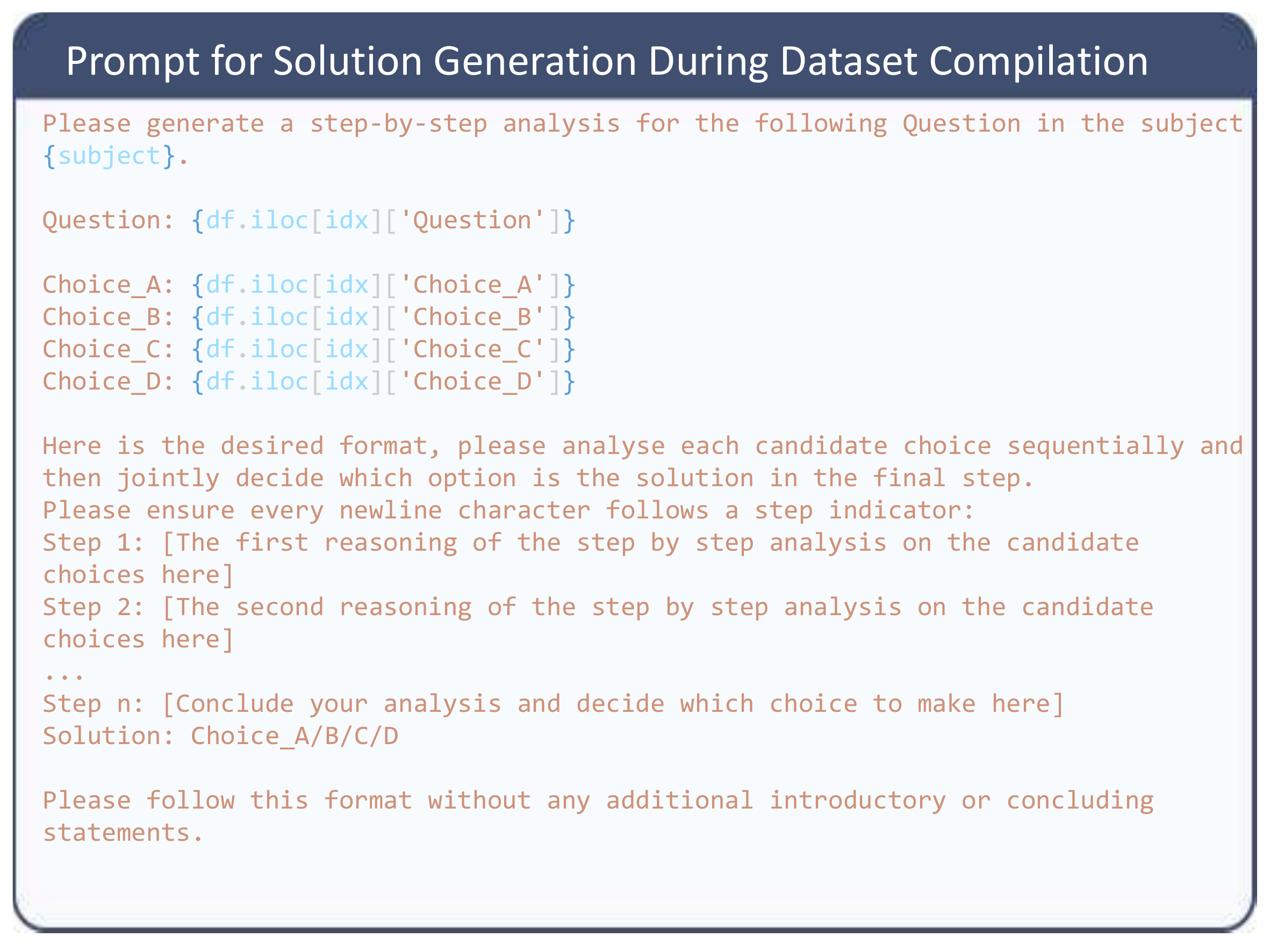}
    \vspace{-0.5cm}
    \caption{This is the prompt we used for solution generation during the dataset compilation stage. Note that besides coding, every subject question in our dataset takes the form of multiple choice problem.}
    \label{fig:sol_gen_prompt}
\end{figure}

\begin{figure}[t]
    \centering
    \includegraphics[width=\textwidth]{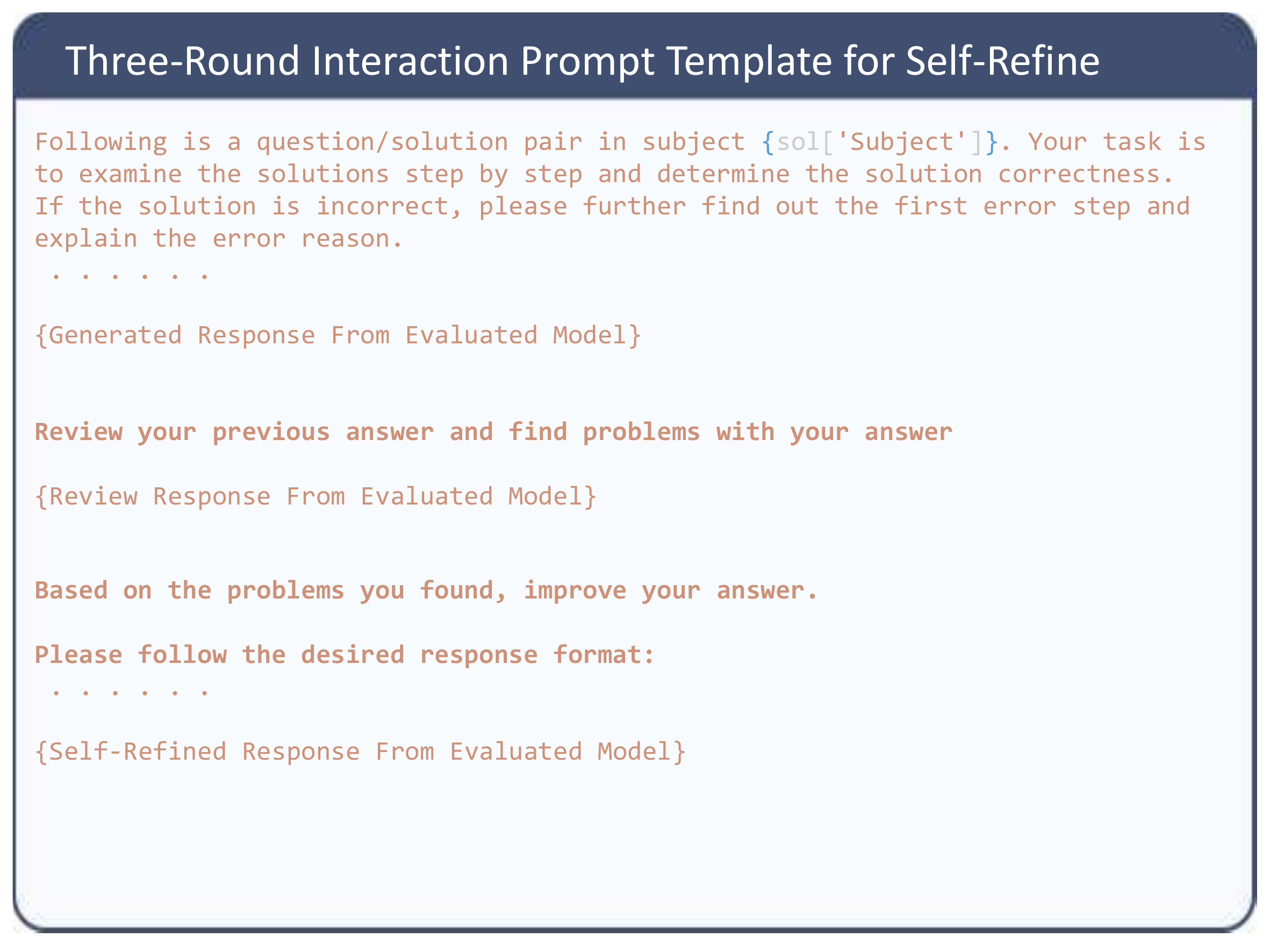}
    \vspace{-0.5cm}
    \caption{This is the prompt we used for self-refine experiment, note that three consecutive inference calls are made in order to perform the most basic self correction.}
    \label{fig:self_refine_prompt}
\end{figure}

\begin{figure}[t]
    \centering
    \includegraphics[width=\textwidth]{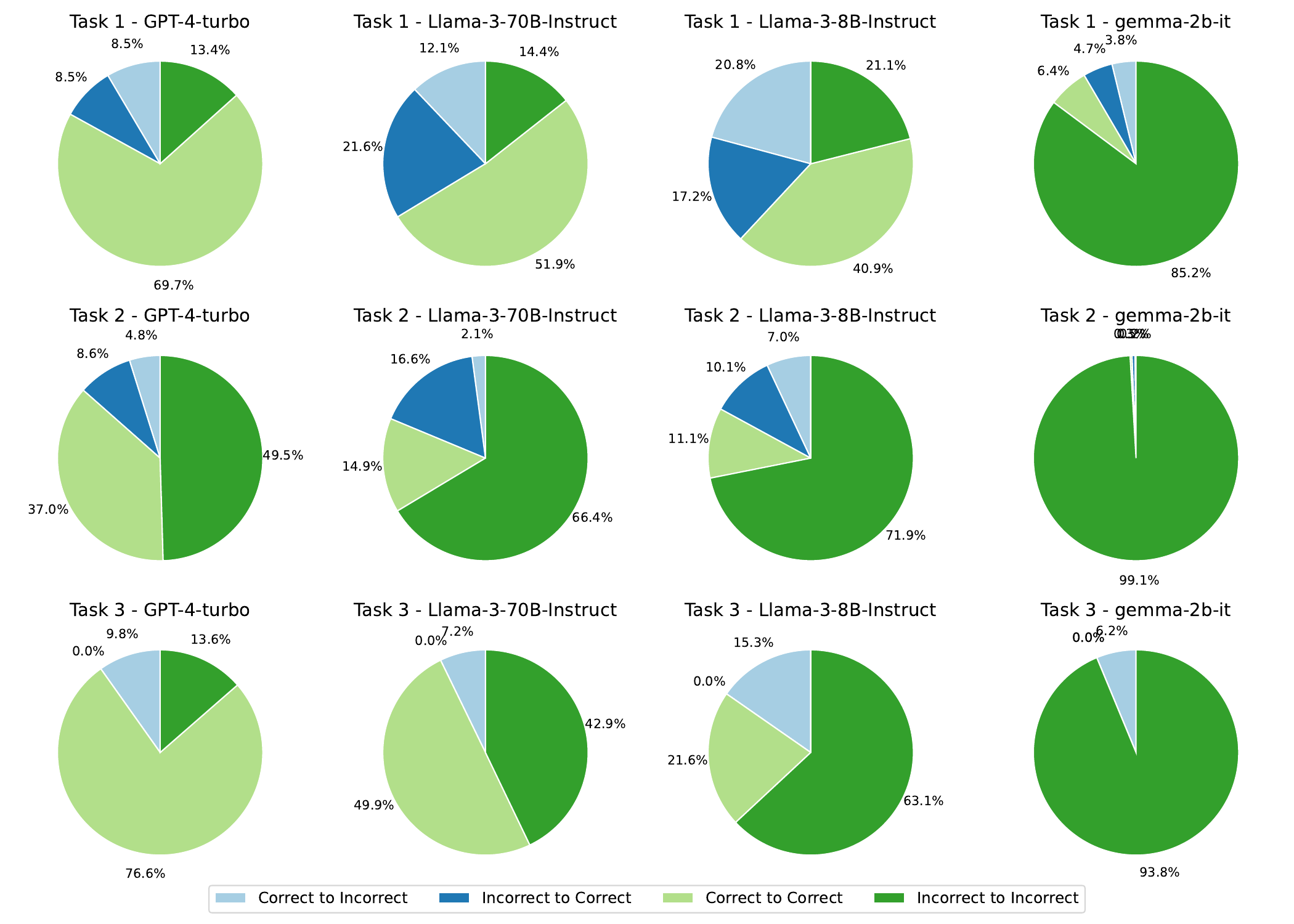}
    \vspace{-0.5cm}
    \caption{This is the performance breakdown for self-refine experiment in the task level, where task 1,2,3 refers to solution correctness, first error step and error reason determination. }
    \label{fig:self_refine_breakdown}
\end{figure}

\begin{figure}[t]
    \centering
    \includegraphics[width=\textwidth]{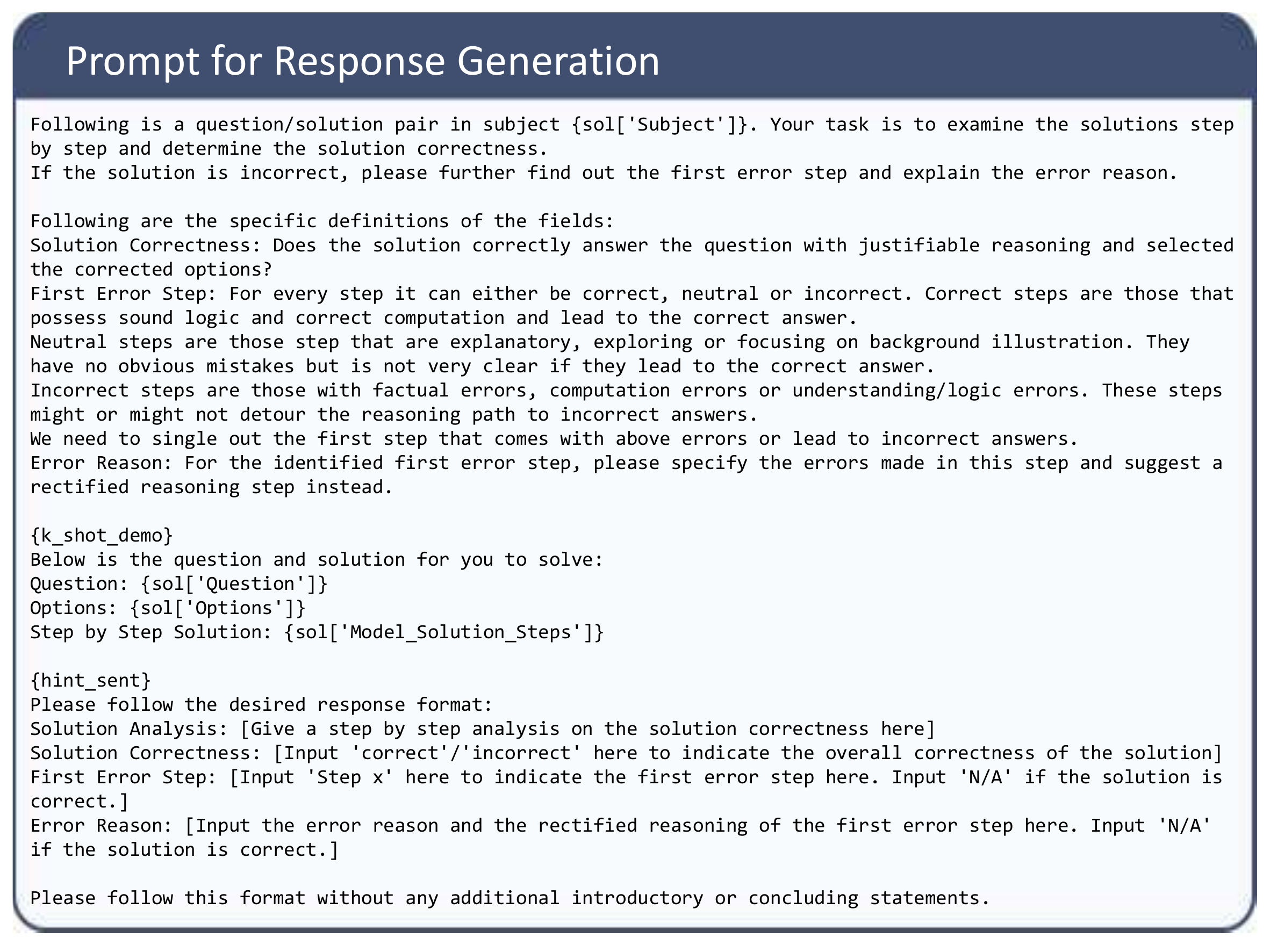}
    \vspace{-0.5cm}
    \caption{This is the prompt template we used to evaluate all the models. The k-shot-demo and hint-sent are either the few shot examples and solution correctness prior or empty string, depending on the experiment setup. }
    \label{fig:eval_prompt}
\end{figure}

\begin{figure}[t]
    \centering
    \includegraphics[width=\textwidth]{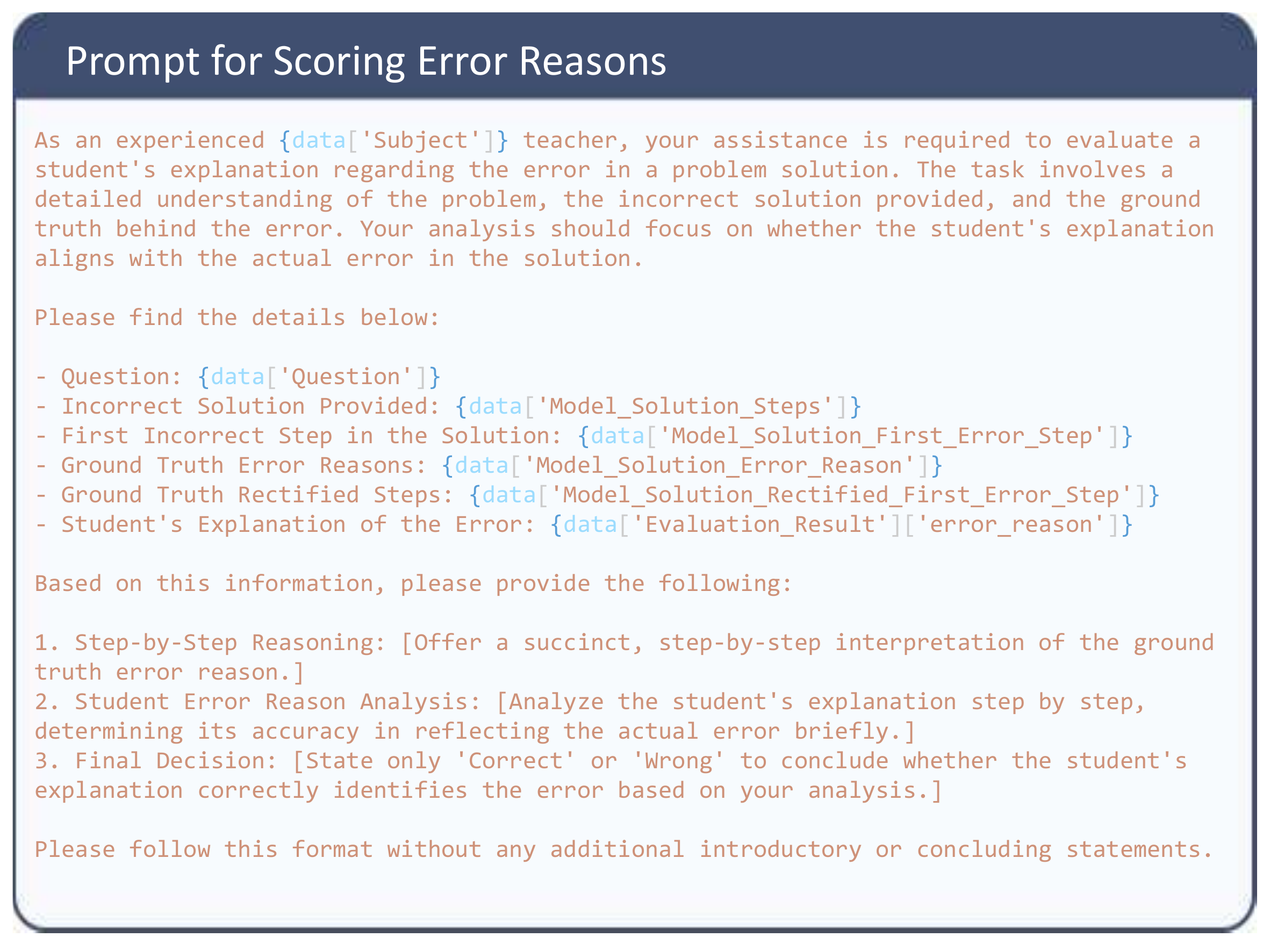}
    \vspace{-0.5cm}
    \caption{This is the prompt template we used to request GPT-4 to help us score the error reasons explained by evaluated models. Note that despite the difficulties of deciding the solution correctness, it is much easier to decide the error reason correctness given the ground truth annotations.}
    \label{fig:error_reason_prompt}
\end{figure}


\section{Evaluation Prompt}
\label{app:Eval_Prompt}

Figure-\ref{fig:eval_prompt} is the prompt template we used to evaluate all the models in our paper. Note that with minor modifications on the following template, the evaluation results can be heavily affected. For example, by introducing a simple hint sentence "Hint: This solution is incorrect. Please focus on looking for the First Error Step and Error Reason.", the model performance can drastically improve as shown in \ref{tab:PriorAccuracy}. Also, by simply taking away the line of 'Solution Analysis' in the response format part of the prompt, the evaluated model will directly output the scoring result without step-wise COT analysis on the solution. This setup will lead to a near zero MR-Score performance as discussed in Section-\ref{sec:experiments}. 

Figure-\ref{fig:sol_gen_prompt} is the prompt we used to query language models for solution generation during the dataset compilation phase. Note that in the prompt, we specifically asked the model to analyse each option in the multiple-choice problem. This is crucial in examining if the model possesses a comprehensive understanding on the topics that the question is asking. 

Figure-\ref{fig:error_reason_prompt} shows the prompt we used to query GPT-4 to score the error reasons returned from evaluated models. Despite the challenging nature of the original task to determine the solution correctness, it is a much easier job to determine if the error reason from the evaluated models aligns with the ground truth error reason. 

Figure-\ref{fig:self_refine_prompt} demonstrates the prompt template we used for self-refine experiment. Note that we followed the setting of \citep{huang2023large} without introducing any prior assumptions or knowledge. This minimum version of extra prompting would mostly rely on the capability of language models to perform self-refine procedure. 

\section{Self Refine Analysis}\label{sec:self_refine}

In this section, we present the results of self-refine in the task level. Specifically, we are looking at the change of labelling by the evaluated models in the determination of solution correctness as shown by Figure-\ref{fig:self_refine_breakdown}. We summarize our observation below:
\begin{itemize}
    \item \textbf{Small Models} like Gemma-2B are too limited to perform effective self-reflection.

    \item \textbf{Competent Models} like GPT4-Turbo are confident in their initial decisions, hardly switching their decisions during self-reflection.

    \item \textbf{Intermediate Models} like Llama3-70B exhibit substantial changes during self-reflection, indicating a lack of consistency in their decisions. However, its change of decisions from incorrect to correct happens to be significantly higher in locating the first error step than in examining solution correctness and explaining the error reason, therefore boosting the overall MR-Score by a large margin. We believe the lack of consistency does not necessarily indicate a more robust or advanced reasoning ability, despite the increase in the evaluation results.

    \item \textbf{Conclusion}: Our results support the observation that LLMs generally lack effective self-refinement capabilities~\citep{huang2023large}.
\end{itemize}

\section{Error Analysis}\label{app:error_cases}

We provide qualitative analyses of how GPT-4 as an example model performed on our benchmark across all seven subjects. The purpose is to offer a deeper understanding of the types and causes of errors made by experimented models to inform future improvements. For each subject in the subsections below, a failure case and a success case are listed. Following the \data~ evaluation framework, each case demonstration consists of the following parts: (1) original questions, options, ground-truth final answers, and LLM-generated CoT solutions; (2) human annotations of step-wise error detection, explanation, and correction; (3) evaluation annotation from the experimented GPT-4 on the aforementioned LLM-generated CoT solutions; (4) scoring results of the error reason if the experimented model identifies the correct first error step.

From our analysis of sampled failure cases, several general observations are made. Firstly, the assessed model GPT-4 exhibits a widely resistant `false positive bias' on our benchmark across all subjects: In cases where the LLM makes incorrect evaluations, the proportion of type I errors is much higher than type II errors. In other words, GPT-4 tends to overlook the mistakes that exist in incorrect model solutions and mislabel them as correct, while seldom actively mislabeling correct model solution steps as incorrect steps. In fact, among the 42 sampled cases we surveyed spanning the seven subjects, all failure cases (size = 21) belong to the type I error category. We provide two possible explanations for such bias: (a) \textbf{input bias}: the implemented LLMs are instructed-tuned, and are inherently biased to follow the prompt input. Therefore, even when the models are asked to fairly judge these CoT solution steps in the prompt input in binary labels, it is likely their labeling threshold is affected and biased towards positive judgments. This is a common issue in using LLMs as generation evaluators and may be mitigated by adjusting the prompt design or other debiasing methods \citep{liu2024aligning, zhou2024batch}; (b) \textbf{self-preference bias}: it has drawn recent attention that state-of-the-art models display self-preference bias: the phenomenon in which an LLM inherently favors their own generated output over texts from other LLMs and humans \cite{liu2024llms, panickssery2024llm}. Therefore, the experiment results of LLMs that are under the same family of the three sampled models (GPT-3.5-Turbo-0125~\citep{GPT35turbo}, Claude2~\citep{Claude2}, and Mistral-Medium~\citep{Mistral23}) may be affected. With the increasingly extensive use of self-evaluation and LLM-as-judge methods, we call for future researchers' attention to the potential issue.



Secondly, the \data benchmark revealed many intricate cases where the assessed model GPT-4 reached a correct final answer through incorrect solution steps, challenging the models' multi-step reasoning capabilities to a greater scale. As shown in the failure cases in math, physics, biology, etc., our benchmark evaluation is able to identify step errors that the sample model made in the solution steps even when its final answer matches the final ground-truth choice. While such step errors can be trivial in terms of generating the correct final answer in the demonstrated failure cases, they can become significant in just slightly nuanced questions, as mentioned in the error analysis section of MMLU \cite{hendrycks2020measuring}. In contrast, our framework, by decomposing the question and model solutions, remains relatively immune to the nuances in question framing. This highlights an important significance of our \data benchmark in that it is not only elaborate but also robust compared to previous benchmarks.

Lastly, there are subtle nuances of model performance in different reasoning paradigms manifest in the case demonstrations of specific subjects. They are interpreted case by case by the captioned figures listed below.

\phantomsection
\label{pdf:all_error_cases}
\includepdf[pages=-]{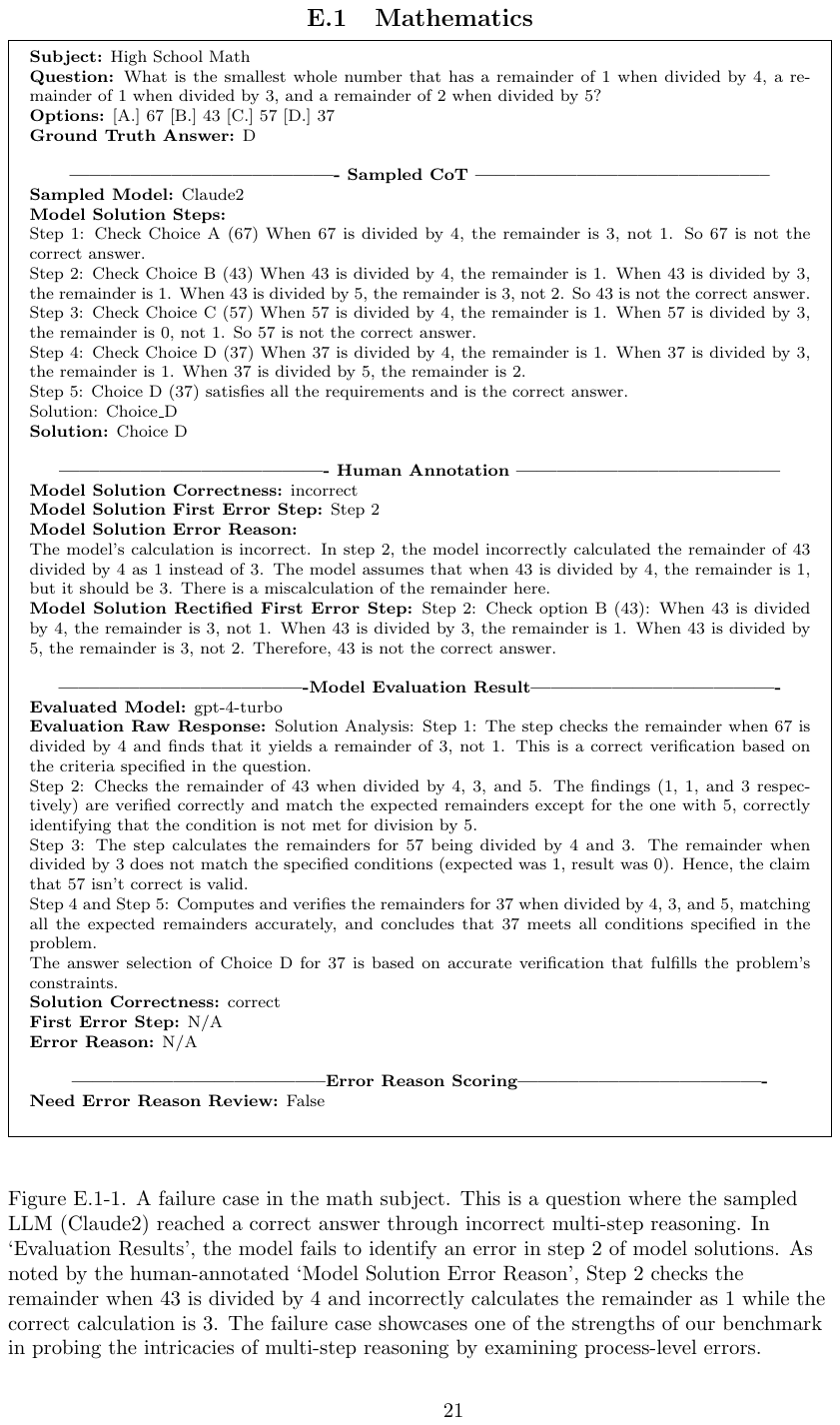}

\section{Computational Resources Used} \label{app:compute-resources}
In this paper, all experiments are either performed on open-source models with local inference or closed-source models with API calls. For local inference, we are using A800 machines with 8 GPUs to run the inferences. The total evaluation time on our 6k benchmark on the 70B language models typically takes around 2 hours using fast inference libraries such as vllm. For smaller language models such as Phi-3 or Gemma, the compute time is smaller.

\section{Annotation Guidelines}
\label{app:guideline}

Below we provide the original annotation guidelines distributed to annotators of distinctive subjects included in the \data benchmark: \hyperref[pdf:math-guidelines]{math}, \hyperref[pdf:biology-guidelines]{biology}, \hyperref[pdf:physics-guidelines]{physics}, \hyperref[pdf:chemistry-guidelines]{chemistry}, \hyperref[pdf:logic-guidelines]{logic}, \hyperref[pdf:medicine-guidelines]{medicine}, and \hyperref[pdf:coding-guidelines]{coding}. The guidelines serve as the primary training material and instructions for annotators to complete the labeling tasks, specified with detailed descriptions, requirements, and standards.


\phantomsection
\label{pdf:math-guidelines}
\includepdf[pages=-]{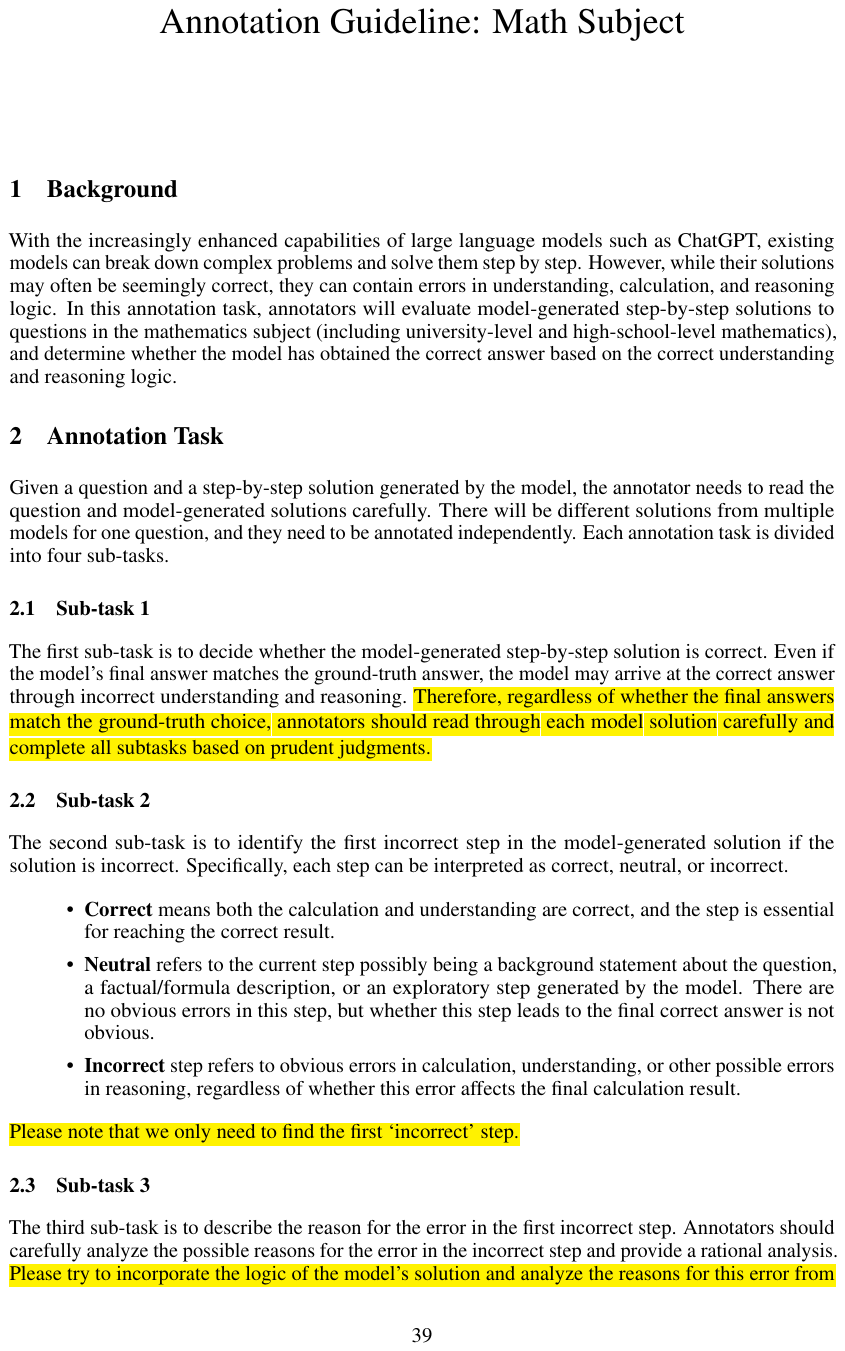}

\phantomsection
\label{pdf:physics-guidelines}
\includepdf[pages=-]{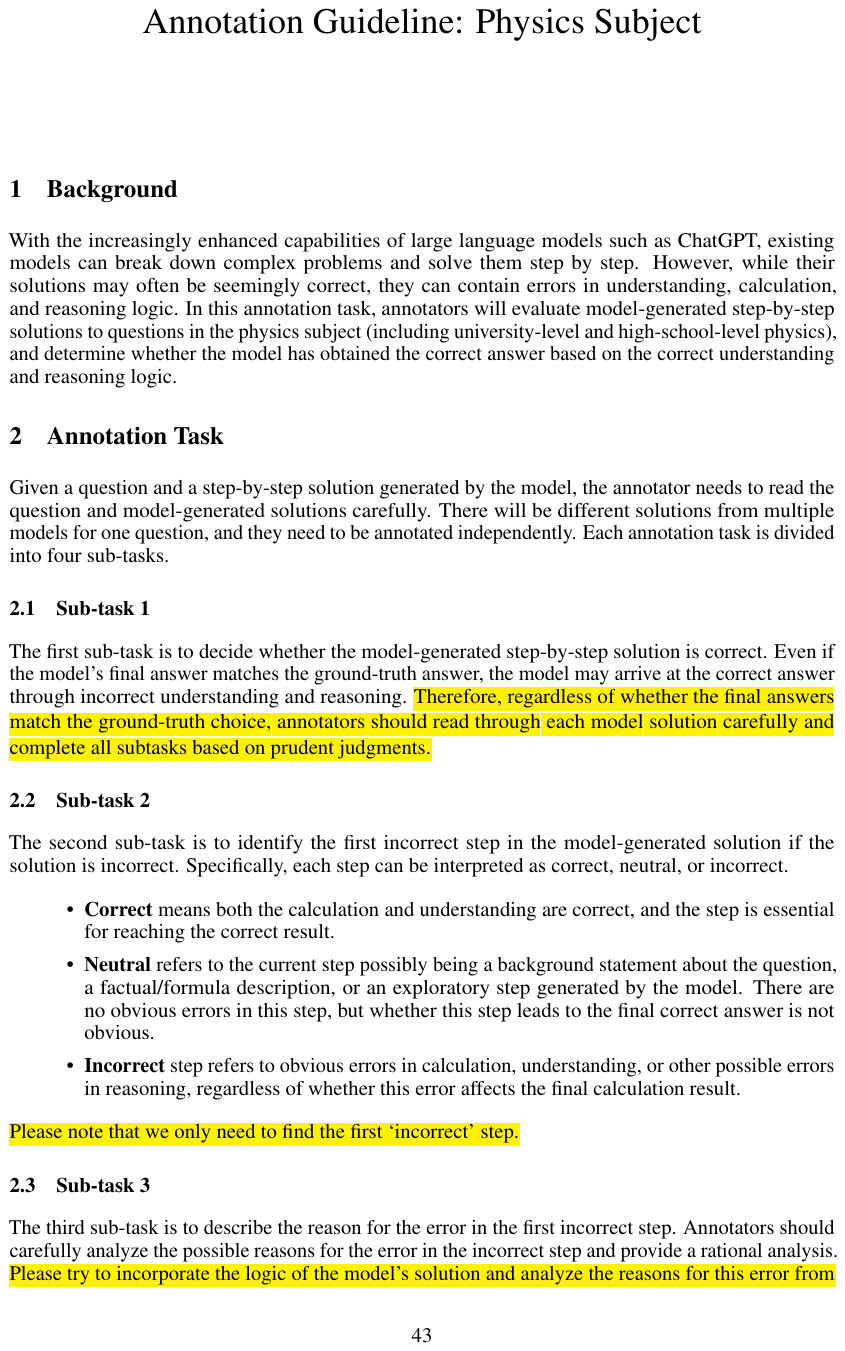}

\phantomsection
\label{pdf:biology-guidelines}
\includepdf[pages=-]{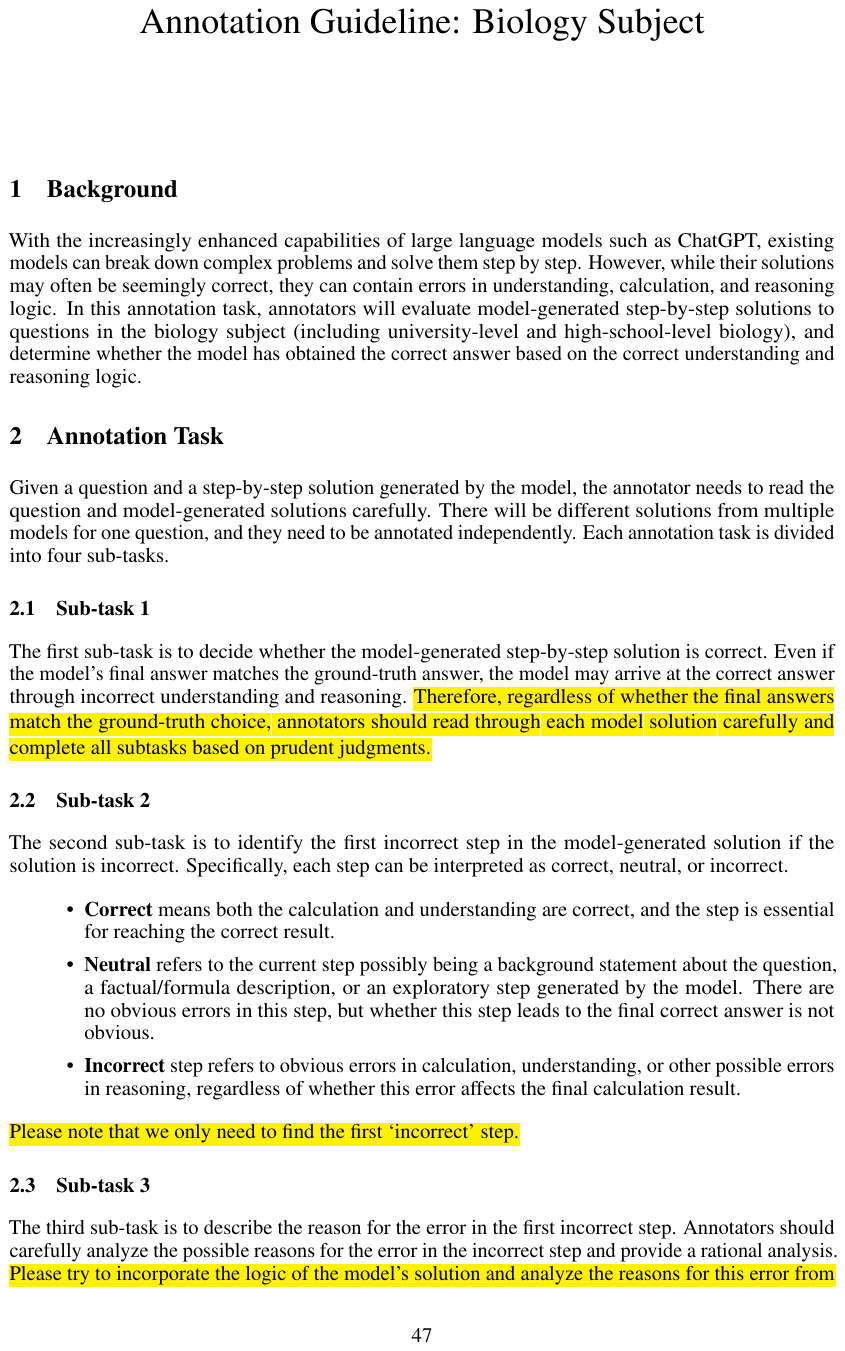}

\phantomsection
\label{pdf:medicine-guidelines}
\includepdf[pages=-]{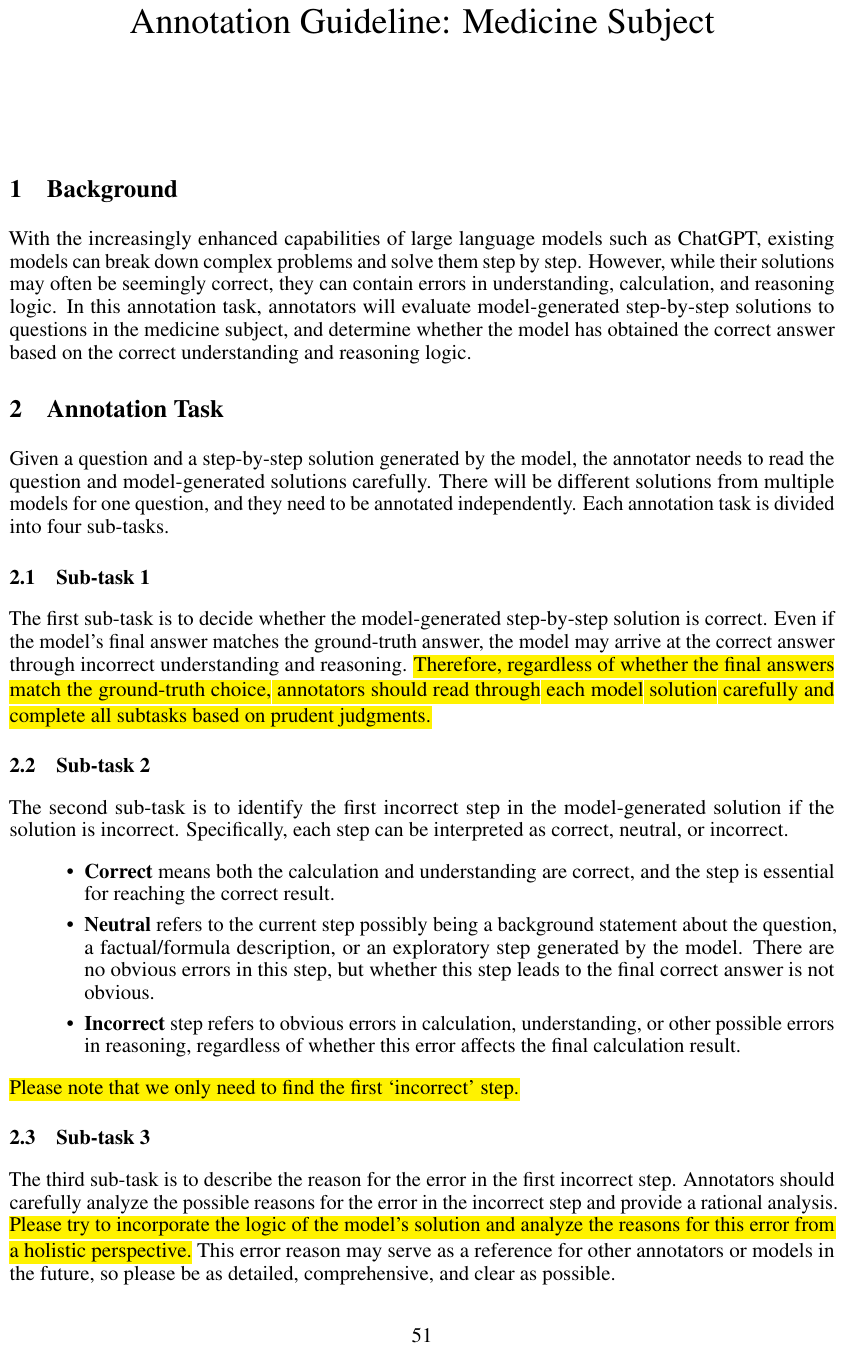}

\phantomsection
\label{pdf:chemistry-guidelines}
\includepdf[pages=-]{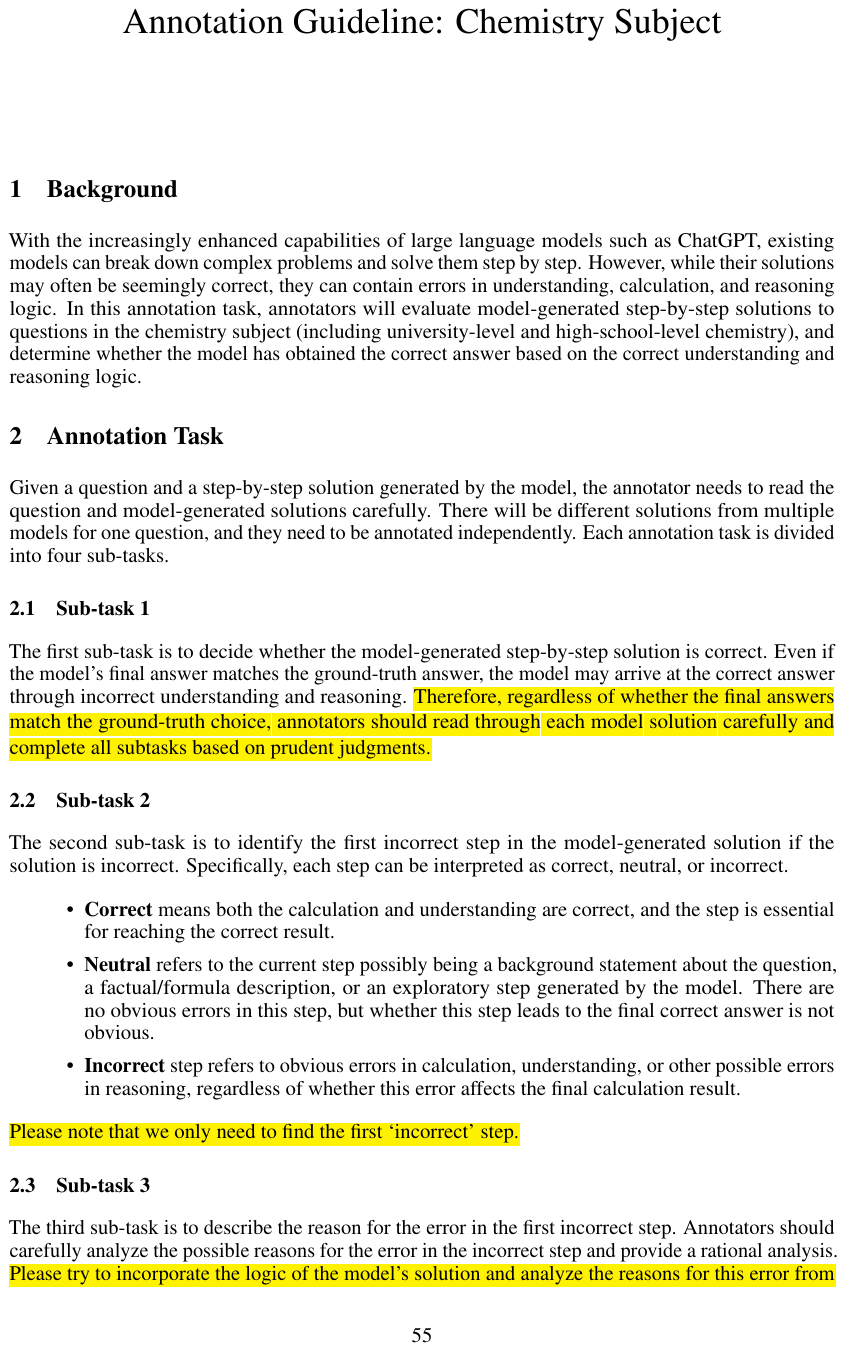}

\phantomsection
\label{pdf:logic-guidelines}
\includepdf[pages=-]{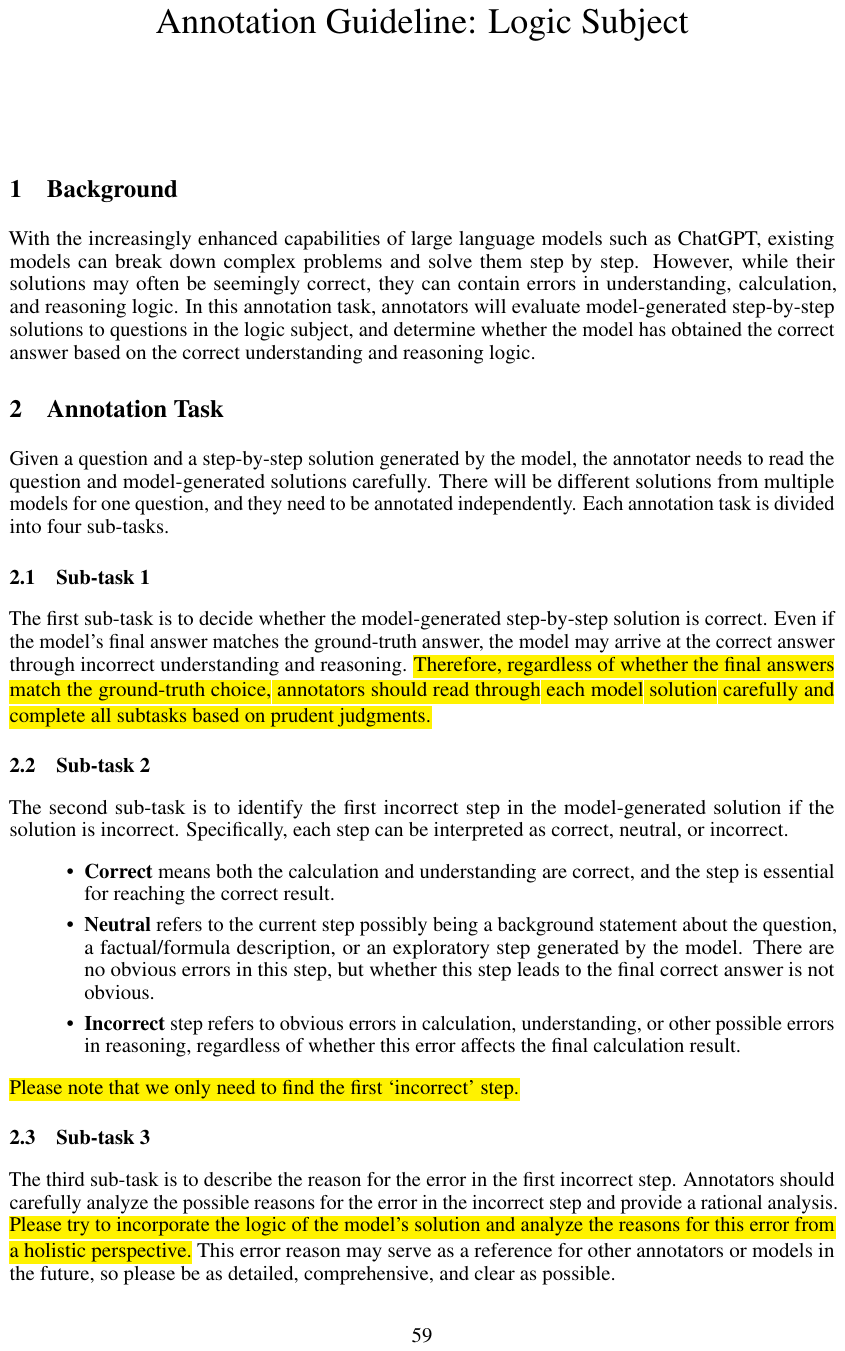}

\phantomsection
\label{pdf:coding-guidelines}
\includepdf[pages=-]{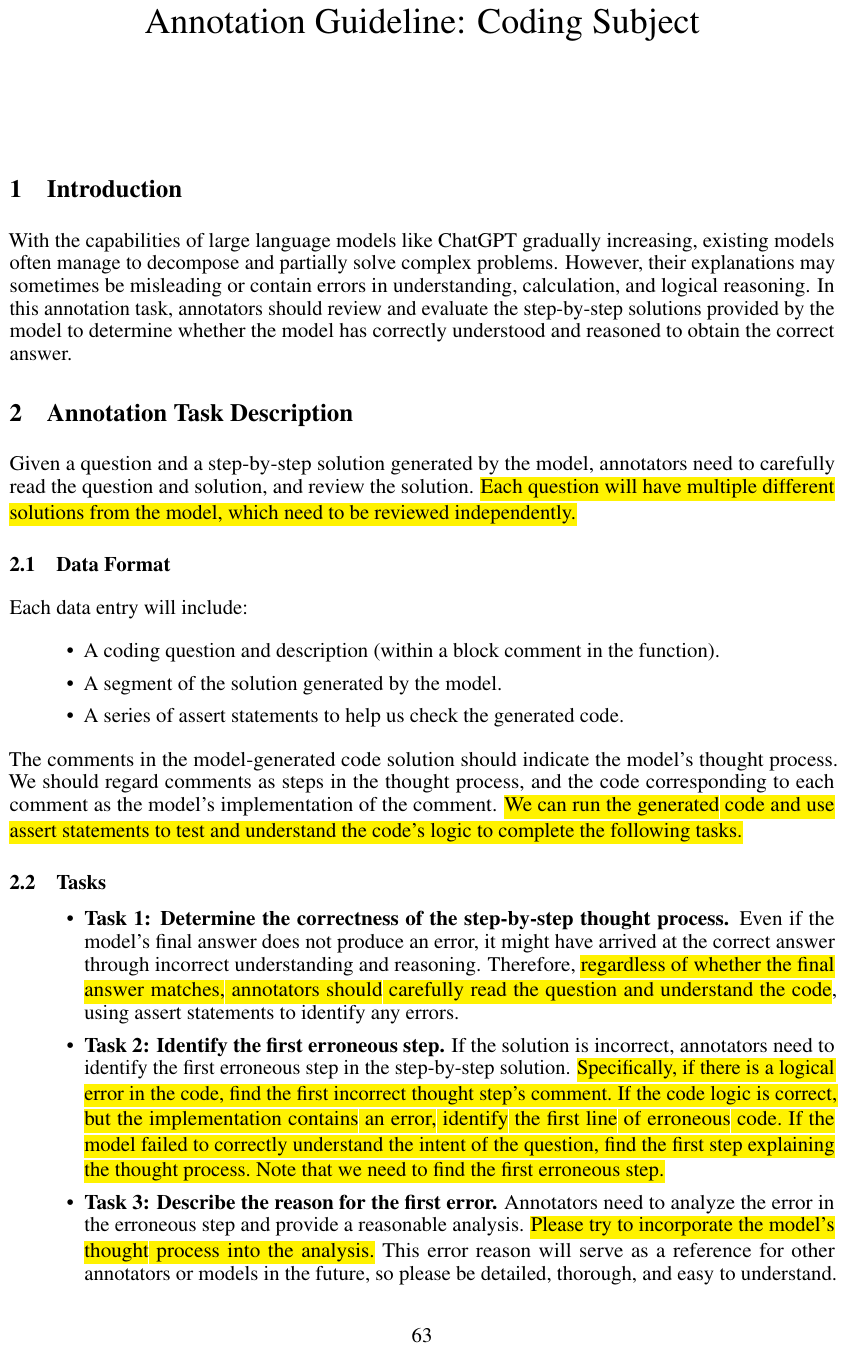}

\clearpage


\newpage
\section*{NeurIPS Paper Checklist}

\begin{enumerate}

\item {\bf Claims}
    \item[] Question: Do the main claims made in the abstract and introduction accurately reflect the paper's contributions and scope?
    \item[] Answer: \answerYes{} 
    \item[] Justification: Yes, the main claims made in the abstract and introduction accurately reflect the paper's contributions and scope.

\item {\bf Limitations}
    \item[] Question: Does the paper discuss the limitations of the work performed by the authors?
    \item[] Answer: \answerYes{} 
    \item[] Justification: The limitations of this work are discussed in \S~\ref{app:limitations}.

\item {\bf Theory Assumptions and Proofs}
    \item[] Question: For each theoretical result, does the paper provide the full set of assumptions and a complete (and correct) proof?
    \item[] Answer: \answerNA{} 
    \item[] Justification: Our work does not involve theoretical assumptions and proofs.

    \item {\bf Experimental Result Reproducibility}
    \item[] Question: Does the paper fully disclose all the information needed to reproduce the main experimental results of the paper to the extent that it affects the main claims and/or conclusions of the paper (regardless of whether the code and data are provided or not)?
    \item[] Answer: \answerYes{} 
    \item[] Justification: We disclose all the information needed to reproduce the main experimental results of the paper to the extent that it affects the main claims and conclusions of the paper, detailed in \S~\ref{sec:experiment setup}.

\item {\bf Open access to data and code}
    \item[] Question: Does the paper provide open access to the data and code, with sufficient instructions to faithfully reproduce the main experimental results, as described in supplemental material?
    \item[] Answer: \answerYes{} 
    \item[] Justification: We opensourced our evaluation benchmark and the script as described in Section-\ref{sec:intro}. Additionally, we have detailed the experimental setup in the paper (\S~\ref{sec:experiments}) , including model selection, hyperparameter settings, data selection, evaluation metrics, hardware resources, etc.

\item {\bf Experimental Setting/Details}
    \item[] Question: Does the paper specify all the training and test details (e.g., data splits, hyperparameters, how they were chosen, type of optimizer, etc.) necessary to understand the results?
    \item[] Answer: \answerYes{} 
    \item[] Justification: We provide comprehensive dataset statistics, evaluation metric descriptions, hyperparameters, and tool usage in \S~\ref{sec:experiments}.

\item {\bf Experiment Statistical Significance}
    \item[] Question: Does the paper report error bars suitably and correctly defined or other appropriate information about the statistical significance of the experiments?
    \item[] Answer: \answerYes{} 
    \item[] Justification: Our proposed method is an inference-only approach for LLM and we adopt the greedy-decoding strategy for all of our experiments, making the experiment results of each session consistent.

\item {\bf Experiments Compute Resources}
    \item[] Question: For each experiment, does the paper provide sufficient information on the computer resources (type of compute workers, memory, time of execution) needed to reproduce the experiments?
    \item[] Answer:\answerYes{} 
    \item[] Justification: We provide comprehensive experimental setup and hardware computation resources used in \S~\ref{app:compute-resources}.
    
\item {\bf Code Of Ethics}
    \item[] Question: Does the research conducted in the paper conform, in every respect, with the NeurIPS Code of Ethics \url{https://neurips.cc/public/EthicsGuidelines}?
    \item[] Answer: \answerYes{} 
    \item[] Justification: We confirm that the research conducted in the paper conform, in every respect, with the NeurIPS Code of Ethics, and all the authors preserve anonymity.

\item {\bf Broader Impacts}
    \item[] Question: Does the paper discuss both potential positive societal impacts and negative societal impacts of the work performed?
    \item[] Answer: \answerYes{} 
    \item[] Justification: The broader impacts of our paper are presented in \S~\ref{app:impacts}.
    
\item {\bf Safeguards}
    \item[] Question: Does the paper describe safeguards that have been put in place for responsible release of data or models that have a high risk for misuse (e.g., pretrained language models, image generators, or scraped datasets)?
    \item[] Answer: \answerNA{} 
    \item[] Justification: Our dataset focuses on evaluation rather than training models. We leverage existing datasets rather than scrape from the Internet.

\item {\bf Licenses for existing assets}
    \item[] Question: Are the creators or original owners of assets (e.g., code, data, models), used in the paper, properly credited and are the license and terms of use explicitly mentioned and properly respected?
    \item[] Answer: \answerYes{} 
    \item[] Justification: All the assets, i.e., codes, data and models used in our paper, are properly credited and we explicitly mention and properly respect the license and terms of use.

\item {\bf New Assets}
    \item[] Question: Are new assets introduced in the paper well documented and is the documentation provided alongside the assets?
    \item[] Answer:  \answerYes{} 
    \item[] Justification: We have submitted the anonymized dataset, codes, and corresponding documents together with the paper.

\item {\bf Crowdsourcing and Research with Human Subjects}
    \item[] Question: For crowdsourcing experiments and research with human subjects, does the paper include the full text of instructions given to participants and screenshots, if applicable, as well as details about compensation (if any)? 
    \item[] Answer: \answerYes{} 
    \item[] Justification: The full text of instructions given to human annotators is presented in \S~\ref{app:guideline}.

\item {\bf Institutional Review Board (IRB) Approvals or Equivalent for Research with Human Subjects}
    \item[] Question: Does the paper describe potential risks incurred by study participants, whether such risks were disclosed to the subjects, and whether Institutional Review Board (IRB) approvals (or an equivalent approval/review based on the requirements of your country or institution) were obtained?
    \item[] Answer: \answerNA{} 
    \item[] Justification: The justification is as follows: We solely engaged human annotators for the dataset, and they were not subjects of our study. Furthermore, we partnered with a legally recognized annotation company in the country, which has obtained all necessary governmental approvals to operate its annotation business.

\end{enumerate}

\end{document}